\documentclass[preprint,12pt]{elsarticle}

\usepackage{graphicx}
\usepackage{amssymb}
\usepackage{amsfonts}
\usepackage{pifont}
\usepackage{lineno}
\usepackage{graphicx}
\usepackage[utf8]{inputenc}
\usepackage{lmodern}
\usepackage[T1]{fontenc}
\usepackage{lscape}
\usepackage{amsmath}	
\usepackage{amssymb}
\usepackage{cases}
\usepackage{graphicx}
\usepackage{caption}
\usepackage{comment}

\usepackage[]{acronym}
\usepackage{makecell}
\usepackage{multirow}
\usepackage{amsfonts}
\usepackage{float} 
\usepackage{diagbox}
\usepackage{fullpage}
\usepackage{url}
\usepackage{rotating}
\usepackage{eurosym}
\usepackage{wrapfig}
\usepackage[final]{pdfpages} 
\usepackage{epstopdf} 
\usepackage[a4paper]{geometry}
\usepackage[nottoc]{tocbibind}
\usepackage{placeins}
\usepackage{float}
\usepackage{listings}
\usepackage{color, colortbl}
\usepackage{hyperref}
\usepackage{xcolor}
\usepackage{graphicx, caption, subcaption}
\usepackage{booktabs}
\usepackage{rotating}
\usepackage{bm}
\usepackage{sidecap}
\newsavebox{\bigimage}
\hypersetup{
colorlinks,
citecolor=black,
filecolor=black,
linkcolor=black,
urlcolor=black
}

\newcommand{\sibo}{\textcolor{green}}

\newcommand{\bx}{\mathbf{x}}

\newcommand{\bA}{\mathbf{A}}

\newcommand{\bV}{\mathbf{V}}

\newcommand{\bW}{\mathbf{W}}

\newcommand{\bU}{\mathbf{U}}
\newcommand{\bX}{\mathbf{X}}

\newcommand{\bI}{\mathbf{I}}

\journal{Information Fusion}

\begin{document}

\begin{frontmatter}

\title{Machine learning for modelling unstructured grid data in computational physics: a review}

\author[CEREA]{Sibo Cheng} \ead{sibo.cheng@enpc.fr}
\author[CEREA]{Marc Bocquet}
\author[Nantong]{Weiping Ding}
\author[CEREA]{Tobias Sebastian Finn}
\author[Tongji]{Rui Fu}
\author[QMUL,Zienkiewicz]{Jinlong Fu}
\author[HKUST]{Yike Guo}
\author[ICL]{Eleda Johnson}
\author[ICL]{Siyi Li}
\author[ICL]{Che Liu}
\author[Tianjin]{Eric Newton Moro}
\author[Cumming1,Cumming2]{Jie Pan}
\author[ICL]{Matthew Piggott}
\author[ICL,Undaunted]{Cesar Quilodran}
\author[UKAEA]{Prakhar Sharma}
\author[ICL]{Kun Wang}
\author[Tongji]{Dunhui Xiao}
\author[UCL1]{Xiao Xue}
\author[Concordia]{Yong Zeng}
\author[ICL]{Mingrui Zhang}
\author[Queensland]{Hao Zhou}
\author[UCL2]{Kewei Zhu}
\author[ICL]{Rossella Arcucci}


\affiliation[CEREA]{organization={CEREA, ENPC, EDF R$\&$D, Institut Polytechnique de Paris},
            state={Île-de-France},
            country={France}}

\affiliation[Nantong]{organization={School of Artificial Intelligence and Computer Science, Nantong University},
            city={Nantong},
            postcode={226019}, 
            state={Jiangsu},
            country={China}}

\affiliation[Tongji]{organization={School of Mathematical Sciences, Key Laboratory of Intelligent Computing and Applications, Tongji University},
            city={Shanghai},
            postcode={200092}, 
            state={},
            country={China}}

\affiliation[QMUL]{organization={School of Engineering and Materials Science, Faculty of Science and Engineering, Queen Mary University of London},
            city={London},
            postcode={E1 4NS}, 
            country={UK}}

\affiliation[Zienkiewicz]{organization={Zienkiewicz Centre for Modelling, Data and AI, Faculty of Science and Engineering, Swansea University},
            city={Swansea},
            postcode={SA1 8EN}, 
            country={UK}}

\affiliation[HKUST]{organization={Department of Computer Science and Engineering, Hong Kong university of science and technology},
            state={Hong Kong},
            country={China}}

\affiliation[ICL]{organization={Department of Earth Science \& Engineering, Imperial College London},
            city={London},
            postcode={SW7 2AZ}, 
            country={UK}}

\affiliation[Tianjin]{organization={Tianjin Key Laboratory of Imaging and Sensing Microelectronics Technology, School of Microelectronics, Tianjin University},
            city={Tianjin},
            postcode={300072}, 
            country={China}}

\affiliation[Cumming1]{organization={Centre for Health Informatics, Cumming School of Medicine, University of Calgary},
            city={Calgary},
            postcode={T2N 1N4}, 
            state={Alberta (AB)},
            country={Canada}}

\affiliation[Cumming2]{organization={Department of Community Health Sciences, Cumming School of Medicine, University of Calgary},
            city={Calgary},
            postcode={T2N 1N4}, 
            state={Alberta (AB)},
            country={Canada}}

\affiliation[Undaunted]{organization={Undaunted, Grantham Institute for Climate Change and the Environment, Imperial College London},
            city={London},
            postcode={SW7 2AZ}, 
            country={UK}}

\affiliation[UKAEA]{organization={UK Atomic Energy Authority (UKAEA), Culham Campus},
            city={Abingdon},
            postcode={OX14 3DB},
            country={UK}}

\affiliation[UCL1]{organization={Centre for Computational Science, Department of Chemistry, University College London},
            city={London},
            postcode={WC1E 6BT}, 
            country={UK}}

\affiliation[Concordia]{organization={Concordia Institute for Information Systems Engineering, Concordia University},
            city={Montreal},
            postcode={H3G 1M8}, 
            state={Quebec (QC)},
            country={Canada}}

\affiliation[Queensland]{organization={School of Mechanical, Medical and Process Engineering, Faculty of Engineering, Queensland University of Technology},
            city={Brisbane},
            state={Queensland (QLD)},
            country={Australia}}

\affiliation[UCL2]{organization={Department of Chemical Engineering, University College London},
            city={London},
            postcode={WC1E 6BT}, 
            country={UK}}

\begin{abstract}
Unstructured grid data are essential for modelling complex geometries and dynamics in computational physics. Yet, their inherent irregularity presents significant challenges for conventional machine learning (ML) techniques. This paper provides a comprehensive review of advanced ML methodologies designed to handle unstructured grid data in high-dimensional dynamical systems. Key approaches discussed include graph neural networks, transformer models with spatial attention mechanisms, interpolation-integrated ML methods, and meshless techniques such as physics-informed neural networks. These methodologies have proven effective across diverse fields, including fluid dynamics and environmental simulations. This review is intended as a guidebook for computational scientists seeking to apply ML approaches to unstructured grid data in their domains, as well as for ML researchers looking to address challenges in computational physics. It places special focus on how ML methods can overcome the inherent limitations of traditional numerical techniques and, conversely, how insights from computational physics can inform ML development. To support benchmarking, this review also provides a summary of open-access datasets of unstructured grid data in computational physics. Finally, emerging directions such as generative models with unstructured data, reinforcement learning for mesh generation, and hybrid physics-data-driven paradigms are discussed to inspire future advancements in this evolving field.
\end{abstract}


\end{frontmatter}


\tableofcontents

\section{Introduction}

Machine learning methods are increasingly being adopted in the field of computational physics to enhance the efficiency of traditional physics-based approaches~\cite{carleo2019machine,willard2020integrating}. These methods have shown significant promise in accelerating simulations, reducing computational costs, and improving the prediction accuracy. When dealing with high-dimensional dynamical systems, if the data structure is regular both spatially and temporally, mature techniques from image and video processing, such as tree-based models, \acp{MLP}, \acp{CNN}, and \acp{RNN}, can be effectively applied to tasks like field prediction, parameter identification, clustering, and super resolution~\cite{fukami2023super,brunton2020machine}. However, these approaches generally require a square grid structure, homogeneous time steps or fixed input dimensions, which limits their applicability to structured grid data. This becomes a significant challenge in cases where data is irregular or unstructured, as is often encountered in high-fidelity simulations~\cite{katz2011mesh,alauzet2016decade}.
Unstructured data often comes in the form of various mesh or grid types, such as triangles and tetrahedra, quadrilaterals and hexahedra, polyhedral meshes, and adaptively refined meshes. Typical applications involve  cerebral hemodynamics~\cite{spiegel2011tetrahedral}, environment modelling~\cite{wang2019effect} and DNA rendering~\cite{benson2015dna}.
These meshes are essential for accurately representing complex geometries in simulations, but they lack the regular structure that traditional machine learning methods typically require \cite{cuomo2022scientific}.

To address these challenges, there is a growing interest in developing machine learning techniques that can manage unstructured data, common in many areas of computational physics. An important family of approaches involves using machine learning models in conjunction with grid interpolation (e.g., nearest neighbor~\cite{loehner2001overlapping}) and reordering techniques (e.g., space-filling curves~\cite{heaney2024applying}). These methods map unstructured or sparse data onto a structured grid, enabling the application of traditional deep learning models such as \ac{CNN}. A recent representative work is the Voronoi-tessellation-assisted \ac{CNN}~\cite{fukami2021global,cheng2024efficient}. Although effective, this approach can introduce interpolation errors and may not fully capture the complexities of the original unstructured data.

Another promising direction is the use of \acp{GNN}, which are specifically designed to work with data represented as values on graphs rather than in grids \cite{wu2020comprehensive}. In computational physics, where the data often consists of points connected in an irregular manner (such as nodes in a mesh), \acp{GNN} excel by directly modelling the relationships between these points. Importantly, \acp{GNN} can also handle adaptive meshes with a time-varying number of nodes or grids, allowing them to dynamically adjust to changes in the mesh structure during simulations~\cite{pfaff2020learning,perera2023dynamic}. This capability makes \ac{GNN}s highly effective at capturing local interactions and generalising across various and evolving mesh configurations.

Transformer-type neural networks, equipped with spatial attention mechanisms~\cite{chu2021twins,Vaswani2017}, have also shown great potential in handling unstructured data. These models can focus on relevant spatial features regardless of the data’s irregular structure, allowing for improved learning and generalisation. Like \acp{GNN}, transformers can manage adaptive meshes where the number of nodes or grids changes over time, providing flexibility in scenarios where the relationships between data points are complex and non-local~\cite{geneva2022transformers}. The ability to dynamically adjust focus based on the importance of different regions in the data makes transformers particularly useful in applications involving complex physical systems with evolving geometries.

\acp{PINN}~\cite{raissi2019physics,cuomo2022scientific,karniadakis2021physics} offer another innovative solution, particularly by enabling meshless predictions. Unlike traditional methods that require a predefined mesh, \acp{PINN} can directly incorporate physical laws into the learning process without relying on a specific grid structure. This meshless approach avoids the challenges associated with unstructured meshes. \acp{PINN} can effectively model complex physical phenomena by embedding the governing equations of the system into the neural network architecture~\cite{raissi2019physics}, ensuring that predictions remain physically consistent. This makes \acp{PINN} ideal for a wide range of applications where mesh generation can be particularly challenging.

On the other hand, by adapting the advanced machine learning techniques previously discussed, modern AI paradigms like \ac{RL} and generative AI could also play a crucial role in enhancing the modelling of dynamical systems with unstructured grid data. 
Reinforcement learning is increasingly applied to optimise unstructured meshes by dynamically adjusting mesh elements based on solution variability, reducing reliance on heuristic rules~\cite{foucart2023deep}. Its success has been recently observed in simulations of fluid dynamics~\cite{lorsung2023mesh,kim2024non}. 
On the other hand, generating unstructured grid data, particularly for spatial-temporal systems, has been a long-standing challenge in the generative AI community. Efforts have spanned \ac{VAE} \cite{kingma2019introduction}, \ac{GAN} \cite{goodfellow2020generative}, and score-based diffusion models \cite{song2020score,croitoru2023diffusion}. The latter have recently been widely applied to dynamical systems due to their robustness to sparse observations \cite{jacobsen2023cocogen,zhuang2025spatially} and flexibility in conditioning \cite{price2023gencast,finn_generative_24}.

\begin{figure*}[!ht]
\centering
\includegraphics[width=0.8\textwidth]{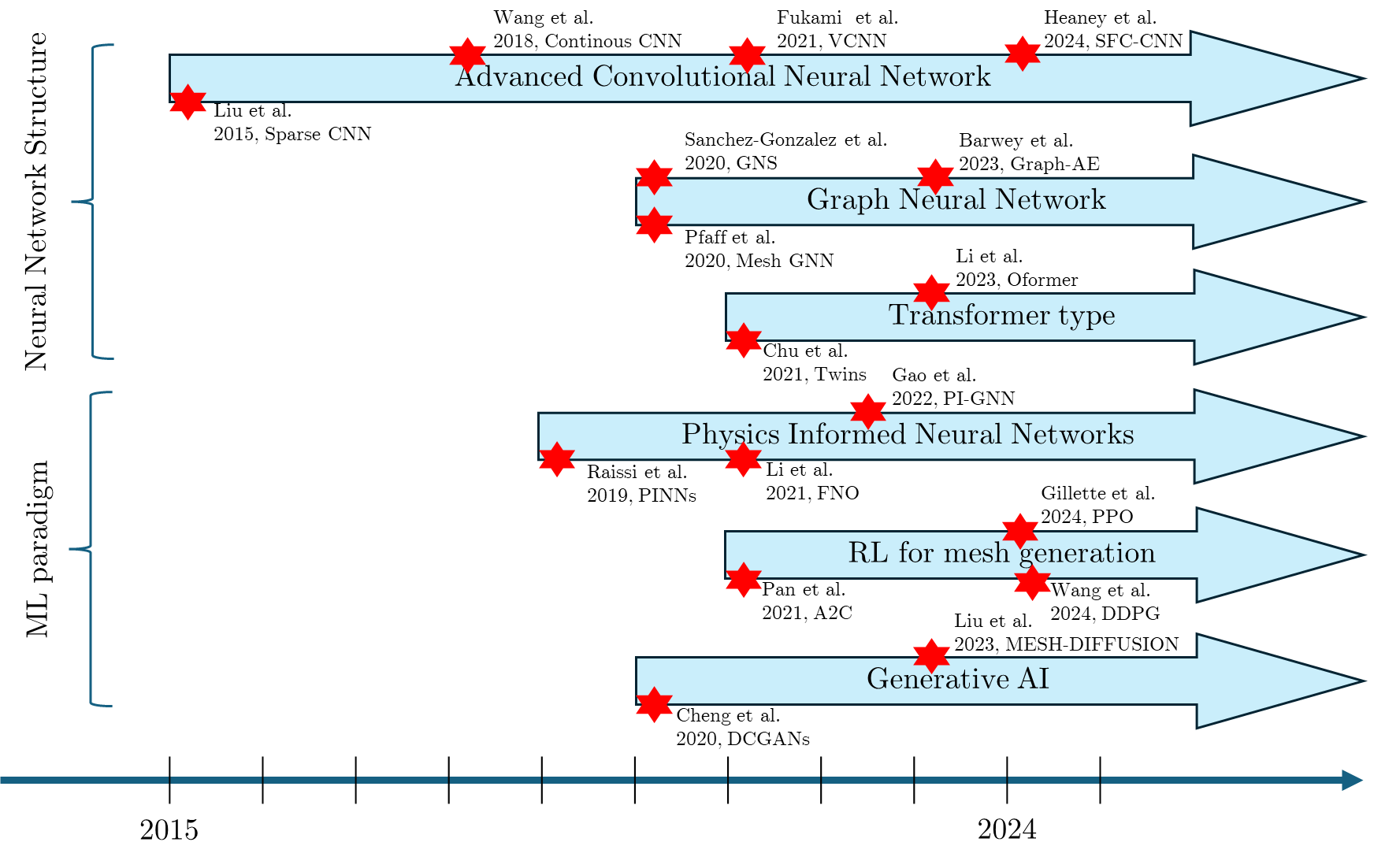}
\caption{Timeline of some representative works that apply advanced machine learning techniques for unstructured grid data}
\label{fig:timeline}
\end{figure*}

These emerging \ac{ML} techniques, as shown in Figure~\ref{fig:timeline}, are revolutionising how unstructured data is processed in computational physics, enabling more accurate, efficient, and flexible simulations of complex systems. This survey explores the latest advancements in machine learning methods specifically tailored for modelling unstructured data, focusing on how these approaches are being adapted to overcome the limitations of traditional techniques and improve the overall efficiency and effectiveness of simulations in complex physical systems. We emphasise that the goal of this work is not to compare the performance of existing methods, as they were designed to solve different challenges. In summary, the key contributions of this paper are as follows:

\begin{itemize}
\setlength{\itemsep}{0pt}
\setlength{\parsep}{0pt}
\setlength{\parskip}{0pt}
\item To the best of the authors’ knowledge, this is the first review paper to comprehensively examine the application of machine learning techniques to unstructured grid data in dynamical systems within computational science.
\item This paper has a special focus on how cutting-edge machine learning techniques, in particular deep neural networks, could tackle the bottleneck of data sparsity and irregularity challenges in simulation data with unstructured or adaptive grids. 
\item This review covers a variety of significant applications in computational science, such as environmental simulation and multiphase flow modelling involving complex geometries, though it does not provide an in-depth analysis of applications, as the primary focus of this paper is on the methodology.
\end{itemize}

The rest of the paper is organised as follows. Section~\ref{Preliminary} introduces the concepts of dynamical systems, irregular grid structures, and data-driven techniques in dynamical systems, ranging from reduced-order modelling and shallow machine learning methods to deep neural networks. Section~\ref{sec:ML_unstructure} discusses advanced neural network architectures and processing techniques for handling data with unstructured grids. A qualitative comparison is provided at the end of this section. Machine learning paradigms, including \acp{PINN}, \ac{RL}, and generative AI, are introduced in Section~\ref{sec:paradigms}. We also list current open-access datasets for benchmarking machine learning approaches with unstructured grids in computational physics in Section~\ref{sec:dataset}. The paper concludes with Section~\ref{sec:conclusion}.

\section{Preliminary}
\label{Preliminary}

In this section, we provide a brief overview of the concept of unstructured grid data in computational science. This is followed by an introduction to \ac{ML} methods, including both shallow approaches (such as \ac{RF} and Gaussian processes) and deep neural networks (primarily \ac{CNN} and \ac{RNN}) for dynamical systems. 
Reduced-order modelling and spatial interpolation techniques are also discussed in this section, as they are important strategies for handling unstructured data and have been integrated into many \ac{ML} models.

\subsection{Unstructured meshes in computational physics }

To describe the behaviour or state of dynamical problems, such as airflow around a spacecraft or stress concentration in a dam, numerical simulations require a finite number of points (in time and space) to describe the fields of physical quantities throughout the whole computational domain.  Numerical methods, such as the finite difference method, finite volume methods, and finite element analysis, are developed to discretise the equations governing these dynamical problems, allowing for the calculation of system's behaviours over time and space shown by variable values at discrete points throughout the computational domain.  In the implementation of these numerical methods, the mesh is the foundation for numerical simulations. Its quality and structure significantly influence the simulations' accuracy, efficiency, and stability, especially in complex simulated domains. To address this situation, unstructured meshes are noted for adaption on irregular boundaries in simulation-based fluid and solid dynamical systems.  Their flexibility and efficiency in handling complex geometries and dynamic problems make them highly suitable for accurately representing the diverse and changing conditions typical of real-world physical systems.

\par

Meshes discretise domains in a space-filling manner such that they provide a description of the associated computational topology and geometry for the fields and equations of functional data. The quality and features of a mesh domain discretisation have been shown through many examples to limit the performance quality of the numerical scheme being employed~\cite{Loseille2017}. Unstructured meshes are advantaged by an arbitrary structure, allowing for flexibility in conformance to geometries and boundaries of the physical system they are used to represent. The topology of meshes typically is arranged hierarchically, built up from nodes connected by edges forming loops which in turn define two-dimensional faces which can be grouped individually or collectively into elements~\cite{TaylorHaimes2018}. Geometrically, the node is related to a point, an edge to a curve and a face to a two-dimensional surface. These components, when referenced in a discretised mesh are sometimes referred to as vertex and segment, respectively~\cite{ParkEtAl2019}.

In general, unstructured meshes are constructed space-filling elements of a signal or mixed shape-type based on the spatial dimension of the domain~\cite{FreyGeorge2008}. In one-dimension, this is a range of finite segments of potentially varying lengths. In two-dimensional surface meshes, a simplex or triangle is the most widely utilised element, though quadrilaterals or a mix of the two are also permissible. Three-dimensional, unstructured volume meshes typically replace the surface mesh triangular base element with a tetrahedral form. Quadrilaterals (pentahedra, hexahedra, prisms, etc.) and other element volumetric shapes are also utilised~\cite{bern2000mesh}. The associated two-dimensional surface meshes can serve as a template for constraining the geometry of the space-filling tetrahedra~\cite{Loseille2017}.

Unstructured surface and volume meshes are advantaged in their flexibility to conform to potentially complex geometries found in both static and dynamical applications. Controlling the size, shape and/or orientation of the mesh element topology allows for problem specific optimisation. On one hand, principles of equidistribution~\cite{BuddEtAl2009} are used to generate unstructured isotropic meshes with elements of roughly the same size, shape and orientation. Isotropic meshes are advantageous for capturing turbulent flows and more uniformly distributed phenomena~\cite{Loseille2017}. On the other hand, anisotropic meshes allow for a larger range of element sizes, shapes and orientations. Anisotropic meshes can be constructed or optimised such that the size, shape and distribution of elements are designed to align with similar anisotropic physical features such as those seen in shocks~\cite{Mavriplis1990a} or strongly directional flow such as tidal regimes~\cite{PiggottEtAl2008}.

The unstructured mesh features which advantage them in complex geometric and multi-scale problem applications can also lead to increased overhead in memory usage, algorithmic complexity, and overall computational cost. For the same node count, unstructured meshes generally have higher memory requirements to store all variations of connectivity and additional topology information unique to arbitrary cells~\cite{Ito2013}. In applications such as in geometrical volume-of-fluid methods extension, where unstructured meshes allow extension to more complex domains and improved accuracy, implementation has lagged due to the complex geometrical operations required to handle the resultant arbitrary, and non-orthogonal orientations~\cite{MaricEtAl2020}. Often there is a trade-off between conducting fast, stable simulations needed for deployed tool performance and high enough mesh resolution to minimise costly numerical prediction errors~\cite{spiegel2011tetrahedral}.

\begin{figure*}[!ht]
\centering
\includegraphics[width=0.8\textwidth]{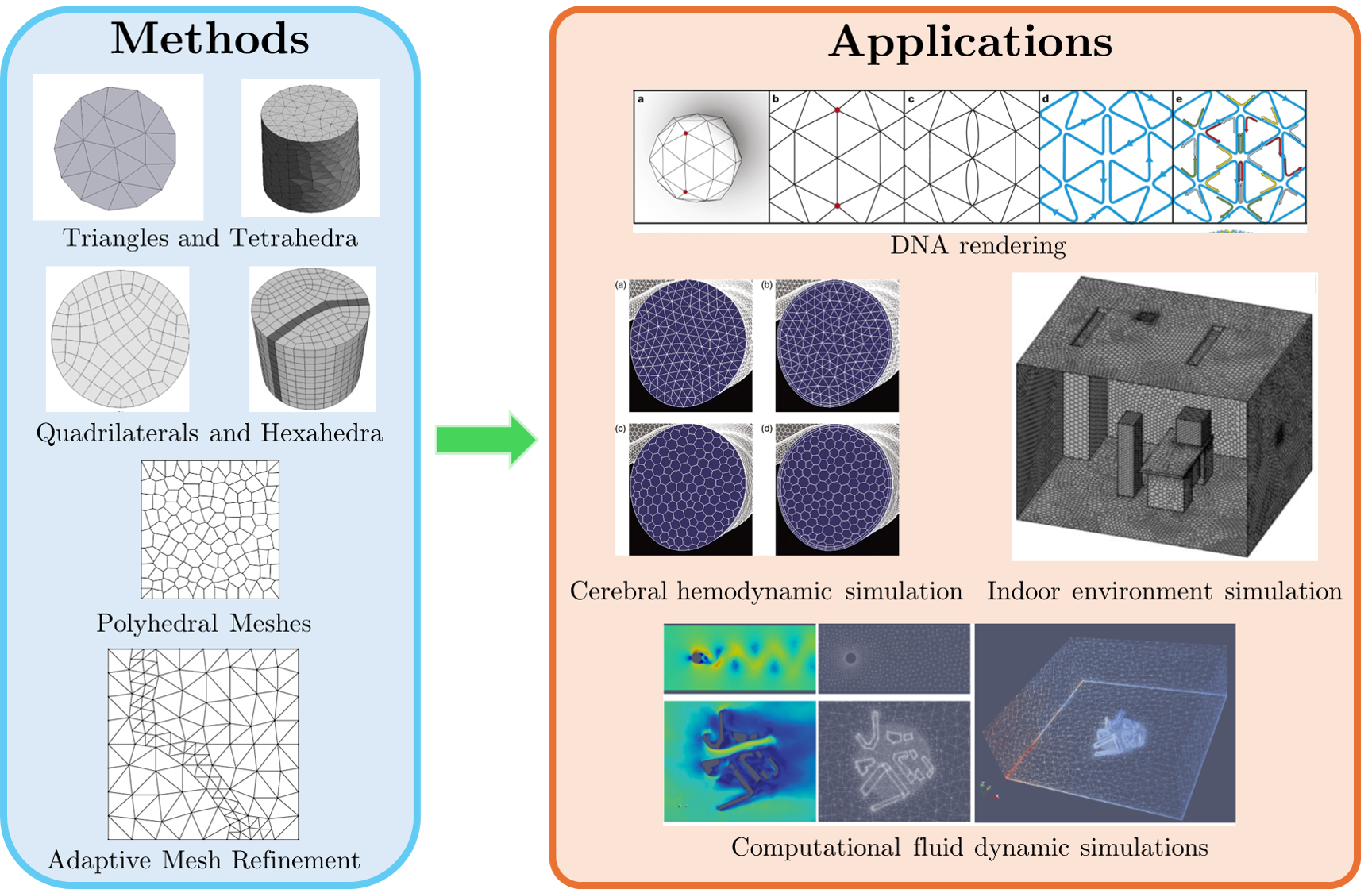}
\caption{Methods and applications of unstructured mesh}
\label{structure_grid}
\end{figure*}

Unstructured meshes are commonly employed in modelling of structures, both to adequately capture complex physical geometries present and to enable more efficient \ac{CFD} simulations. Quadrilateral meshes (as seen in Figure~\ref{structure_grid}) are aptly suited to modelling complex, holistic aircraft designs where structural member interactions need to be captured and can accelerate airframe structural analysis in automated workflows~\cite{HwangMartins2016}. The flexibility of unstructured meshes is further illustrated in the coupling of complex fluid-structure interactions, where evolving flow dynamics along boundary conditions requires compatibility of structure and spatial discretisation between the different domains~\cite{SloneEtAl2002}. The feasibility of a 3D finite-volume method for dynamic fluid-structure interactions was shown in~\cite{SloneEtAl2002}  using a loaded fixed-free cantilever wing-like structure in incompressible flow with no turbulence modelling. However, without further optimisation, the cost of extending this workflow to more complex geometries and flow physics, such as flutter dynamics, was predicted to be computationally prohibitive~\cite{SloneEtAl2002}.


\subsection{Reduced order modelling for structured and unstructured data} 
\label{sec:POD}

Despite current advancements in computer technology, computational mesh sizes remain constrained by available computational resources, especially unstructured mesh, which produces a much larger number of meshes for the same conditions of dynamical problems~\cite{baker2005mesh, katz2011mesh}. For example, to improve the computational accuracy, a rectangle is usually divided into 5 tetrahedra~\cite{chen1996domain}. The \ac{ROM} technique has become a vital method for reducing computational burdens through low-dimensional reduced models~\cite{brunton2020machine}. These models are fast to solve and approximate well high-fidelity simulations of dynamic systems~\cite{benner2015survey, Rowley2017Model}. As scientific development focus shifts from first-principles to data-driven approaches, data-driven ROMs are developed in optimising ROMs by utilising existing data efficiently~\cite{brunton2020machine, Arcucci2023Reduced}. Especially, non-intrusive reduced order modellings (NIROMs) have become popular across various research and engineering fields~\cite{Xiao2015Nonintrusivea, audouze2013nonintrusive, LeGuennec2018Parametrica, hesthaven2018non, Guo2018Reduceda, Guo2019Datadrivena, Wang2018Modelb}. They operate independently as robust models, offering accurate descriptions from high-fidelity simulations without any modifications to original source codes~\cite{Xiao2019Reduced}. It is important to note that \ac{ROM} techniques can also be considered a means of transforming unstructured data into a regular and often fixed-size reduced space, which facilitates downstream tasks such as field prediction or parameter identification.
 
In \ac{ROM}s for computational simulations, \ac{POD}, also known as \ac{PCA} or \ac{EOF} has proven to be a classic method for spatial dimensionality reduction on data~\cite{benner2015survey}. \ac{POD} and its variants have been applied with unstructured meshes in various research areas successfully, such as eigenvalue problems in reactor physics~\cite{Buchan2013PODa}; ocean models~\cite{Fang2009Pod}; aerospace design and optimisation~\cite{Manzoni2015Reduced}; fluid dynamics with applications in porous media~\cite{Luo2013Reducedordera}, Navier-Stokes Equations~\cite{Du2013PODa, Xiao2015Nonintrusive}, air pollution~\cite{Fang2014Reduced}, and shallow water equations~\cite{Stefanescu2013POD}. It is often applied with Galerkin projection to form intrusive \ac{ROM}s and applied with various interpolation or regression methods, like radial basis function, neural network in NIROMs~\cite{audouze2013nonintrusive, Xiao2019Reduced, Mohan2018Deep, Wu2020Data}.

Developed from \ac{SVD} techniques, \ac{POD} aims to identify an optimal least squares subspace for approximating data sets. In this method, we assumed that any variable $\mathbf{X} \in \mathbb{R}^{N_s\times N_h}$ could be expressed as: 
\begin{equation}
    \mathbf{X} = \sum_{i=1}^{N_h}\alpha_i\phi_i,
\end{equation}
where $N_s$ denotes the total number of snapshots and $N_h$ denotes the general number of nodes on meshes. $\alpha_i$ denotes the $i$-th \ac{POD} coefficient and $\phi_i$ denotes the  $i$-th \ac{POD} basis function. 

\begin{figure*}[!ht]
\centering
\includegraphics[width=1.\textwidth]{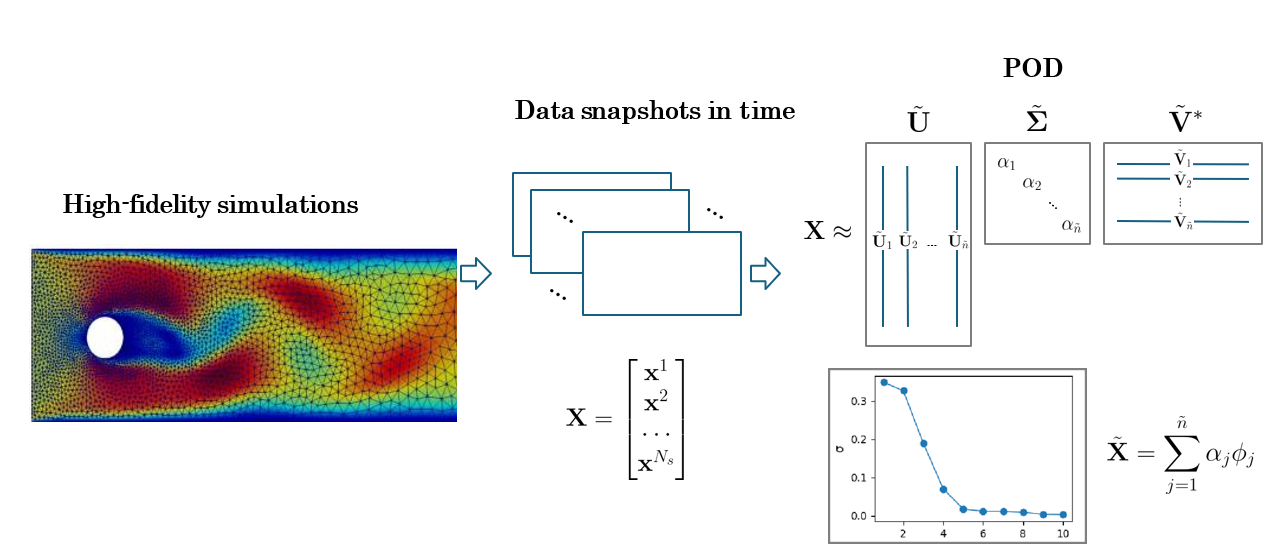}
\caption{POD of High-Fidelity Simulation Data Over Time}
\label{fig: POD}
\end{figure*}

The \ac{POD} basis functions are obtained through \ac{SVD} applied on discredited numerical solutions $\mathbf{X}$, which could be shown as 
\begin{equation}
\mathbf{X} = \begin{bmatrix}
\mathbf{x}^{1}\\ 
\mathbf{x}^{2}\\ 
\cdots \\ 
\mathbf{x}^{N_s}
\end{bmatrix}.
\end{equation}
The \ac{SVD} computed on $\mathbf{X}$ forms:
\begin{equation}\label{SVD_eq}
    \mathbf{X} = \mathbf{U} \mathbf{\Sigma} \mathbf{V^{\ast}},
\end{equation}
where orthogonal matrix $\mathbf{U}\in  \mathbb{R}^{N_s\times N_s}$, $\mathbf{V^{\ast}}\in  \mathbb{R}^{N_h\times N_h}$ contains the left and right singular vectors separately, matrix $ \mathbf{\Sigma}\in  \mathbb{R}^{N_s\times N_h}$ denotes the singular values of $\mathbf{X}$ and these values are listed by their magnitude. In the matrix $\mathbf{U}$, the columns represent the \ac{POD} basis functions or modes, with the singular values indicating the importance of each basis function. The square of these singular values signifies the energy captured by each basis function. Therefore, basis functions corresponding to low singular values contribute less to the system's overall energy and can be ignored. The energy capture formula applies this principle: for a given tolerance $\eta$, the smallest integer $\tilde{n}$ is determined by:
\begin{equation}
    \frac{\Sigma _{i=1}^{\tilde{n}}\sigma _{i}^{2}}{\Sigma _{i=1}^{N_h}\sigma _{i}^{2}}\geq \eta ,
\end{equation}
where the left side of the equation is the fraction of total energy captured or variance of the system by the first $\tilde{n}$ 
\ac{POD} basis functions and $ \tilde{n}$ is much less than $N_h$, the total number of singular values. Following this truncation, an approximation $\tilde{\mathbf{X}}$ of the variables $\mathbf{X}$ could be calculated by truncated matrices and expressed as a linear combination of the retained \ac{POD} basis functions:
\begin{equation}
   \tilde{\mathbf{X}} = \mathbf{X}\cdot \tilde{\mathbf{V}} \approx  \tilde{\mathbf{U}}  \tilde{\mathbf{\Sigma}} \tilde{\mathbf{V^{\ast}}}\cdot \tilde{\mathbf{V}}= \tilde{\mathbf{U}}  \tilde{\mathbf{\Sigma}}=\sum_{j=1}^{\tilde{n}} \alpha_{j}\phi_j,
\end{equation}
where truncated matrices $\tilde{\mathbf{U}} \in  \mathbb{R}^{N_s\times \tilde{n}}$, $\tilde{\mathbf{V^{\ast}}}\in  \mathbb{R}^{\tilde{n}\times N_h}$  and $ \tilde{\mathbf{\Sigma}}\in  \mathbb{R}^{\tilde{n}\times \tilde{n}}$ are used to create a reduced-order dataset $\tilde{\mathbf{X}}$ that maintains the most significant patterns in the data $\mathbf{X}$. This process is illustrated on flow past a cylinder case simulations in Figure~\ref{fig: POD}.

\begin{figure*}[!ht]
\centering
\includegraphics[width=1.\textwidth]{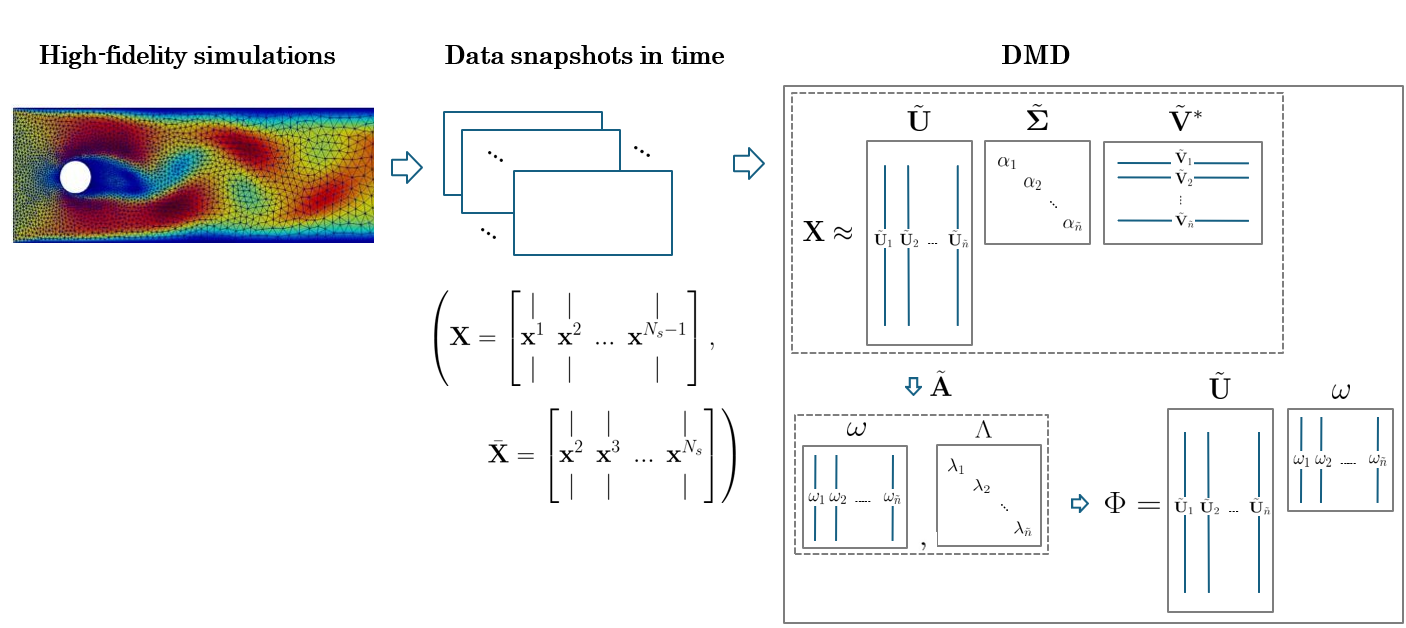}
\caption{DMD of dynamic modes in high-fidelity simulation data}
\label{fig: DMD}
\end{figure*}

Moreover, \ac{DMD} is also developed for \ac{ROM}s  based on the \ac{SVD} technique~\cite{schmid2022dynamic, kutz2016dynamic}. It is defined to identify low-order dynamics, providing insights into the system evolution over time~\cite{schmid2010dynamic}. \ac{SVD} in \ac{DMD} serves as a dimensionality reduction and feature extraction tool to identify the dominant spatial and temporal patterns in complex systems~\cite{tu2013dynamic, taira2017modal}. Furthermore, \ac{DMD} was developed to connect to the underlying nonlinear dynamics through Koopman operator theory~\cite{koopman1931hamiltonian} by Mezi\'{c}~\cite{Mezic2005Spectral, Mezic2013Analysis}, and Rowley~\cite{Rowley2009Spectral}.
In a general \ac{DMD} algorithm, we assume that data are generated by linear dynamics:
\begin{equation}
    \mathbf{x}^{t+1}=\mathbf{A}\mathbf{x}^{t},
\end{equation}
where an operator $\mathbf{A}$ is assumed to exist and approximate the dynamics. The \ac{DMD} modes and eigenvalues are intended to approximate the eigenvectors and eigenvalues of $\mathbf{A}$.
For \ac{DMD} on the selected variable, the discredited numerical solution is split into two matrices: 

\begin{equation}
\mathbf{X} = \begin{bmatrix}
 \mid & \mid &        & \mid \\
    \mathbf{x}^{1}& \mathbf{x}^{2} & ... & \mathbf{x}^{N_{s}-1}    \\
    \mid & \mid &        & \mid
\end{bmatrix},\ 
\bar{\mathbf{X}}= \begin{bmatrix}
 \mid & \mid &        & \mid \\
    \mathbf{x}^{2}& \mathbf{x}^{3} & ... & \mathbf{x}^{N_{s}}    \\
    \mid & \mid &        & \mid
\end{bmatrix},
\end{equation}
as a set of pairs $\begin{Bmatrix}
\begin{pmatrix}
{\mathbf{X}}^k, {\bar{\mathbf{X}}}^k
\end{pmatrix}
\end{Bmatrix}_{k=1}^{N_s-1}$. The \ac{SVD} of $\mathbf{X}\in  \mathbb{R}^{N_h\times (N_s-1)}$ is computed by Eq.~\ref{SVD_eq} as:
\begin{equation}
    \mathbf{X} = \mathbf{U} \mathbf{\Sigma} \mathbf{V^{\ast}},
\end{equation}
where $\mathbf{U}\in  \mathbb{R}^{N_h \times N_h}$, $\mathbf{V}\in  \mathbb{R}^{(N_s-1)\times (N_s-1)}$, $ \mathbf{\Sigma}\in  \mathbb{R}^{N_h\times (N_s-1)}$. 
A low-order approximation of $\mathbf{X}$ can be made by retaining only the first $\tilde{n}$ singular values:
\begin{equation}
   \mathbf{X} \approx  \tilde{\mathbf{U}}  \tilde{\mathbf{\Sigma}} \tilde{\mathbf{V^{\ast}}},
\end{equation}
where $\tilde{\mathbf{U}}\in  \mathbb{R}^{N_h\times \tilde{n}}$ , $\tilde{\mathbf{V^{\ast}}}\in  \mathbb{R}^{\tilde{n}\times (N_s-1)}$, $ \tilde{\mathbf{\Sigma}} \in  \mathbb{R}^{\tilde{n}\times \tilde{n}}$. 
Then the approximation of the dynamic matrix $\mathbf{A}$ in the low-dimensional space, $\tilde{\mathbf{A}}\in\mathbb{R}^{\tilde{n}\times \tilde{n}} $, is defined as:
\begin{equation}
    \tilde{\mathbf{A}}=\tilde{\mathbf{U}}^{\ast} \bar{\mathbf{X}} \tilde{\mathbf{V}}\tilde{\mathbf{\Sigma}}^{-1}
\end{equation}
The eigenvalues $\Lambda$ and eigenvectors $\boldsymbol \omega$ of $\tilde{A}$ are computed by:
\begin{equation}
    \tilde{\mathbf{A}}\boldsymbol \omega =\boldsymbol \omega \Lambda ,
\end{equation}
where $\Lambda$ is a diagonal matrix with DMD eigenvalues $\lambda _{\tilde{n}}$ then the corresponding DMD mode is given with $\tilde{\mathbf{U}}$:
\begin{equation}
    \Phi  = \tilde{\mathbf{U}} \boldsymbol \omega,
\end{equation}
where $\Phi \in  \mathbb{R}^{N_h\times \tilde{n}}$ contains the eigenvectors or dynamic modes corresponding to the \ac{DMD} eigenvalue $\Lambda$ in the original high-dimensional space. 
The whole DMD process on flow past a cylinder case is shown in Figure \ref{fig: DMD}.

In addition, \ac{ML} tools have been developed and shown great potential in representing complex nonlinear mapping. With strong nonlinear fitting ability, \ac{ML} models are implemented in \ac{ROM}s' modal decomposition and evolution regression parts~\cite{hinton2006reducing, cheng2023machine}.  For the establishment of the appropriate coordinate system in latent space, the deep learning-based \ac{AE} network has been widely used as a nonlinear model reduction alternative to linear methods, like \ac{POD}, capturing features more efficiently~\cite{guo2016deep}. A typical autoencoder is a self-supervised neural network with identical inputs and outputs. It includes an encoder $\mathcal{E}$ to map the inputs $x$ into the reduced latent space and a decoder $\mathcal{D}$ to reconstruct the high-fidelity simulations from the latent representations $\tilde{\mathbf{x}}$:
\begin{equation}
    \tilde{\mathbf{x}} = \mathcal{E}(\mathbf{x})\ \ \textrm{and}\ \  \mathbf{x}^{{AE}} =\mathcal{D}(\tilde{\mathbf{x}}).
\end{equation}
The encoder $\mathcal{E}$ and decoder $\mathcal{D}$  are optimised simultaneously for minimising the reconstruction loss function:
\begin{equation}
     L_{\mathcal{E},\mathcal{D}}(\mathbf{x}) = \left \| \mathbf{x}-\mathcal{D}(\mathcal{E}(\mathbf{x})) \right \|=\left \| \mathbf{x}-\mathbf{x}^{AE} \right \| ,
\end{equation}
where $|| . ||$ is the Euclidean norm.
A series of algorithms based on \ac{AE} have been developed to construct \ac{ROM}s for latent-space learning~\cite{quilodran2021adversarial, phillips2021autoencoder, pu2016variational}. 

However, a limitation of these reduced order methods is their lack of awareness of points position. This means that while they efficiently reduce the dimensionality of data by capturing dominant features, they might overlook spatial dependencies and interactions critical in the mesh structure~\cite{Rowley2017Model, dutta2022reduced}. To address this limitation, additional strategies could be integrated into \ac{ROM}s, such as spatial tagging of modes or coupling with spatially-aware algorithms, to enhance the position sensitivity of these techniques. Meanwhile,  these reduced-order methods inherently require a fixed mesh configuration. They operate under the assumption that the number of mesh points remains constant and that these points are consistently positioned throughout the dataset~\cite{alsayyari2019nonintrusive}. This limitation can be particularly restrictive in fields like fluid dynamics or structural mechanics, where adaptive unstructured meshes are often essential to capture complex behaviours efficiently. The integration of \ac{ROM}s with adaptive mesh strategies is being explored to overcome these challenges, such as Adaptive Mesh Refinement~\cite{baiges2020finite, carlberg2015adaptive} and re-meshing techniques~\cite{pain2001tetrahedral}.

Once an appropriate coordinate system is established, various \ac{ML} algorithms are utilised to model the dynamics evolution for forecasting, especially in Non-Intrusive Reduced Order Modellings~\cite{Xiao2019Reduced}. \ac{RNN}s (see Section \ref{sec:CNNRNN}) have improved the modelling of temporal dependencies in \ac{ROM}s~\cite{wang2020recurrent, maulik2021reduced}, such as \ac{LSTM} networks excel in capturing long-range time dependencies, addressing the vanishing gradient issue~\cite{nakamura2021convolutional, Mohan2018Deep}. Self-attention algorithm-based methods like Transformers and its variants have been explored in \ac{ROM}, facing challenges in computational efficiency~\cite{wu2021reduced,fu2023nonlinear}. Approaches like \ac{SINDy} efficiently derive dynamic system models from data~\cite{brunton2016discovering, rudy2017data, fasel2022ensemble, fukami2021sparse}. Enhanced accuracy in \ac{ROM} predictions is achievable through specialised physical knowledge integration by \ac{PINN} as physical-informed ROMs~\cite{gong2022data, fu2023physics}. While \ac{ML} models combined with \ac{ROM} reduce computational loads in system modelling, the prediction process incrurs deviation from the physical model due to error accumulation over successive predictions.

\subsection{Shallow machine learning for dynamical systems                
             }
\label{sec:shallow}
Conventional \ac{ML} applications are increasingly pivotal in the analysis and understanding of dynamic systems and unstructured mesh environments. 
These systems, characterised by their complexity and ever-changing nature, present unique challenges for computational modelling and simulations. 
Several well-established \ac{ML} algorithms discussed herein have been effectively applied to
address issues pertaining to dynamical systems. \par

The \ac{KNN} algorithm~\cite{fix1989discriminatory,cover1967nearest} is a simple yet effective method for classification and regression. 
It operates on the principle of selecting a predetermined number of neighbors ($k$) to determine the categorisation or value of new data points, measured using metrics such as Euclidean, Manhattan and Minkowski~\cite{chomboon2015empirical}. 
\ac{KNN} finds considerable application in modelling the dynamics of physical systems, where it can analyse spatial or temporal relationships among data points to simulate complex behaviours~\cite{hoang2022projection}.
A notable implementation is the \ac{KNN}-\ac{DMD}~\cite{gao2022reduced}, which utilises \ac{KNN} to select and average the closest $k$ \ac{DMD} solutions (introduced in Section \ref{sec:POD}). 
This is based on the distances between the parameters of interest and other parameters, thereby adapting \ac{DMD} to parametrised problems effectively.\par

\ac{KNN}'s main challenges are high computational demands with large datasets~\cite{gao2022reduced} and difficulties handling chaotic time-series data~\cite{bollt2000model,bollt2017regularized}. Research continues to improve its efficiency and adaptability~\cite{gao2022reduced,bollt2017regularized}.\par

\begin{figure}[!ht]
    \centering
    \begin{subfigure}[b]{0.48\columnwidth}
        \includegraphics[width=\textwidth]{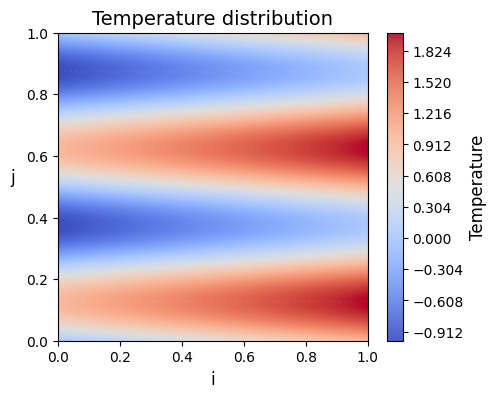}
        \caption{Simulated temperature: $x_{i,j} = \mathsf{i} + \sin(4\pi * \mathsf{j})$}
        \label{fig:T1}
    \end{subfigure}
    \hfill 
    \begin{subfigure}[b]{0.48\columnwidth}
        \includegraphics[width=\textwidth]{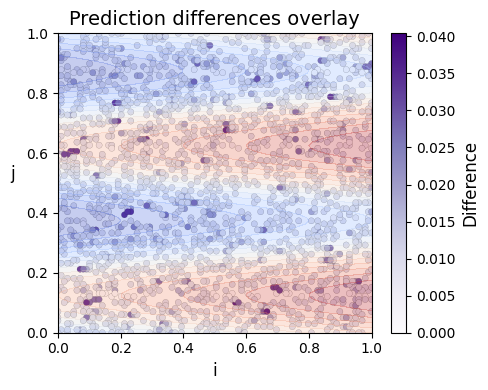}
        \caption{Prediction results: $\text{Difference} = \left| 
        \text{Prediction} - x_{i,j} \right|$}
        \label{fig:T2}
    \end{subfigure}
    \begin{subfigure}[b]{0.8\columnwidth}
        \includegraphics[width=\textwidth]{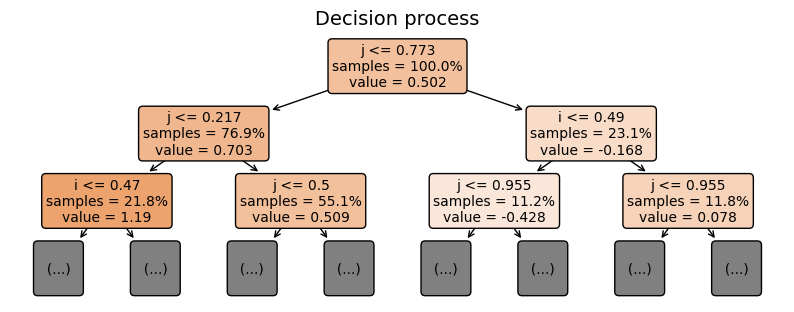}
        \caption{The tree's top-down hierarchy starts with the root node classifying samples. }
        \label{fig:T3}
    \end{subfigure}
    \caption{A regression tree predicts across continuous domains. } 
    \label{fig:DT}
\end{figure}

A decision tree~\cite{von1986decision} is a binary tree model used to handle unstructured data~\cite{wille1996local} and dynamic systems~\cite{freno2019machine}, making it ideal for classification (categorising phenomena) and regression (predicting values, as shown in Figure \ref{fig:DT}). 
The construction of the model, derived from the training data, is a one-time process~\cite{sun2007online}. It enables rapid predictions that remain efficient regardless of subsequent expansions in the size of the dataset ~\cite{wang2011implementation, karabadji2019data}. 
The hierarchical nature of this model (illustrated in Figure \ref{fig:T3}), with its nodes and directed edges, offers clear interpretability~\cite{sagi2021approximating,sagi2020explainable,izza2022tackling}, which is a critical feature in physics that allows for the detailed tracing of the decision-making process~\cite{farhi1998quantum}.  Applications include dynamic assessments and specialised models like Dynamic Fault Trees~\cite{ge2015quantitative} for complex systems. \par

Further advancing on the decision tree's capabilities, the \ac{RF}~\cite{breiman2001random} model emerges as a robust ensemble technique designed to enhance predictive accuracy and prevent overfitting. 
This ensemble method proves particularly potent in dynamic systems, offering enhanced performance and adaptability~\cite{hoang2022projection, hackel2016contour, auret2010change}. 
\ac{XGBoost}~\cite{chen2016xgboost}, on the other hand, marks a significant evolution in the realm of ensemble learning, specifically advancing the use of boosting techniques.  
Unlike \ac{RF} approach, which constructs an ensemble of decision trees in parallel, \ac{XGBoost} builds its tree models sequentially. 
Each tree in the sequence is designed to address and correct the errors of its predecessors, leading to progressively improved model accuracy. 
\ac{XGBoost} has demonstrated exceptional effectiveness in capturing complex relationships within datasets~\cite{kumar2024effect, zhu2023analyzing} and performing reliably across various interpolation scenarios~\cite{yan2023semi}.

\begin{figure}[!ht]
    \centering
    \includegraphics[width=0.7\columnwidth]{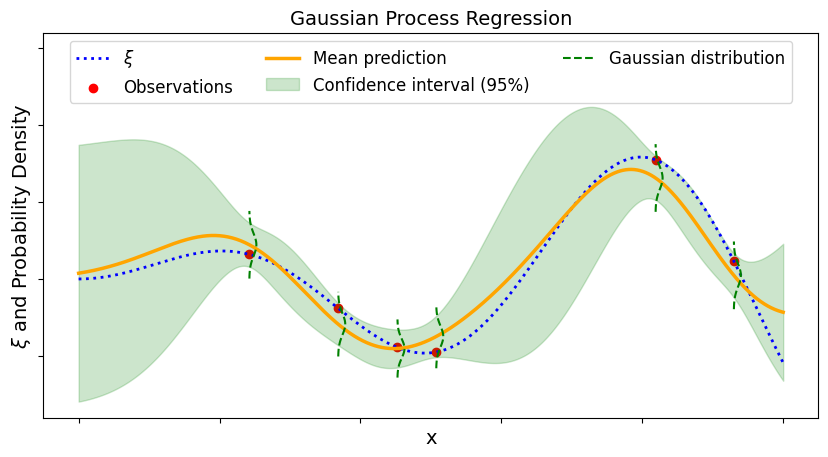}
    \caption{Gaussian process regression (as known as Kriging~\cite{williams1998prediction}) to predict complex, nonlinear relationships.}
    \label{fig:GPR}
\end{figure}

A Gaussian Process~\cite{seeger2004gaussian} is a probabilistic model for continuous domains that represents any set of points as a multivariate Gaussian distribution (Fig.~\ref{fig:GPR}). 
Defined by a mean function $\mu$ and a covariance function $\sigma^2$, GPs model distributions over functions, enabling flexibility in regression and classification for dynamical systems~\cite{casenave2024mmgp}. 
Their ability to handle nonlinear relationships stems from the kernel, which encodes assumptions like smoothness or periodicity. 
By optimizing kernel hyperparameters and employing non-stationary or autoregressive formulations~\cite{wang2005gaussian, damianou2011variational}, GPs adapt to complex, high-dimensional dependencies, such as time-varying noise or evolving dynamics, without rigid parametric constraints. \par

Like \ac{KNN}, \ac{RF}, \ac{XGBoost}, and Gaussian Process models are widely used with reduced-order models to capture nonlinear dynamics in reduced spaces \cite{saraswat2024enhanced,cheng2022parameter,gong2022efficient,kocijan2004gaussian,wang2007gaussian}. While effective for dynamical systems on unstructured grids, tree models are prone to overfitting in high dimensions and are sensitive to noise, while GP models face limitations like the assumption of stationarity and challenges with high-dimensional systems, complicating model construction and inference~\cite{csato2002sparse,nguyen2008local}.

\subsection{Convolutional and recurrent neural networks}
\label{sec:CNNRNN}
Deep neural networks are machine-learning techniques that are capable of handling high-dimensional, nonlinear dynamical systems, and automatically learn complicated patterns from data without extensive manual feature engineering, which are complex to conventional methods as described in Section \ref{sec:shallow}, especially for large data inputs. There are various types of neural networks each designed for a specific task; however, \ac{CNN}~\cite{alzubaidi2021review,li2021survey} and \ac{RNN}~\cite{sherstinsky2020fundamentals} will be briefly discussed in this section regarding their capability of handling spatial-temporal systems. 

\begin{figure}[!ht]
    \centering
    \includegraphics[width=\linewidth]{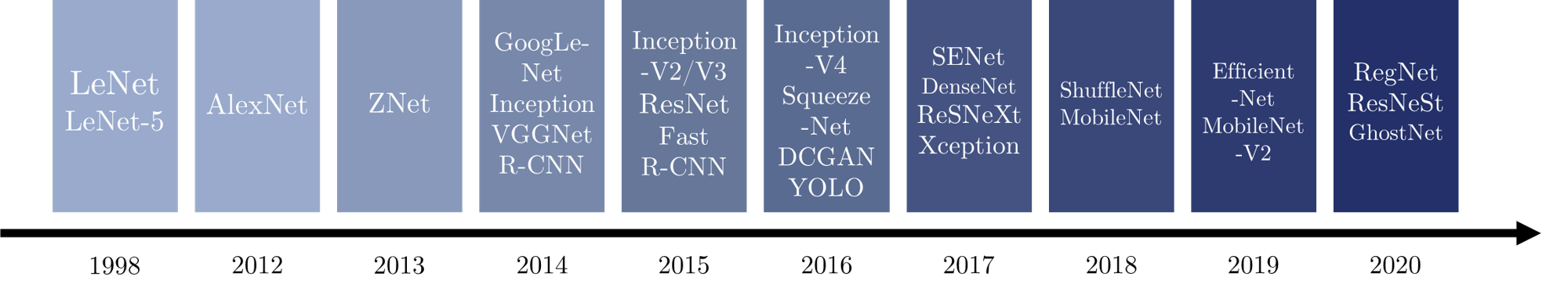}
    \caption{Evolutionary trend of classic CNN models: Visual Geometry Group Network, Regional-Based Convolutional Neural Network, ResNet, Deep Convolutional Generative Adversarial Network, Squeeze-And-Excitation Network.}
    \label{fig:Evolutionary_trend_of_classic_CNN_models}
\end{figure}

\ac{CNN}s are widely used in analysing high-dimensional spatial data such as images~\cite{alzubaidi2021review,li2021survey,dhillon2020convolutional} and dynamical systems like identifying patterns or features in spatially distributed data~\cite{sony2021systematic,bhatnagar2019prediction,jin2023pattern}. In dynamical systems with sensor data inputs, \ac{CNN}s can be leveraged for feature extraction and pattern recognition. For instance, in environmental monitoring systems, \ac{CNN}s can help in identifying anomalies or predicting future states based on sensor readings~\cite{sony2021systematic}. The architecture of \ac{CNN} has undergone a significant transformation, as illustrated in ~\ref{fig:Evolutionary_trend_of_classic_CNN_models}, with 2017 marking a notable surge in the development of \ac{CNN} models. 

\ac{CNN}s were designed originally to process structured data, particularly images~\cite{dhillon2020convolutional} due to their ability to capture hierarchies through convolutional layers. However, through advancements in research and engineering, \ac{CNN}s have been extended to handle unstructured data~\cite{coscia2023continuous,heaney2024applying}, like text, audio, and time-series data. Residual Network (ResNet)~\cite{he2016deep} is a specialised CNN that employs skip connections to learn residual functions, addressing the vanishing gradient problem and enabling the training of deeper networks. Improved Residual Networks (iResNet)~\cite{duta2021improved} further enhance information flow, optimize shortcuts, and improve spatial feature learning, allowing for extremely deep networks without added complexity~\cite{feng2024handle}. Skip connections are also an essential component of transformer type neural networks which could effectively handle irregular data as detailed in Section \ref{sec:transformer}.




The significance of sequential data processing and time series forecasting has grown substantially across various domains, encompassing finance, economics, weather prediction, and natural language processing. Within this landscape, \ac{RNN}s~\cite{kaur2019review,sherstinsky2020fundamentals,beiran2023programming}, notably \ac{LSTM} networks, have risen to prominence as a favorite approach for managing sequential data and predicting time series. The fundamentals of \ac{RNN} are presented in~\cite{sherstinsky2020fundamentals} utilising differential equations encountered in many branches of science and engineering. The canonical formulation of \ac{RNN} by sampling delayed differential equations can be applied in modelling of complex processes in physics. Chen et al~\cite{chen2022pyramid} proposes pyramid convolutional \ac{RNN} for MRI image reconstruction based on low, middle, and high-frequency information in a sequential pyramid order with better recovery of fine details different from \ac{CNN} which learns the three frequency categories individually. The more robust \ac{LSTM}, a highly used type of \ac{RNN} overcomes the challenges of the standard \ac{RNN}, and is capable of representing high-dimensional and complex systems, which are typically difficult to model using conventional machine learning techniques~\cite{sherstinsky2020fundamentals,farooq2023multiscale}. For instance, \ac{RNN} and \ac{LSTM} are used in capturing dynamical systems with multivariate parameters optimisation to minimise errors~\cite{inapakurthi2021deep}. Blandin et al~\cite{blandin2022multi} use a multi-variate \ac{LSTM} to predict rapidly changing geomagnetic conditions electric fields form within the Earth’s surface and induce currents. A Convolutional \ac{LSTM} network provides a fast and affordable prediction of spatio-temporal systems in expensive computation fluid solvers~\cite{adeli2023advanced}.

The main models for handling sequential data use advanced recurrent or convolutional neural network that incorporate an encoder and a decoder~\cite{Vaswani2017}. Complex dynamic systems with multiple time lags are regarded as high-dimensional dynamical systems with time lags and the temporal dependence plays a key role in modelling them. However, in classical \ac{CNN} models, the number of operations needed to relate signals of two arbitrary input or output positions grows in the distance between positions, creating complexities in learning dependencies between distant positions. \ac{RNN}s appear as ideal candidates to model, analyse, and predict complex and dynamical systems due to their temporal occurrence. However, classical \ac{RNN}s do not carry out sequential reasoning, a process based on attention~\cite{Vaswani2017,hernandez2021attention}. Position embedding technique in attention mechanism incorporates information about position of tokens within a sequence that conventional \ac{RNN} and \ac{CNN} are unable to capture. It has the capabilities of handling sequences of variable length without tempering performance and hence is able to handle sparse, unstructured, and missing data. More details about the attention mechanism and the transformer type neural networks are given in Section  \ref{sec:transformer} of this review paper.


\subsection{Gridding irregular data via interpolation }
\begin{figure*}[htb]
\centering
\includegraphics[width=0.60\textwidth]{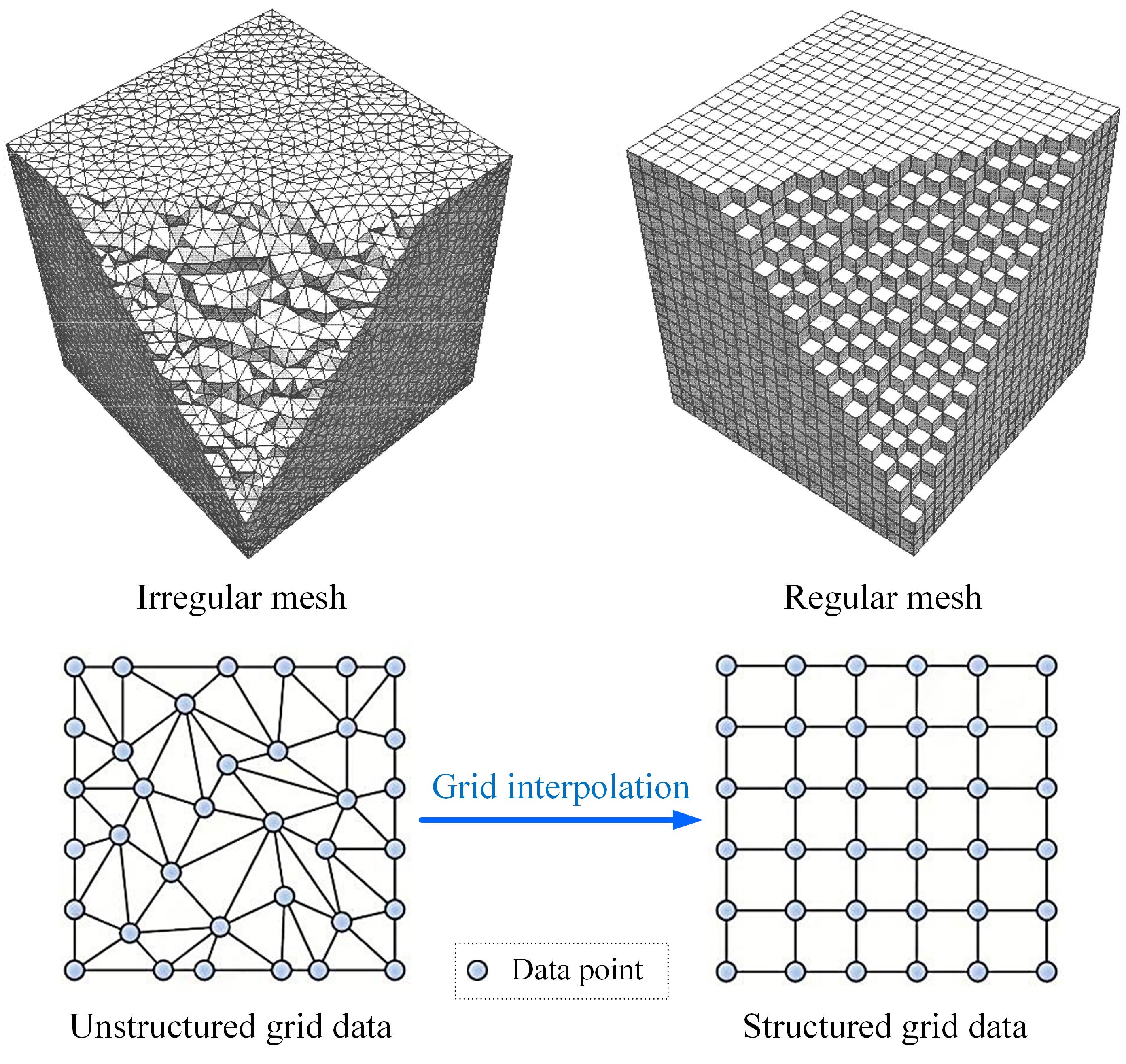}
\caption{Illustration of converting unstructured grid data to structured gird data via interpolation.}
\label{fig:Data_conversion}
\end{figure*}

Discretisation~\cite{bern2000mesh, sevilla2022mesh} constitutes an indispensable step for numerically resolving \acp{PDE} that govern the evolution of dynamical systems. In this process, continuous physical fields in the space-time domain are transformed into discrete representations by meshing the domain into thousands or even millions of small elements, making the problem computationally solvable. 
Generally, meshing can be categorised into regular and irregular types, as depicted in Figure~\ref{fig:Data_conversion}. While regular meshes facilitate computer programming, irregular meshes are far more prevalent for approximating complex, arbitrary geometrical shapes in practical applications.

With the increasing use of computational intelligence across various fields, there is a growing demand to convert large volumes of unstructured data into structured formats that can be processed by machine and deep learning models. This is particularly important because many models, including perceptrons and convolutional neural networks, are usually optimised to work with data arranged in a regular Cartesian grid as mentioned in Section \ref{sec:CNNRNN}. To meet this need, interpolation methods are commonly employed to convert irregularly gridded data into a raster dataset composed of regularly spaced square pixels, enabling further analysis and model training in machine learning applications. Commonly used interpolation methods include nearest neighbor interpolation, linear or barycentric interpolation, radial basis function interpolation, and Kriging interpolation.

\begin{figure*}[htb]
\centering
\includegraphics[width=0.90\textwidth]{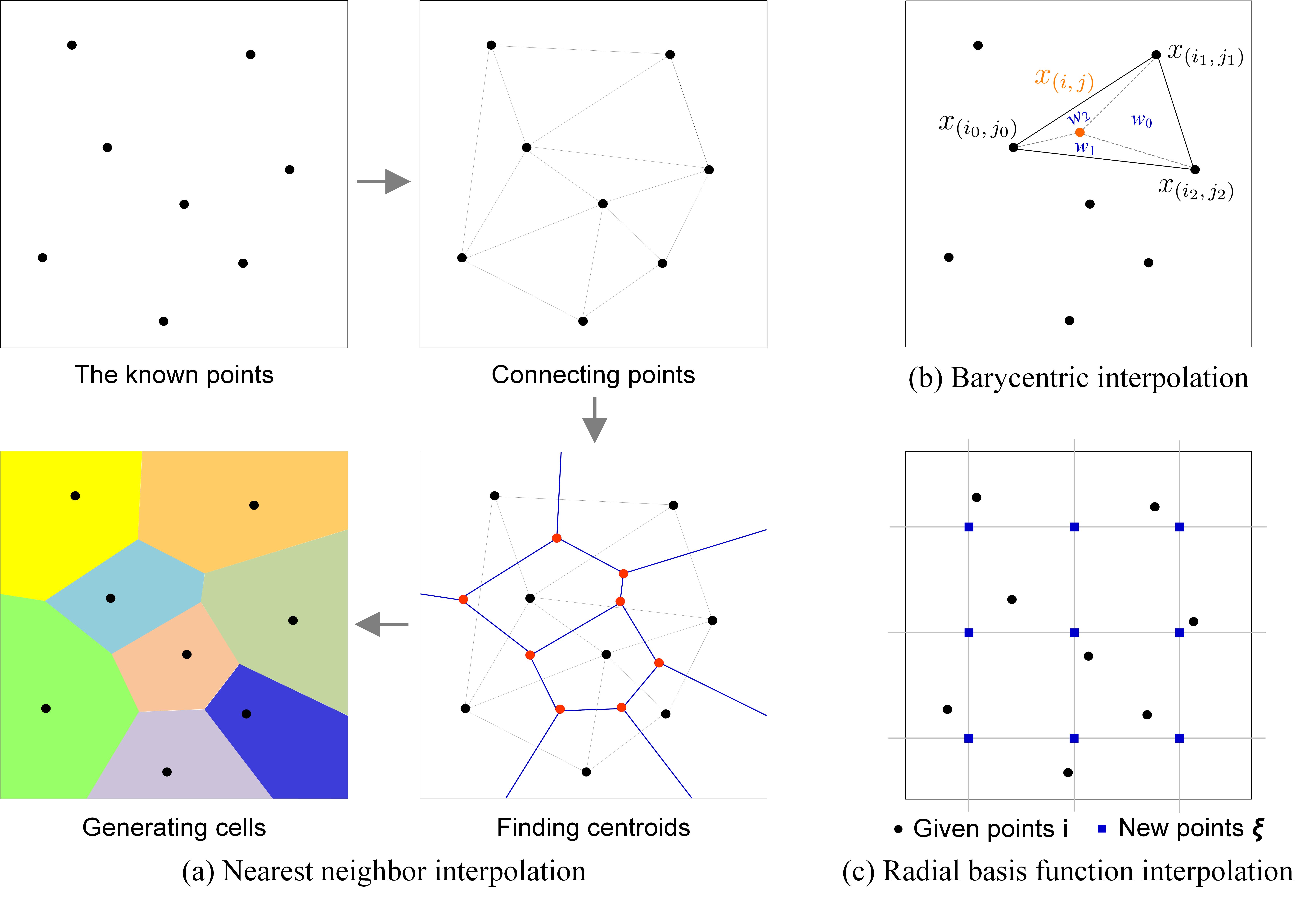}
\caption{Graphical illustration of unstructured grid interpolation.}
\label{fig:grid_interpolation}
\end{figure*}

Nearest neighbor interpolation~\cite{xing2022benefit, cover1967nearest} estimates the value of a variable at any given location by using the value of the nearest known data point. The concept of Voronoi tessellation \cite{aurenhammer2000voronoi} can be employed to measure the distance between points and determine proximity, as illustrated in Figure \ref{fig:grid_interpolation}a. In this method, the entire domain is subdivided into cells based on the locations of the known data points, with each cell assigned the value of its corresponding data point. This means that the value of each known data point is assigned to all unknown data points within its associated cell. While nearest neighbor interpolation is computationally efficient due to its simplicity, it typically results in lower interpolation accuracy, especially when the data points are sparsely distributed or exhibit significant spatial variation.
 
Linear interpolation~\cite{lancaster1981surfaces, de1978practical} creates new data points within the range of known data points by fitting a linear polynomial. For the 1D case, given a field vector $\mathbf{x}$ and the values of two know data points $x_{i_0}$ and $x_{i_1}$, the linear interpolation polynomial $x_{i}$ within the range of $i \in [i_0, i_1]$ is mathematically expressed as:
\begin{equation}
    x_i =\frac{i_1-i}{i_1-i_0}x_{i_0}+\frac{i-i_0}{i_1-i_0}x_{i_1}\,.
\end{equation}
This univariate interpolation can also be straightforwardly extended to multivariate interpolation on 2D and 3D regular grids, called bilinear and trilinear interpolation, respectively.
Barycentric interpolation~\citep{hormann2014barycentric, floater2005surface} generalises linear interpolation for arbitrary (non-regular) grids, allowing transformation of unstructured grid data to structured grid data. For the 2D case, barycentric interpolation creates new data points from three near-neighbours that form a triangle, as depicted in Figure~\ref{fig:grid_interpolation}b. Given the values of three vertices of the triangle $x_{(i_0,\,j_0)}$, $x_{(i_1,\,j_1)}$ and $x_{(i_2,\,j_2)}$, the value of an unknown data point $(i,\,j)$ within this triangle can be estimated via weighted average:
\begin{equation}
    x_{(i,\,j)} = w_0 \hspace{1mm} x_{(i_0,\,j_0)}+w_1 \hspace{1mm} 
 x_{(i_1,\,j_1)}+w_2 \hspace{1mm} 
 x_{(i_2,\,j_2)}\,,
\end{equation}
where $w_0$, $w_1$ and $w_2$ are the barycentric weights, which can be calculated from the coordinates of the three data points, given by:
\begin{equation}
\begin{aligned}
w_0&=\frac{(j_1-j_2)(i-i_2)+(i_2-i_1)(j-j_2)}{(j_1-j_2)(i_0-i_2)+(i_2-i_1)(j_0-j_2)}\,,\\
w_1&=\frac{(j_2-j_0)(i-i_2)+(i_0-i_2)(j-j_2)}{(j_1-j_2)(i_0-i_2)+(i_2-i_1)(j_0-j_2)}\,,\\
w_2&=1-w_0-w_1\,.\\
\end{aligned}
\label{Eq:Data_driven_training}
\end{equation}
These weights, often called barycentric coordinates, are crucial for accurately estimating values at unknown locations based on the positions and values of the known data points.

Radial basis function interpolation \citep{franke1982scattered, buhmann2000radial} represents a sophisticated technique for constructing high-order accurate interpolants from unstructured data. In this method, the interpolant is expressed as a weighted sum of radial basis functions. Given a set of data points consisting of $\big(\mathbf{i}_k, x_{\mathbf{i}_k}\big)$ for $k=1,2,...,n$, where $\mathbf{i}_k$ denotes the vector of coordinates, the goal of radial basis function interpolation is to find an interpolant $s(\mathbf{i}_k)$, satisfying
\begin{equation}
s(\mathbf{i}_k)=x_{\mathbf{i}_k},\ \ \ k=1,2,...,n.
\label{Eq:RBF1}
\end{equation}
For any point $\mathbf{i} \notin \{ \mathbf{i}_k \}_{k=1,2,...,n}$, the interpolant $s(\mathbf{i})$ is expressed as a linear combination of radial basis functions $\phi(\mathbf{i})$:
\begin{equation}
s(\mathbf{i})=\sum_{k=1}^{n}w_k\phi\big(\|\mathbf{i}-\mathbf{i}_k\|\big), \ \ \mathbf{i}\in\mathbb{R}^d\,.
\end{equation}
From the condition in Eq.~\eqref{Eq:RBF1}, the following system of equations for the weights $w_k$:
\begin{equation}
\sum_{k=1}^{n}w_k\phi\big(\|\mathbf{i}_q-\mathbf{i}_k\|\big)=x_{\mathbf{i}_k}, \ \ \ q\neq k,
\end{equation}
where $\mathbf{i}_q$ and $\mathbf{i}_k$ are the vectors of coordinates of different known data points.

By solving the above system of linear equations, the unique solution for the weight vector $\bm{w} = [w_1, w_2...,w_n]$ can be obtained. The unknown value $x_{\bm{\xi}}$ at a point $\bm{\xi}$ on the structured grid can then be estimated as
\begin{equation}
x_{\bm{\xi}}=s(\bm{\xi})=\sum_{k=1}^{n}w_k\phi\big(\|\bm{\xi}-\mathbf{i}_k\|\big), \ \ \bm{\xi}\in\mathbb{R}^d\,,
\end{equation}
as illustrated in Figure \ref{fig:grid_interpolation}c.

Kriging \cite{cressie1990origins, oliver2014tutorial} is a spatial interpolation technique employed to derive predictions at unsampled locations based on observed geostatistical data. Given a set of observation points represented as $(\mathbf{i}_k,\, x_{\mathbf{i}_k})$ for $k=1,2,...,n$, Kriging interpolation estimates the value at an unobserved location $\bm{\xi}$ through a weighted average:
\begin{equation}
\widehat{x}_{\bm{\xi}}=\sum_{k=1}^{n}w_k x_{\mathbf{i}_k}, \ \ \ k=1,2,...,n.
\label{Eq:Kriging}
\end{equation}
This estimation minimises the mean squared prediction error over the entire state space, defined as: 
\begin{equation}
E\big[\big(\widehat{x}_{\bm{\xi}}-x_{\bm{\xi}})^2\big]\,.
\end{equation}
The unknown Kriging weights $\bm{w} = [w_1, w_2...,w_n]$ in Eq. (\ref{Eq:Kriging}) can be derived from the estimated spatial structure of the known dataset. Specifically, these weights are obtained by fitting a variogram model to the observed data, which elucidates how the correlation between observation values varies with distance between locations. 

\begin{figure*}[htb]
\centering
\includegraphics[width=1.0\textwidth]{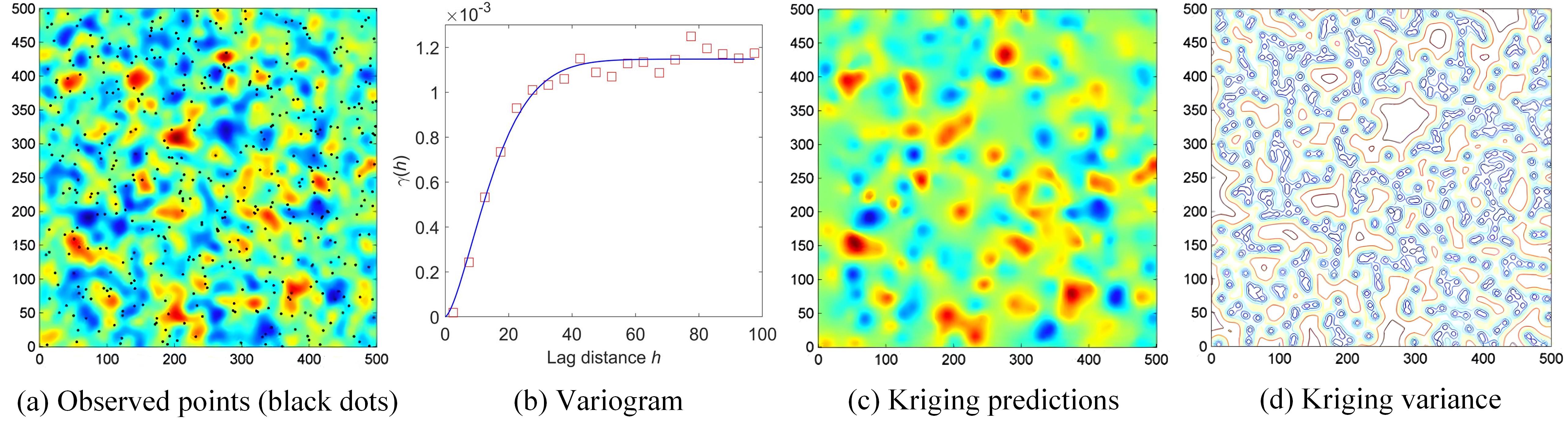}
\caption{Graphical illustration of Kriging interpolation.}
\label{fig:Kriging}
\end{figure*}

Once the Kriging weights are obtained, they are applied to the known data values at observed locations to compute predicted values at unobserved locations, as illustrated in Figure~\ref{fig:Kriging}. These weights reflect the spatial correlation inherent in the data, accounting for both geographical proximity and similarity among data points. Consequently, observed locations that exhibit stronger correlation and are closer to prediction sites receive greater weight compared to those that are uncorrelated and/or more distant. 
Additionally, the weighting scheme considers the spatial arrangement of all observations; thus, clusters of observations in oversampled areas carry less weight due to their lower information content relative to single locations. Under certain assumptions, Kriging predictions serve as best linear unbiased estimators. There exist various types of Kriging methods differentiated by their underlying assumptions and analytical objectives \cite{cressie1990origins, oliver2014tutorial}. For instance, Simple Kriging \cite{omre1989bayesian} presumes that the mean $\mu(\textbf{i})$ of the random field is known; Ordinary Kriging \cite{cressie1988spatial} assumes an unknown but constant mean $\mu(\textbf{i}) = \mu_0$; while Universal Kriging \cite{zimmerman1999experimental} is applicable for datasets characterised by an unknown non-stationary mean structure.

Other interpolation methods, such as those based on polynomial functions, finite element basis functions \cite{farrell2009conservative}, or splines \cite{thomas2022u}, have also been widely adopted for unstructured data. Their combinations with \ac{ML} will be further discussed in Section \ref{sec:preprocessing}.

\section{Machine learning models designed for unstructured grid data}
\label{sec:ML_unstructure}

In this section, we review machine learning models, focusing on neural network structures and specific preprocessing methods that can be applied to dynamical systems with unstructured grid data. These include architectures such as specific \ac{CNN}s, \ac{GNN}s, and transformers.

\subsection{Machine learning with preprocessing}
\label{sec:preprocessing}

\subsubsection{Neural network with interpolation methods}
Traditional \ac{ML} approaches, in particular, \ac{CNN}-based methods are limited to structured grid data and can only process fixed-size input data when an \ac{MLP} layer is included in the neural network architecture. These limitations restrict their applicability to real-world problems, where incomplete or sparse observations and time-varying sensors are common~\cite{barker2004three, elbern2001ozone}. 
Significant effort has been dedicated to integrating machine learning with interpolation methods to address the challenges of data sparsity and irregularity (see review papers~\cite{liu2020gaussian, li2011application}). 

Sparse and continuous \acp{CNN} have been developed to handle sparse and irregular data points in convolutional operations~\cite{liu2015sparse,wang2018deep,xu2021ucnn}. Some pioneering works have applied these advanced \ac{CNN} approaches in \ac{CFD} with unstructured grid data. For instance, Wu et al~\cite{wu2023computationally} use a sparse \ac{CNN} model for end-to-end prediction of supersonic compressible flow fields around airfoils from spatially sparse geometries, while Wen et al~\cite{wen2024hybrid} adapt continuous \acp{CNN} for irregularly placed multiphase flow particles. However, sparse \acp{CNN} involve specialised algorithms for handling sparse data, leading to increased computational overheads and optimisation difficulties. Continuous \acp{CNN}, designed for continuous input spaces, can encounter discretisation errors and require more computational resources for precise computations~\cite{wang2018deep}. Consequently, their application in high-dimensional dynamical systems, such as numerical weather prediction or fine-resolution \ac{CFD}, remains limited.

\begin{figure*}[!ht]
\centering
\includegraphics[width=0.63\textwidth]{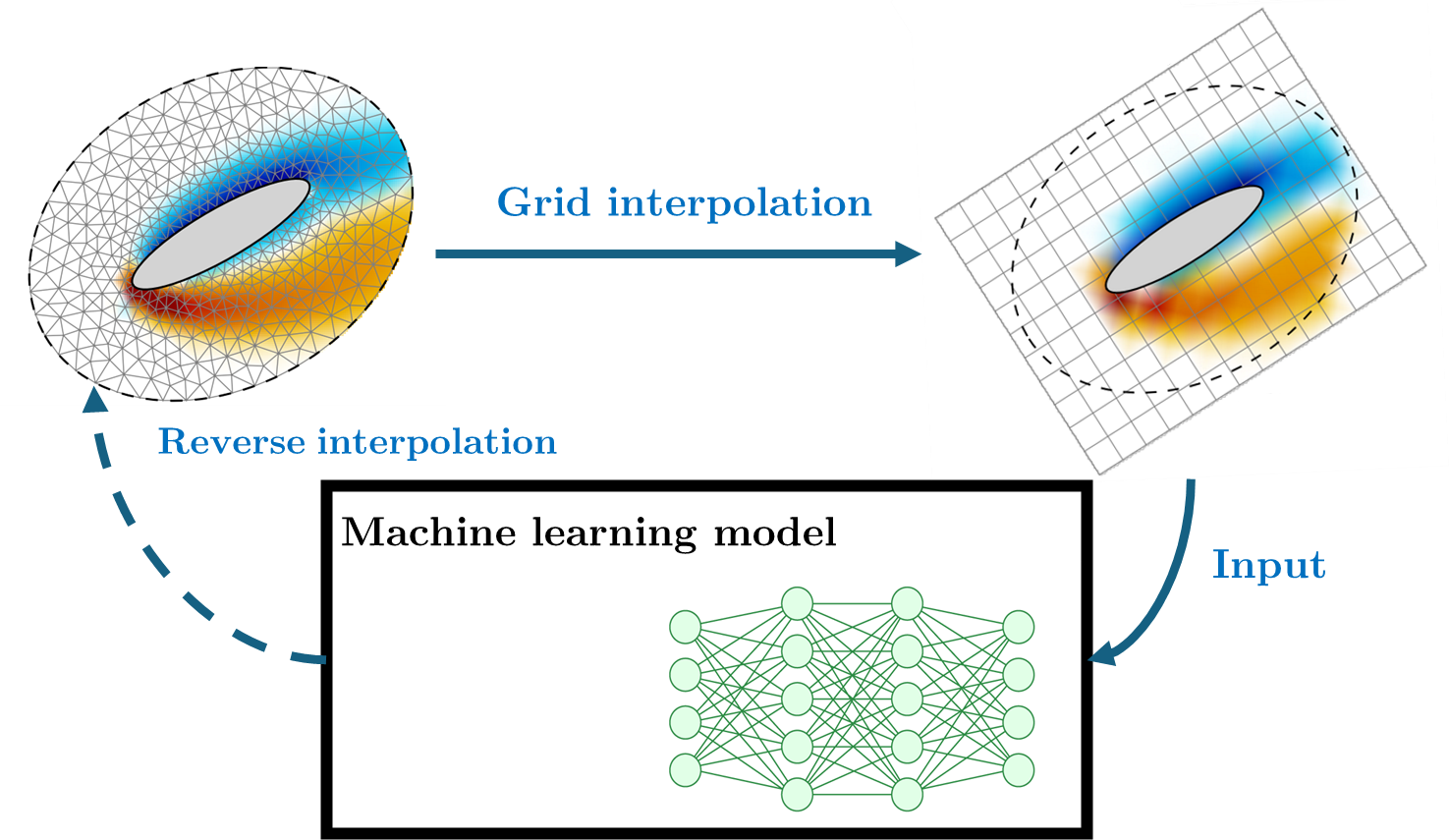}
\caption{Illustration of \ac{ML} methods with interpolated inputs for unstructured data. The figure is based on elements in Figure~9 of \cite{palha2015hybrid}}
\label{fig:int_ML}
\end{figure*}

Another common strategy to address the challenge of unstructured data involves converting unstructured grid data into structured grid data, often resizing it to a fixed dimension before feeding it into a machine learning model, as shown in Figure~\ref{fig:int_ML}. Typically, Hou et al~\cite{hou2023spatial} and Cao et al~\cite{cao2024field} combined Kriging interpolation with \ac{CNN} and \ac{RNN} for spatially irregular input data.
Following a similar idea, the recent work by Fukami et al.~\cite{fukami2021global} leverages Voronoi tessellation, as introduced by~\cite{watson1981computing}, to overcome the challenges posed by sparse observations and the varying number of sensors in CNN-based field reconstruction.
Their method considers a set of observable points (sensors) at a given time, located at $\{(i_{k},j_{k})\}_{k \in \{1,...,k*\}} $ where
\begin{align}
\{i_{k},j_{k}\} \in [1,...,N_x] \times [1,...,M_x] \quad \textrm{and}\quad \textrm{$k*$ is the number of sensors}. 
\end{align}
Suppose $x_{k}$ is the observed value at $\{i_{k},j_{k}\}$.
 A Voronoi cell $R_{k}$ associated to the observation $\{x_{k}, i_{t,k},j_{t,k}\}$ can be defined as
\begin{align}
    R_{k} = \big\{ \{i_r,j_r\}  \hspace{2mm} | \hspace{2mm} d((i_r,j_r), (i_{k},j_{k})) \leq d((i_r,j_r), (i_{q},j_{q})), \hspace{2mm} \forall \hspace{2mm} 1 \leq q \leq k^* \hspace{2mm} \textrm{and} \hspace{2mm} q \neq k \big\}.
\end{align}

Here, $d(\cdot)$ is the Euclidean distance. Therefore, the state space can be partitioned into several Voronoi cells regardless of the number of sensors, that is,
\begin{align}
    \mathcal{U}_{[1,...,N_x] \times [1,...,M_x] } = \bigcup_{k=1}^{k*} R_{k} \quad \textrm{and} \quad R_{k} \cap R_{q} = \emptyset \quad (\forall  \quad k \neq q),
\end{align}
where $\mathcal{U}_{[1,...,N_x] \times [1,...,M_x] }$ designates the full discretised space.
 A tessellated observation $\hat{\bx} = \{ \hat{x}_{i_x,j_x}\} \in \mathbb{R}^{N_x  \times M_x}$ in the full state space can be obtained by
\begin{align}
    \hat{x}_{i_x,j_x} = x_{k} \quad \textrm{if} \quad (i_x,j_x) \in R_{k}.
\end{align}

Once the tessellated observation is obtained, regression \ac{ML} models could be implemented to perform field reconstruction, prediction of future time steps or parameter calibration.

 The Vornoi tessellation-assisted \ac{CNN} has demonstrated superior accuracy and efficiency in \ac{CFD} and geoscience datasets compared to traditional interpolation techniques like Kriging \cite{fukami2021sparse,wang2024dynamical}. 
The recent work by~\cite{cheng2024efficient} integrates Voronoi tessellation-assisted \ac{CNN} into a variational data assimilation framework, enhancing prediction accuracy using the same unstructured and sparse observation data.

On the other hand, the work of~\cite{mohammadpour2024machine} incorporates Kriging interpolation into the loss function calculation for unstructured data. This approach differs from previously mentioned methods by explicitly accounting for interpolation errors during the training phase. A differentiable nearest neighbour algorithm has also been proposed by~\cite{plotz2018neural}. Although it has primarily been tested on image classification and restoration tasks, this method shows promise for enhancing interpolation in unstructured grid systems.

\ac{ML} techniques have also been developed to perform interpolation and super-resolution directly on sparse or unstructured grid data \cite{fukami2023super}. Typical examples in computational physics include airfoil wake and porous flow~\cite{obiols2021surfnet, zhou2022neural}.
Furthermore, Kashefi et al.~\cite{kashefi2021point} leverages PointNet architecture~\cite{qi2017pointnet} to predict flow fields in irregular domains by treating \ac{CFD} grid vertices as point clouds, preserving the accuracy of unstructured meshes without data interpolation. The approach accurately represents object geometry, maintains boundary smoothness, and predicts flow fields much faster than conventional \ac{CFD} solvers, while generalising well to unseen geometries. Compared to conventional interpolation methods such as Kriging or nearest neighbor, which require strong mathematical assumptions, \ac{ML}-based approaches offer more flexible and non-parametric interpolations for irregular data~\cite{li2022application}.

\subsubsection{Machine learning with mesh reordering and transformation}
\label{sec:reordering}

As mentioned in Section \ref{sec:POD}, 
conventional linear projection methods for \ac{ROM}, such as \ac{POD} and \ac{DMD}, can be seamlessly applied to unstructured data by flattening the entire physics field into a single column vector. This approach simplifies the integration of machine learning models for tasks like field prediction or parameter identification. 
For instance, Xiao et al~\cite{Xiao2019Reduced} and Casenave et al~\cite{casenave2024mmgp} apply \ac{POD} to unstructured grid data of dynamical systems, followed by a Gaussian process to model the dynamics of air pollution within a reduced-order space. Similar approaches have also been applied to nuclear reactor physics \cite{kang2022application} and unsteady flow simulations \cite{xu2023comparative}.

\begin{figure*}[!ht]
\centering
\subfloat[Reordering using Cuthill-McKee algorithm]{\includegraphics[width=0.55\textwidth]{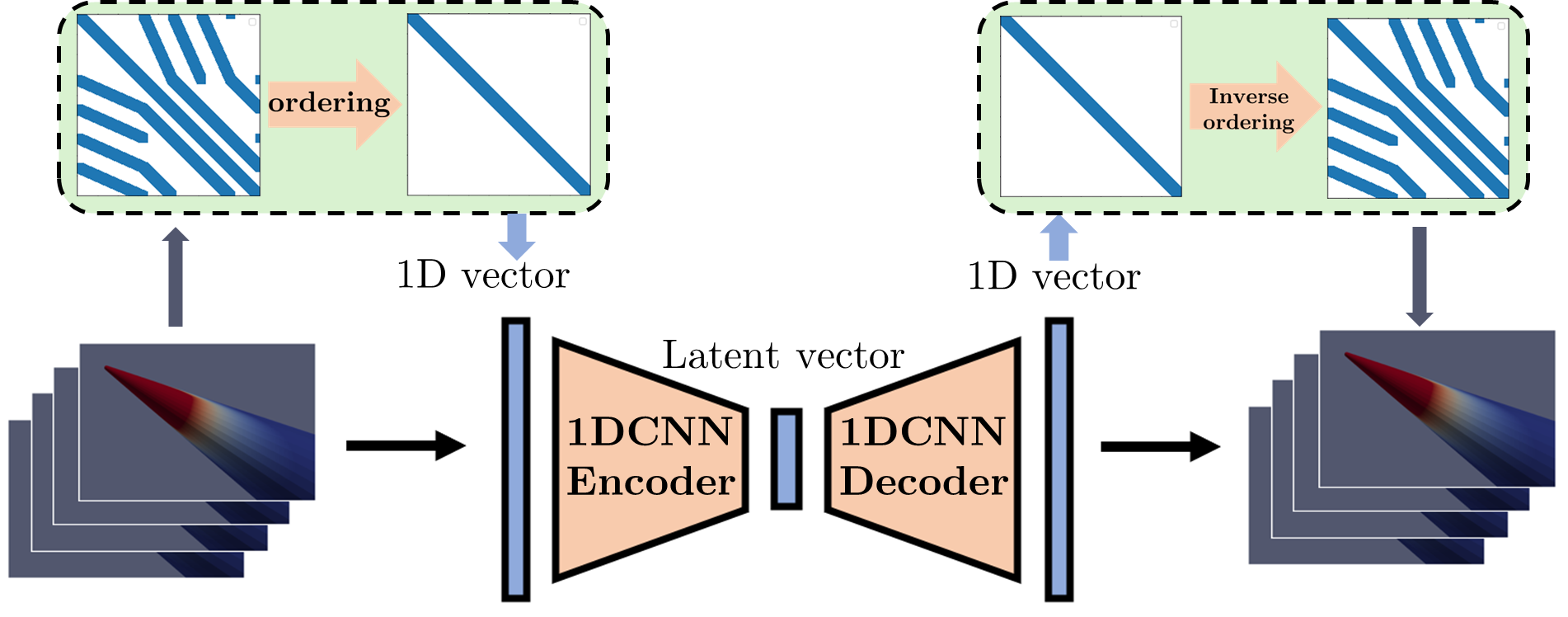}}
\subfloat[Illustration of a space filling curve]{\includegraphics[width=0.45\textwidth]{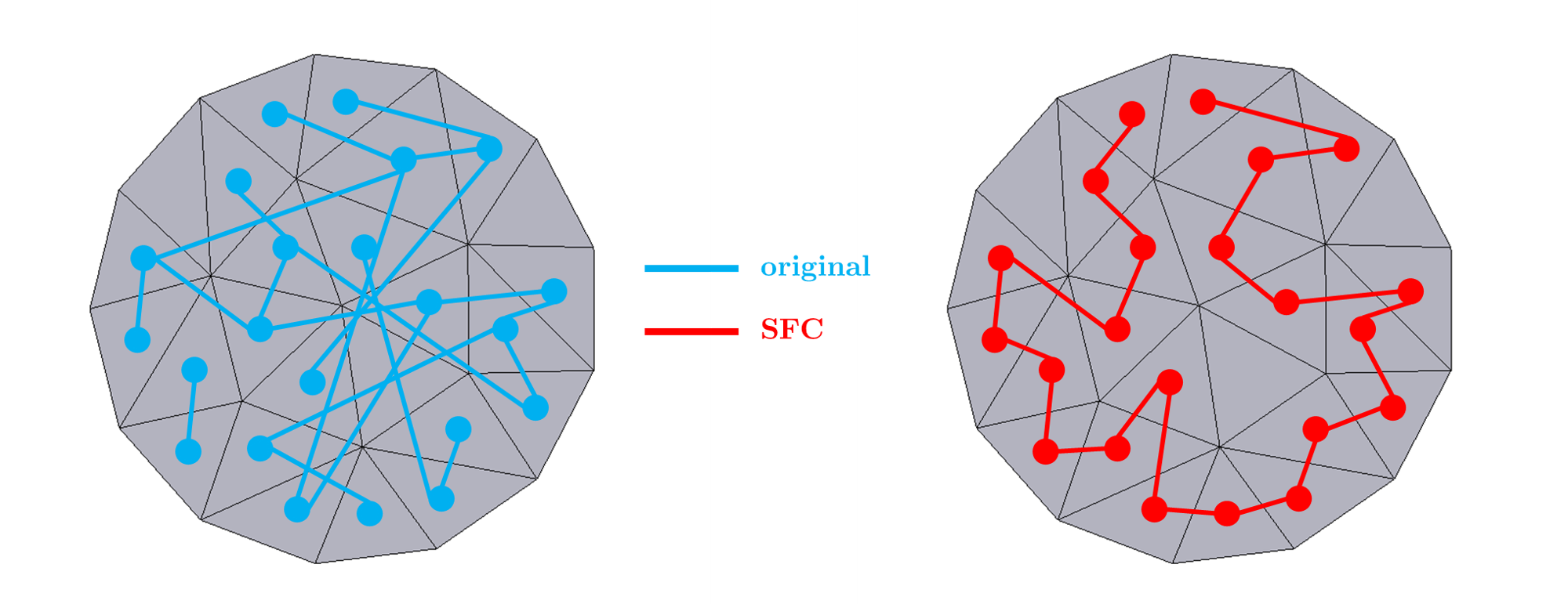}}
\caption{1D CNN-based autoencoder with grid data reordering}
\label{fig:CNN_reorder}
\end{figure*}

Although \ac{POD}-type methods can be directly applied to unstructured grid data, they are less efficient than deep learning techniques, particularly autoencoders, for handling significanlty nonlinear dynamical systems. Building on the concepts of \ac{POD} and \ac{DMD}, researchers have explored the use of autoencoders with one-dimensional \acp{CNN} by first flattening the physical field into a single column vector.
Since the one-dimensional \acp{CNN} uses a fixed size filter to capture the pattern and correlation in input vectors, the ordering of unstructured grids plays a key role in the performance of such approaches~\cite{qian2023soft,heaney2024applying}.  
In \cite{cheng2023generalised}, the authors utilise the Cuthill-McKee algorithm~\cite{cuthill1969reducing}, which draws on concepts from graph theory, to reorder grid points in a way that minimises the bandwidth of the adjacency matrix as shown in Figure~\ref{fig:CNN_reorder}(a). As a result, the reordering of grids into the same convolutional window enhances the numerical accuracy of the autoencoder~\cite{cheng2023generalised}. 
In a related approach, the recent study by~\cite{heaney2024applying} employs space-filling curves to determine an ordering of nodes or cells that converts multi-dimensional data on unstructured meshes into a one-dimensional format, as illustrated by Figure~\ref{fig:CNN_reorder}(b). This transformation allows the optimal application of 1D convolutional layers to the unstructured data.
Such reordering method with space-filling curves has also been applied to regular image data~\cite{wang2022neural}, where it achieves a superior performance compared to conventional \ac{CNN} approaches. 

In line with the principle of grid reordering, coordinate transformation can be employed to adapt standard machine learning models for non-homogeneous mesh structures. For instance, PhyGeoNet~\cite{gao2021phygeonet} utilises a coordinate transformation that maps an irregular domain onto a structured mesh space, enabling the application of convolutional operations on flow fields for fluid flow regression.
Along similar lines, Chen et al.~\cite{chen2020developing} select the cell lengths (both horizontal and vertical) and maximum included angle to form a three-channel quality feature, analogous to the RGB three-channel feature of images, to perform \ac{CNN} for mesh quality evaluation. 
Another family of approaches involve transforming data into the spectral domain to address the challenges posed by complex geometries~\cite{lemeunier2022representation,zhou2024machine,lingsch2023beyond}. In these methods, the spatial coordinates of the mesh cells are typically converted into frequency coefficients within the spectral domain before being input into a machine learning model~\cite{lemeunier2022representation}.
In the frequency domain, these methods apply a global convolution operation, which is a key differentiator from traditional \acp{CNN}. Instead of using local convolutions that only capture local interactions, spectral domain machine learning~\cite{li2020fourier} can capture global patterns across the global input space.

There has also been an effort to merge the flexibility of linear projection techniques like \ac{POD} with the ability of neural networks to manage nonlinear patterns. One common approach, known as \ac{SVD}-\ac{AE} or \ac{POD}-\ac{AE}, involves applying \ac{ML}-based autoencoding to the modal coefficients derived from \ac{SVD}-based methods~\cite{phillips2021autoencoder,casas2020urban,fetni2023capabilities,pham2022pca}. As a result, the machine learning model can effectively manage data with complex geometries, though it requires a fixed input dimension. More precisely, both the input and ouput of the \ac{AE} (with encoder $\mathcal{E}$ and decoder $\mathcal{D}$) in \ac{SVD}-\ac{AE} are the compressed vectors $\tilde{\bx}_{\textrm{SVD}}$ obtained through \ac{SVD}, i.e.,
\begin{align}
    \tilde{\bx}_{\textrm{SVD}} = {\bU} \bx, \quad \tilde{\bx} =  \mathcal{E}(\tilde{\bx}_{\textrm{SVD}}) \quad \textrm{while} \quad  \tilde{\bx}^r_{\textrm{SVD}} =  \mathcal{D} (\tilde{\bx}), \quad \bx^r_\textrm{SVD AE} = {\bU}^T \hspace{1mm} \tilde{\bx}^r_{\textrm{SVD}},
\end{align}
where $\tilde{\bx}^r_{\textrm{SVD}}$ and $\bx^r_\textrm{SVD AE}$ denote the reconstruction of the \ac{POD} or \ac{DMD} coefficients and the reconstruction of the full physical field, respectively. 

\begin{figure*}[!ht]
\centering
\includegraphics[width=0.9\textwidth]{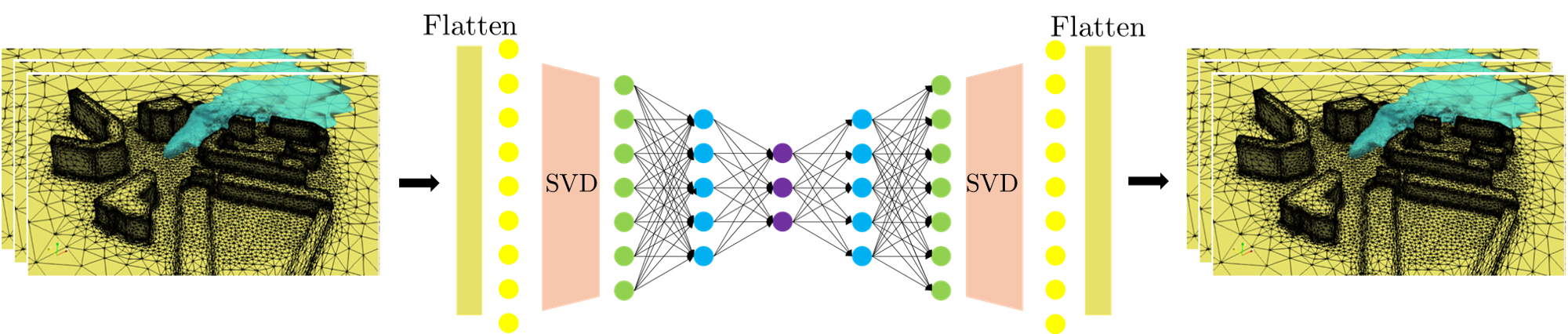}
\caption{\ac{SVD}-\ac{AE} for simulated air pollution data with unstructured meshes}
\label{fig:SVDAE}
\end{figure*}

In fact, numerous studies have investigated the equivalence between \ac{POD} and linear autoencoder networks~\cite{plaut2018principal,bourlard1988auto}, leading to the possibility of jointly training these two-stage data processing methods.
The use of \ac{SVD}-\ac{AE} approaches extends to various applications, including nuclear reactor physics~\cite{phillips2021autoencoder}, multiphase flow \ac{CFD}~\cite{cheng2023generalised} and street-level air pollution estimation~\cite{casas2020urban}, both of which involve unstructured grid data. The latter is illustrated in Figure~\ref{fig:SVDAE}.

\subsection{Graph neural networks}
\ac{GNN}s are specialised in learning structured data on graphs, where nodes represent interconnected data points and edges denote the relationships between them. \ac{GNN}s are particularly suitable for modelling unstructured meshes due to their inherent compatibility with such data structures. Unstructured meshes can be directly represented as graphs, with mesh vertices as nodes and connections formed by mesh edges. This allows for more efficient utilisation of training data by eliminating the need to interpolate unstructured meshes onto uniform grids, which is, as discussed in Section~\ref{sec:preprocessing}, a process typically required when using \ac{CNN}s designed for grid-like datasets such as images. It should also be noted that multiple means exist for constructing a graph from an unstructured mesh. Alternatives such as defining elements or cells as nodes and creating edges based on adjacent elements or cells sharing a face, or incorporating nearest neighbour methods can be equally valid. The selection among these methods depends on the specific requirements of the modelling problem.

In the context of physical simulations, a mesh graph can be formulated as $\mathcal{G} = (\mathcal{V}, \mathcal{E})$, where $\mathcal{E} \subseteq \mathcal{V} \times \mathcal{V}$ denotes a set of edges defining connections between pairs of nodes whose set is $\mathcal{V}$. Node-level features can be represented by $\mathbf{V}^G \in \mathbb{R}^{|\mathcal{V}| \times N^{v}_{f}}$, where $N^{v}_{f}$ is the number of node-level features, which typically includes spatial locations and physical (flow) field parameters measured at the corresponding points in the physical domain. The mesh graph connectivity can be described with the adjacency matrix $\mathbf{A}^G \in \mathbb{Z} ^ {|\mathcal{V}| \times |\mathcal{V}|}$, where an entry $a_{uv}$ is set to one if there exists an edge from node $u$ to node $v$, and zero otherwise. $\mathbf{E}^G \in \mathbb{R}^{|\mathcal{E}| \times N^{e}_{f}}$ represent edge-level features, with $ N^{e}_{f}$ indicating the number of edge-level features per edge. Moreover, it can also be beneficial to define a graph-level attribute $\mathbf{u} \in \mathbb{R}^{N^{g}_f}$ that contains global information pertinent to the entire graph, such as simulation parameters and inflow conditions, which are universally applicable to all nodes and edges. The final input to the \ac{GNN} can be represented as the tuple $\mathbf{G} = (\mathbf{V}^G, \mathbf{E}^G, \mathbf{u}, \mathbf{A}^G)$. A node-level regression task can then be setup to use the input graph processed by layers of GNNs to predict various physical quantities at each node, such as pressure, velocity, or temperature, depending on the specific application.

While \ac{CNN}s excel at capturing local patterns and maintaining translational invariance in images with convolutional filter of fixed sizes, \ac{GNN}s can better adapt to the irregular connectivity structures found in unstructured meshes and can accomodate to graphs of diverse sizes and structures. \ac{GNN}s function by processing and integrating information from neighbouring nodes within a graph structure.
The 1-hop neighbourhood of node $u$ can be defined as $\mathcal{N}_u = \{v | (u, v) \in \mathcal{E} \textrm{ or } (v, u) \in \mathcal{E}\}$, with the neighbourhood's features represented as the multiset $\mathbf{X}_{\mathcal{N}_u} =\{\{ \mathbf{x}_v: v \in \mathcal{N}_u \}\}$ where $\mathbf{x}_v$ represents the feature vector of node $v$. \ac{GNN}s aggregate features over local neighborhoods within the graph by applying shared, permutation invariant functions $\phi(\mathbf{x}_u, \mathbf{X}_{\mathcal{N}_u})$.
 Most graph neural networks can generally be categorised into one of the three flavours~\cite{bronstein2021gdl, velickovic2023gnn}: convolutional~\cite{kipf2017gcn, hamilton2017graphsage, defferrard2016chebconv, wu2019sgconv}, attentional~\cite{velickovic2018gat, brody2022gatv2} and message-passing~\cite{gilmer2017mpnn, battaglia2018graphnet}. Each flavour determines how the features of neighouring nodes are processed and aggregated, offering different levels of complexity for capturing the interactions across the graph. The output feature representation for a node $u$, denoted $\mathbf{h}_u$, varies across different \ac{GNN} flavours as follows:

\begin{align}
    &\text{Convolutional}: 
    \mathbf{h}_u = \phi\bigr(\mathbf{x}_u, \underset{v \in \mathcal{N}_{u}}{\bigoplus} c_{uv} \psi(\mathbf{x}_v)\bigr), \\
    &\text{Attentional}:
    \mathbf{h}_u = \phi\bigr(\mathbf{x}_u, \underset{v \in \mathcal{N}_{u}}{\bigoplus} a(\mathbf{x}_u, \mathbf{x}_v) \psi(\mathbf{x}_v)\bigr), \\
    &\text{Message-passing}:
    \mathbf{h}_u = \phi\bigr(\mathbf{x}_u, \underset{v \in \mathcal{N}_{u}}{\bigoplus}  \psi(\mathbf{x}_u, \mathbf{x}_v)\bigr),
\end{align}
where $\phi$ and $\psi$ are trainable neural networks, $\bigoplus$ is a permutation-invariant aggregation function such as mean, sum or maximum. In the convolutional flavour, the importance of a neighbouring node $v$ on node $u$'s feature representation is quantified by a constant $c_{uv}$, which is a direct function of the graph's structural connectivities. The attentional flavour, on the other hand, computes this influence through a trainable self-attention mechanism $a$ (see Section \ref{sec:transformer} for details). In the message-passing flavour, $\psi$ is a trainable function that can compute and convey arbitrary vectors or messages from node $v$ to $u$. While the message-passing flavour offers more expressive and flexible modelling by computing vector-valued messages, they tend to be more memory-intensive and difficult to train. Attentional \ac{GNN}s provide a more scalable middle ground by passing scalar-valued messages across edges, and convolutional \ac{GNN}s are most suited for the efficient processing of homophilous graphs.

\begin{figure*}[!h]
\centering
\includegraphics[width=0.85\textwidth]{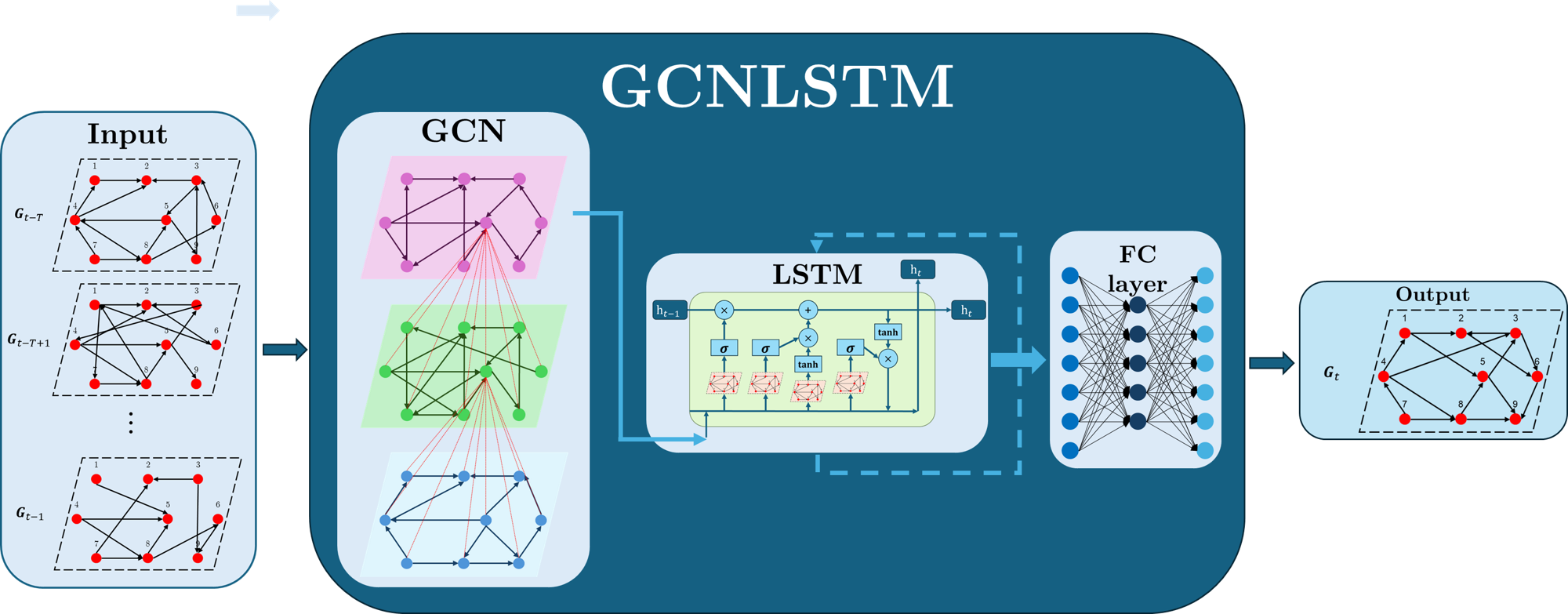}
\caption{An example schematic for combining GCNs and LSTMs for modelling spatial-temporal unstrcutured mesh data.}
\label{fig:GCNLSTM}
\end{figure*}

Integrating \ac{GNN}s with \ac{LSTM} networks is an effective strategy for modelling dynamic fluid flow problems. This method leverages the \ac{GNN}s' compatibility with unstructured meshes and their capability to capture spatial dependencies and relationships inherent in fluid flow data. Simultaneously, \ac{LSTM}s excel at managing sequential data and temporal dependencies. An illustrative schematic of this approach is shown in Figure~\ref{fig:GCNLSTM}.

In recent years, \ac{GNN}s of these different flavours have been widely applied to modelling complex computational physics problems with unstructured meshes. For instance, the work of~\cite{ogoke2021graph} trained convolutional graph networks to accurately predict laminar flows around airfoils of different shapes. The work of~\cite{belbuteperes2020gcncfd} combined convolutional graph networks and a differentiable \ac{CFD} solver that operates at a significantly coarser resolution to develop a surrogate model that can better generalise to previously unseen flow conditions. He et al~\cite{he2022flow} developed graph neural networks that can be trained to reconstruct missing information from partial or incomplete flow fields and accurately predict fluid dynamics from sparse inputs. Li et al~\cite{li2022graph} applied \ac{GNN}s to accelerate molecular dynamics simulations, where \ac{GNN}s are used to learn interactions between particles and predict the forces between atoms in molecular systems. Liu et al~\cite{liu2021multi} developed UNet inspired multi-resolution \ac{GNN}s that uses a hiercrchical graph structure to capture different levels of resolution of the data in order to allow for efficient and accurate approximations of \ac{PDE} solutions. Chen te al~\cite{chen2021graph} trained Convolutional Graph Networks on laminar flow around various two-dimensional shapes and the developed surrogate model showed promising results in predicting flow fields and aerodynamic properties including drag and lift. Suk et al~\cite{suk2024gcn} developed SE(3)-equivariant Convolutional Graph Network that is inherently invariant to translations and equivairant to rotations of the unstructured mesh, to accurately predict hemodynamic fields on high-resolution surfaces meshes of artery walls. Li et al.~\cite{li2023gnn} trained convolutional and attentional \ac{GNN}s on 3D \ac{RANS} data to accurately predict flow fields around wind turbines and power generation. Sanchez-Gonzalez et al~\cite{sanchezgonzalez2020meshgnn} developed graph network based simulators using message passing \ac{GNN}s that can learn to simulate a wide range of challenging particle-based physical problems, including smooth particle hydrodynamics, rigid bodies and interacting deformable materials. By injecting strong inductive biases into the models, representing physical states with graphs of interacting particles and approximating physical dynamics by learned message-passing among nodes, the models were able to generalise to much larger graphs and longer time scales than those encountered during training. The work of~\cite{pfaff2020learning} introduced MeshGraphNets which is a framework for learning mesh-based simulations of diverse physical problems, including aerodynamics, structural mechanics and cloth dynamics, all using message passing \ac{GNN}s. In particular, the developed models are able to adaptively adjust the mesh discretisation during simulation and supports the learning of resolution-independent dynamics and scale to more complex discretisations at test time. Song et al.~\cite{song2022m2n} proposed a \ac{GNN} based mesh deformer for mesh movement-based mesh adaptation, which accelerates PDE solving.  Barwey et al~\cite{barwey2023multiscalegnn} developed a multi-scale message-passing \ac{GNN} autoencoder with learnable coarsening operations that can generate interpretable latent graphs that reveal regions important for flowfield reconstruction.

\subsection{Transformer and attention mechanism}
\label{sec:transformer}
\subsubsection{Transformers: attention-mechanism}
The advent of transformer models~\cite{Vaswani2017}, originally developed within the domain of \ac{NLP}, has offered novel methodologies for addressing the challenges of sparse and irregular data. Characterised by their self-attention mechanisms, transformers have demonstrated remarkable ability to manage unstructured data and interpret data dependencies over long ranges, making them particularly suited for the intricacies involved in machine learning of computational physics.

\begin{figure}[!ht]
\centering 
\includegraphics[width=0.7\textwidth]{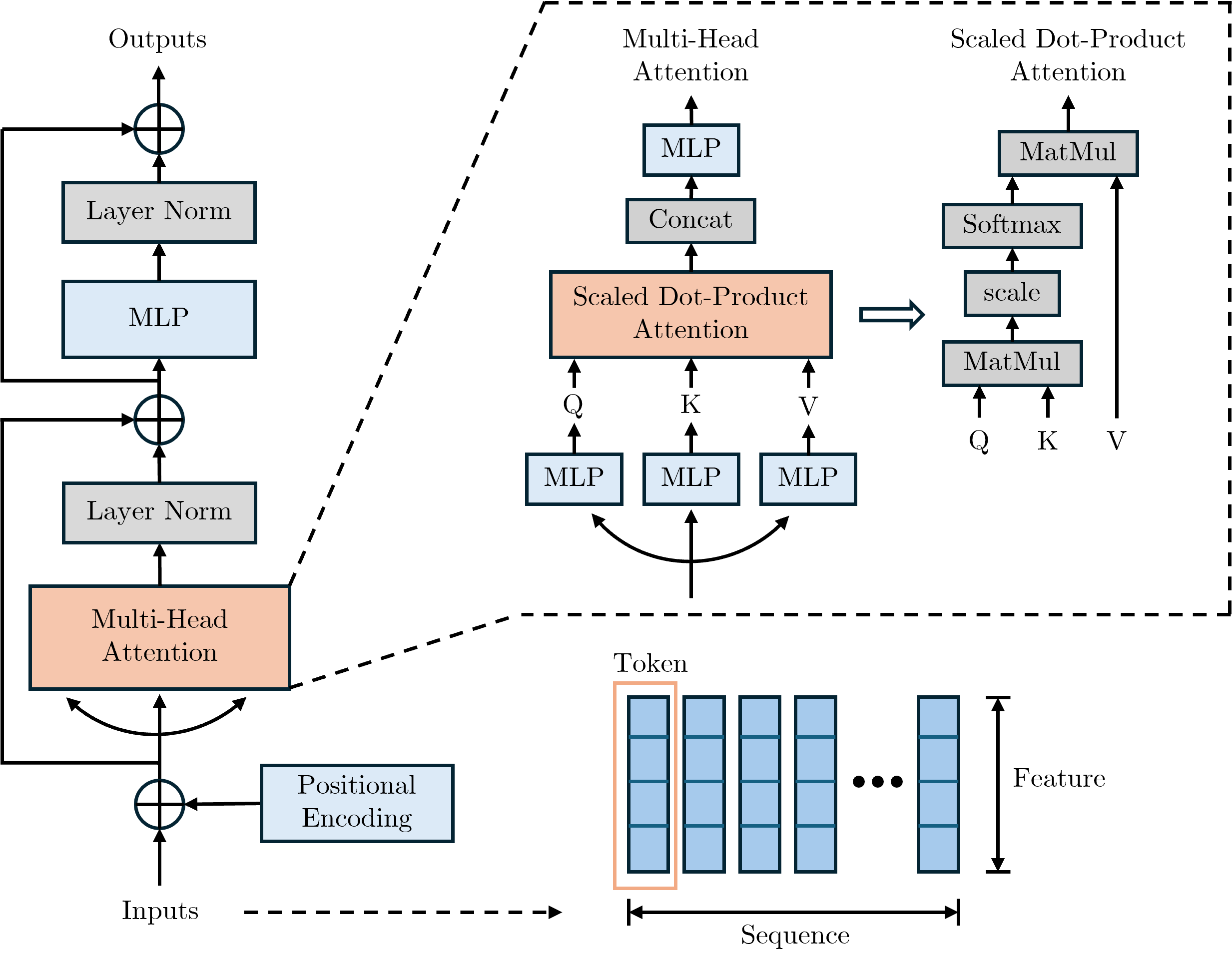} 
\caption{Transformer architecture. For brevity, only the encoder part of the original transformer is shown in the figure as the decoder part has similar architecture. } 
\label{fig:transformer_arch} 
\end{figure}

The overview of the transformer model architecture is shown in Figure~\ref{fig:transformer_arch}. Note that for brevity, only the encoder part of the original transformer is depicted here. The decoder part has similar architecture and is designed for machine translation tasks, which is not relevant to many applications in computational physics. For readers who are interested in the decoder, please refer to the seminal paper on transformers~\cite{Vaswani2017}.
Given an input sequence $\mathbf{x} \in \mathbb{R}^{n \times d}$, where $n$ and $d$ are the number of tokens and feature dimension of each token. In the context of language processing, an input sequence is a sentence and a token is the word embedding (i.e., encoded feature vector) of a word from the sentence. In computational physics, an input sequence can be a list consisting of all the nodes of an unstructured mesh and a token is the feature vector of a node. In the following part, we will introduce self-attention, the key component of transformers, and positional encoding mechanism in details.

\paragraph{Self-attention}
The self-attention mechanism enables a model to selectively focus on different parts of a sequence when processing each token of that sequence. It achieves this by generating three vectors for each input token: a query vector ($\mathbf{Q}$), a key vector ($\mathbf{K}$), and a value vector ($\mathbf{V}$), derived from the input through learned linear transformations (the \ac{MLP}s shown in Figure~\ref{fig:transformer_arch}). The essence of self-attention lies in computing attention scores by taking the scaled dot product of a query vector with all key vectors, including itself, which are then used to create a weighted sum of value vectors. This process allows the model to weigh the significance of all other tokens when encoding each token, thereby encapsulating the context of each token within the sequence. To obtain vectors  query $\mathbf{Q} \in \mathbb{R}^{n \times d_q}$, key $\mathbf{K} \in \mathbb{R}^{n \times d_k}$ and value $\mathbf{V} \in \mathbb{R}^{n \times d_v}$, we define three learnable projection matrices $\mathbf{W}^Q \in \mathbb{R}^{d \times d_q}$, $\mathbf{W}^K \in \mathbb{R}^{d \times d_k}$, $\mathbf{W}^V \in \mathbb{R}^{d \times d_v}$ and apply them to the input as linear transformations:
\begin{equation}
\mathbf{Q} = \mathbf{x} \mathbf{W}^Q, \quad \mathbf{K} = \mathbf{x} \mathbf{W}^K, \quad \mathbf{V} = \mathbf{x} \mathbf{W}^V.
\end{equation}

Here we consider computing the simple scaled dot product attention.
The attention scores (i.e., the measure of similarity between two vectors) are computed by taking the dot product of the query vector $\mathbf{Q}$ with all key vectors $\mathbf{K}$, followed by a scaling factor to stabilise gradients during training. The scaling factor is usually the square root of the dimension of the key vectors $\sqrt{d_k}$. To ensure that the attention scores sum up to $1$ across the sequence for each query, a softmax function is applied after the attention score computation:
\begin{equation}
\text{Attention}(\mathbf{Q}, \mathbf{K}, \mathbf{V}) = \text{softmax}\left(\frac{\mathbf{Q}\mathbf{K}^T}{\sqrt{d_k}}\right)\mathbf{V} .
\end{equation}
To enhance the ability to capture diverse feature and increase representation capacity, multi-head attention mechanism~\cite{cordonnier2020multi} is usually applied instead of relying on a single head. When computing the multi-head attention, we project the input into queries, keys and values multiple times with different projection matrices  $\mathbf{W}^Q_i \in \mathbb{R}^{d \times d_q}$, $\mathbf{W}^K_i \in \mathbb{R}^{d \times d_k}$, $\mathbf{W}^V_i \in \mathbb{R}^{d \times d_v}$. Given $m$ heads, the attention of each head $\mathbf{h}_i$ is computed in parallel. All computed attentions $\{\mathbf{h}_i\}_i^m$ are concatenated together and projected with a projection matrix $\mathbf{W}^O \in \mathbb{R}^{md_v \times d}$:
\begin{equation}
\begin{aligned}
\text{MultiHeadAttention}(\mathbf{x}) = \text{Concat}(\mathbf{h}_1, \mathbf{h}_2, ..., \mathbf{h}_m)\mathbf{W}^O, \\ 
\text{where} \, \mathbf{h}_i = \text{Attention}\left(\mathbf{Q}\mathbf{W}^Q_i, \mathbf{K}\mathbf{W}^K_i, \mathbf{V}\mathbf{W}^V_i\right),
\end{aligned}
\end{equation}
where $m$ is the number of heads of MultiHeadAttention, $d_v$ is the features dimension of the output after concatenating multihead attentions. The computed attentions have the same size to value vectors $V$ and each vector in the attention is a weighted summation of each vector in value vectors $V$. Intuitively, each vector in the attentions is an encoded vector containing aggregated local and global information from all other vectors. The amount of the information aggregated from each vector to the encoded vector is determined by their similarity i.e., weights or attention scores. 

\paragraph{Positional encoding}
Since the self-attention mechanism does not inherently process the sequential order of the input, transformers use positional encoding to incorporate information about the position of tokens in the sequence. This allows the model to understand the order of tokens, which is crucial for tasks in language processing.

In the original transformer paper~\cite{Vaswani2017}, sine and cosine functions of different frequencies are proposed for positional encoding: 
\begin{subequations}
\begin{align}
\mathbf{p}_{pos, 2\iota} &= \sin\left(\frac{pos}{10000^{2\iota/d}}\right), \\
\mathbf{p}_{pos, 2\iota+1} &= \cos\left(\frac{pos}{10000^{2\iota/d}}\right),
\end{align}
\end{subequations}
where the $pos$ is the position of a token in a sequence and $\mathbf{p}_{pos, 2\iota}$ is the $2\iota$-th element of the d-dimensional token $\mathbf{p}_{pos}$ (i.e., $0 \leq 2\iota \leq d$). The positional encoding is added to input sequence $\mathbf{x}$ element-wisely for injecting position information explicitly for each token. Therefore, the query, key and value with positional encoding are extended as:
\begin{equation}
\mathbf{Q} = (\mathbf{x} + \mathbf{p})\mathbf{W}^Q, \quad \mathbf{K} = (\mathbf{x} + \mathbf{p})\mathbf{W}^K, \quad \mathbf{V} = (\mathbf{x} + \mathbf{p})\mathbf{W}^V.
\end{equation}

The positional encoding introduced above is designed for language processing. There are different positional encoding methods tailored for graph data structures, e.g.,~\cite{li2022transformer,xu2024self}. For more discussions about positional encoding methods, which are out of scope of this review, readers can refer to~\cite{Ladislav2022} for more details.

\subsubsection{Applications in computational physics}
\begin{figure}[!htb]
\centering 
\includegraphics[width=0.85\textwidth]{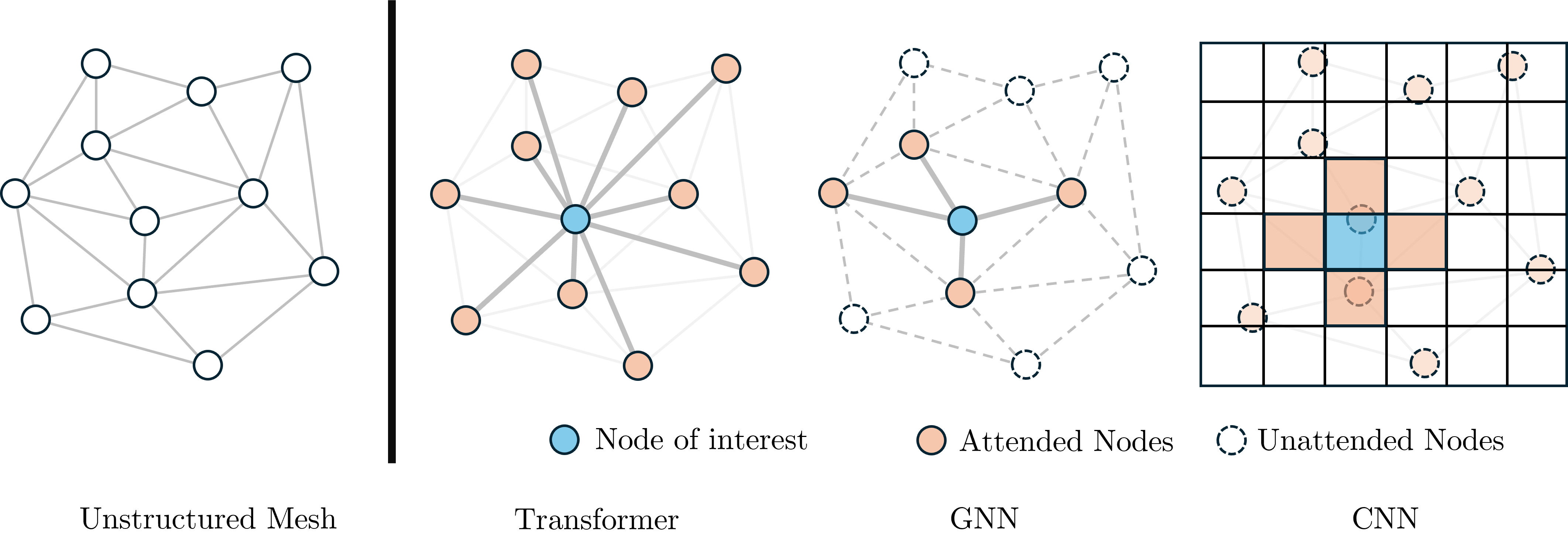}
    \caption{Comparison between models on processing unstructured mesh data. A transformer model employs a self-attention mechanism to consider all mesh nodes. Single layer \ac{GNN}s attend their direct neighbors. \ac{CNN}s apply fixed size kernel on a structure grid with interpolated data from original unstructured mesh.} 
\label{fig:comparision} 
\end{figure}

Although transformers are originally proposed for language processing tasks, their strong modelling abilities and capacities have drawn increasing research interest in machine learning for areas among computational physics, for examples, Hamiltonian dynamics~\cite{Hutchinson2022}, particle physics~\cite{qu2022}, computational fluid dynamics~\cite{geneva2022, LI2024122758}, weather and climate forecasting~\cite{gao2022earthformer}. Transformers are combined with \ac{GNN} to capture long-term temporal dependencies through its attention mechanism for long sequence prediction in physical simulations~\cite{han2022}, or capturing long-distance spatial information for mesh adaptation ~\cite{zhang2024um2n}. The self-attention mechanism underpins the transformers' ability to aggregate long-range (or global) information and naturally to handle unstructured data. The explicit positional encoding enables transformers to understand and incorporate the order or position of tokens (or nodes in unstructured meshes). These features present advantages comparing to \ac{CNN}s and \ac{GNN}s in handling spatial data. A visualisation of the comparison between different models on processing unstructured mesh data is shown in Figure~\ref{fig:comparision}. In addition to using transformers as pure data-driven model for physical system modelling, there is a special interest in building neural \ac{PDE} solvers using transformers especially in neural operator learning~\cite{Kovachki2023}. Intuitively, the self-attention mechanism can be interpreted as a learnable Galerkin projection~\cite{Cao2021transformer} or a learnable kernel integral~\cite{kissas2022, nguyen2022} in neural operator learning methods. To handle both uniform and non-uniform discretisation grids, OFormer~\cite{li2022transformer} is proposed based on an encoder-decoder architecture with self-attention and cross-attention. General Neural Operator Transformer~\cite{hao2023gnot} proposes a heterogeneous normalised attention layer to flexibly handle unstructured meshes, multiple and multi-scale input functions. 
The quadratic-complexity of standard scaled-dot product attention computation limits its applications on large scale problems. To improve the efficiency and scalability, linear attention is investigated for \ac{PDE} modelling~\cite{Cao2021transformer}. FactFormer~\cite{li2023scalable}, on the other hand, proposes a computationally efficient low-rank surrogate for the full attention based on an axial factorised kernel integral. Large-scale pre-training has emerged as a potential approach demonstrating robust generalisation capabilities across a range of downstream tasks in both natural language processing~\cite{brown2020} and computer vision~\cite{he2022} domains. There have been initial attempts to explore pre-training for \ac{PDE}s~\cite{McCabe2023, Subramanian2023, Mialon2024, hao2024DPOT}, in which transformers serve as backbones. 

There has also been significant work applying transformers to 3D irregular meshes, addressing tasks such as mesh recovery and 3D mesh generation~\cite{chen20223d, yoshiyasu2023deformable, siddiqui2023meshgpt,chen2022transformer,hong2024,zhang2024clay}. A key aspect of these approaches is the use of spatial attention mechanisms, which allow the models to effectively capture and utilise the spatial relationships within the mesh data. For instance, in the context of mesh recovery, spatial attention enables the transformers to focus on crucial areas of the mesh, leading to more accurate reconstructions. The Deformable Mesh Transformer, as explored in recent work~\cite{chen2022transformer}, leverages these attention mechanisms to improve the accuracy of human mesh recovery by dynamically adjusting the attention across different mesh regions, effectively handling complex deformations. Additionally, transformer-based models have shown promise in generating 3D meshes by focusing on the intricate spatial dependencies within the data~\cite{yoshiyasu2023deformable}, which is crucial for creating detailed and coherent structures. These advances highlight the growing role of spatial attention in enhancing the performance of transformers on 3D mesh tasks.

In summary, transformers have shown promising abilities in modelling complex physical systems of computational physics. Compared to \ac{CNN}s and \ac{GNN}s, transformers can naturally handle unstructured meshes as well as capture long-range dependencies using the self-attention mechanism. Furthermore, leveraging their scalability and parallelisation capabilities, transformers are extensively utilised in large-scale pre-training models. They serve as foundational backbones for constructing models that facilitate the learning of \ac{PDE}s in the computational physics area. 

\subsection{Summary and comparison }

It is clear that the three families of approaches introduced in this section are widely adopted in the computational science community to handle spatially unstructured and irregular data. However, each of them may exhibit certain advantages and disadvantages depending on the application field. Therefore, we believe a qualitative comparison would be beneficial to highlight the strengths of each family of approaches.

\ac{CNN}s capture local patterns efficiently by applying the same filter across different parts of the input, leveraging translational invariance. 
Conventional \ac{CNN}s struggle with irregular data such as unstructured meshes, requiring interpolation onto regular grids, which introduces potential errors.
An alternative solution could be 1D \ac{CNN}s, particularly with mesh-dependent reordering as introduced in Section \ref{sec:reordering}. Although 1D \ac{CNN}s can seamlessly handle unstructured grid data, their unidimensionality constrains their ability to capture relative and global position awareness within the neural network.

\ac{GNN}s are designed to work with graph-structured data, where the concept of position is not as straightforward as in grid-like data structures. Instead of relying on position, \ac{GNN}s focus on the structure and relationships between nodes. They are aware of the node's position in terms of its connectivity and neighborhood in the graph. Therefore, \ac{GNN}s excel at capturing the relational invariance and dependencies between elements but struggle to understand the absolute or relative positions which are important in computational physics. In addition, as \ac{GNN}s deepen, they can suffer from over-smoothing~\cite{li2018deeper}, where node features become too homogenised, reducing model performance in tasks requiring fine distinctions. 

Transformers are the most explicitly position-aware, designed to incorporate position information directly into their processing, which makes them highly effective for a wide range of tasks that require an understanding of element order or position within the data. An important strength of the attention mechanism is position awareness, i.e., the ability to understand and incorporate the order or position of elements. In the context of machine learning in computational physics, the "position" indicates both the relative and absolute spatial positions of elements in data structures such as meshes. On the other hand, transformers, as the foundation of many 'large' AI models, require considerably more data to train compared to \acp{CNN} and \acp{GNN}, a phenomenon often referred to as 'data-hungry' \cite{pandey2024vision}.
\begin{table}[ht]
\begin{center}

\caption{Summary of neural network and linear projection methods (\checkmark: correct; \ding{53}: incorrect; $\bigcirc$: partially correct)}%
\label{table:LA}
\resizebox{\textwidth}{!}{%
\begin{tabular}{cccccccc}
\hline
\hline
Methods  &  \makecell{POD/DMD}  & \makecell{MLP} & \makecell{1DCNN} &  \makecell{2DCNN*} & \makecell{GNN} & \makecell{Transformer} & \makecell{PINNs}\\
\hline
\hline
Global information   & \checkmark   & \checkmark  & \ding{53} & \ding{53}   & \ding{53}  & \checkmark  & \ding{53}\\
Grid information   & \ding{53}   & \ding{53}   & \ding{53} & \ding{53}  & \checkmark   & \checkmark  & $\bigcirc$\\
Adaptive mesh & \ding{53}   & \ding{53}   & $\bigcirc$ & $\bigcirc$  & \checkmark   & \checkmark  & \checkmark\\
Physics information   & \ding{53}   & \ding{53}  & \ding{53} & \ding{53}   & \ding{53}  & \ding{53}  & \checkmark\\
Position awareness & \ding{53}   & \ding{53}   & $\bigcirc$ & \checkmark  & \ding{53}   & \checkmark  & \checkmark\\
\hline
\hline
\end{tabular}
}
\end{center}
\end{table}

A summary of abilities comparison between single layer CNNs, GNNs and transformers is shown in Table~\ref{table:LA}. The column 2DCNN* indicates 2D \ac{CNN} methods with grid reordering, interpolation or transformation.

\section{Learning paradigms with unstructured data}
\label{sec:paradigms}

Machine learning performance on unstructured data is influenced not only by neural network structures but also by training objectives, including loss function design and data augmentation strategies. This section reviews Physics-Informed Neural Networks with meshless loss functions, reinforcement learning for mesh generation, and generative AI approaches for handling unstructured grid data. These workflows are largely independent of the neural network structure, ensuring adaptability to those introduced in Section \ref{sec:ML_unstructure}.

\subsection{PINNs: a meshless solution}
\label{sec:PINNS}

\acp{PINN} mark a significant evolution in scientific machine learning, by embedding the physical laws described by the differential equations in the neural network~\cite{raissi2019physics}. Traditional numerical methods, while foundational in engineering, struggle with challenges such as mesh dependency and computational burdens in high dimensions. Early neural network applications were limited to purely data-driven techniques but recent advances in automatic differentiation have revitalised this approach, with \ac{PINN}s at the forefront, offering solutions to forward and inverse problems by enforcing physical laws within the learning process~\cite{lee_neural_1990r, lagaris_artificial_1998}.

\ac{PINN}s excel in parametrising high-dimensional \ac{PDE}, incorporating parameters directly into training data to navigate expansive parameter spaces efficiently, which is a significant advantage over conventional numerical techniques~\cite{schiassi_extreme_2020}. This capability is further enhanced by Physics-Informed Deep Operator Networks, which learns an operator learning architecture to solve parametric problems~\cite{wang_learning_2021}. 

\begin{figure}[!ht]
    \centering
    \includegraphics[width=\linewidth]{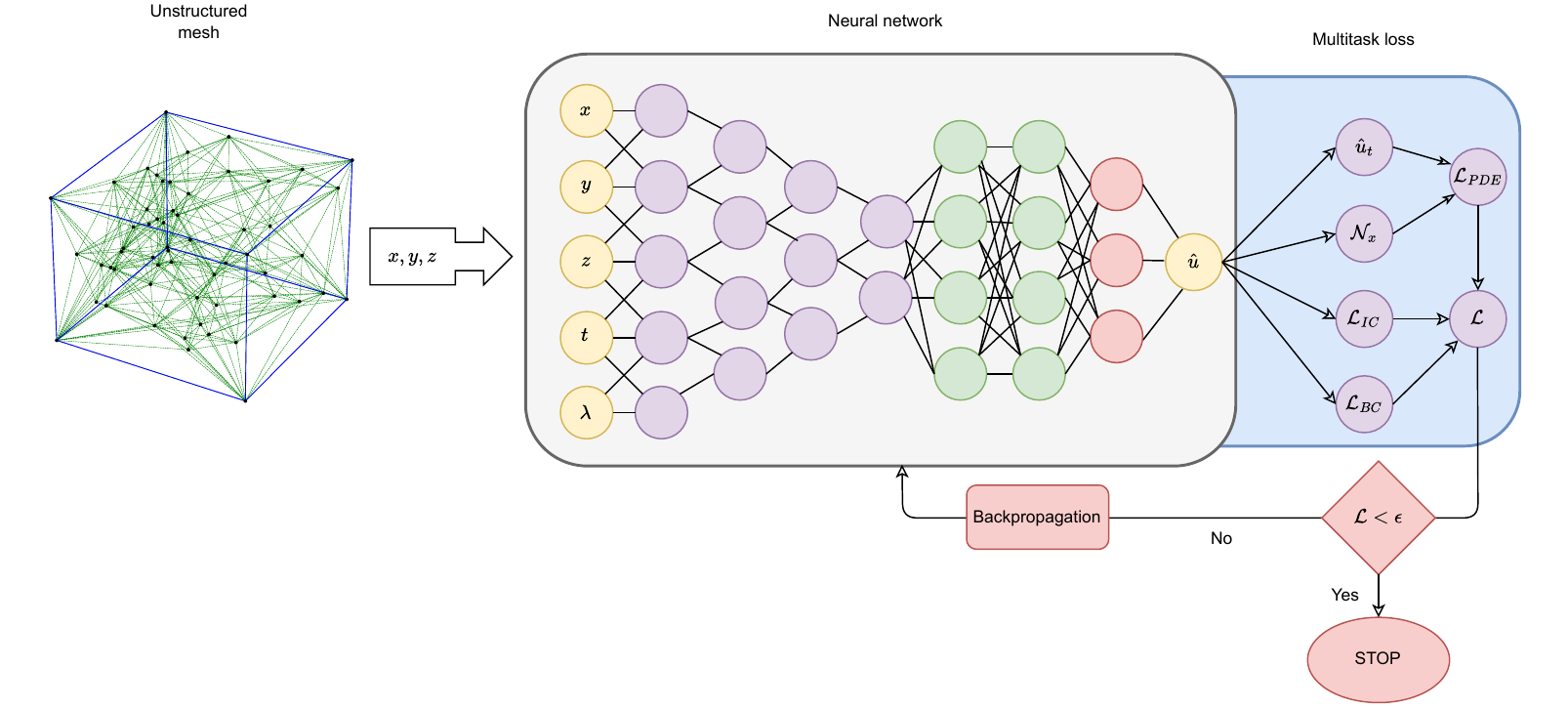}
    \caption{Schematic of a parametric \ac{PINN} for a 3D time dependent problem.}
    \label{fig:PINN_schematic}
\end{figure}

Figure~\ref{fig:PINN_schematic} shows the schematic of a parametric \ac{PINN} for a 3D time dependent problem. The inputs to the neural network are the spatio-remporal coordinates $x,y,z,t$ along with a set of parameters $\lambda$. Although one can obtain these spatial coordinates from the nodal points of unstructured meshes, employing sophisticated pseudo-random sampling techniques--such as Latin hypercube sampling, Sobol sequences~\cite{sun2021comparing}, Halton sequences~\cite{faure2009generalized}, or Hammersley sequences~\cite{wong1997sampling}--offers a more efficient alternative. These techniques generate unstructured point clouds that require significantly fewer points for effective domain sampling, compared to using the nodal points from unstructured meshes directly.

\ac{PINN}s inherently manage unstructured data more efficiently than conventional architectures like \ac{CNN}s and \ac{RNN}s due to their ability to integrate differential equations directly into the learning process. \ac{CNN}s, primarily designed for grid-like data, such as images, often struggle with irregular, non-gridded data formats without extensive pre-processing to regularise the input space. \ac{RNN}s, while adept at handling sequences, similarly face challenges with spatial irregularities and multidimensional data. In contrast, \ac{PINN}s leverage the underlying physical laws, represented through differential equations, to learn the solution to the \ac{PDE} in the domain. This capacity allows \ac{PINN}s to learn complex patterns with fewer data points~\cite{zheng2023physics,fang2021high}.


The output from the \ac{PINN} is the solution to the \ac{PDE}. This output is constrained to satisfy multitask loss constituting of the initial condition, boundary conditions and the \ac{PDE}. These loss terms are often evaluated as \ac{MSE}.

In the analysis of incompressible fluid dynamics, it is fundamental to enforce the conservation of mass, expressed as zero divergence of the velocity field. For bounded domains with solid boundaries, such as in pipes or channels, the no-slip condition is typically imposed. The mathematical formulation is as follows:
\begin{subequations}
\begin{align}
  \nabla \cdot \mathbf{x} &= 0, \\
  \mathbf{x} &= \mathbf{0} \text{ on } \partial\Omega.
\end{align}
\end{subequations}
Here, $\mathbf{x}$ represents the velocity field vector, comprising components $u$, $v$ and $w$ in the three directions respectively. $\partial\Omega$ denotes the boundary of the domain. The \ac{PINN} architecture  for divergence-free flow problems incorporates two primary loss components to guide the training process: the \ac{PDE} loss ($\mathcal{L}_{PDE}$) and the Boundary Condition loss ($\mathcal{L}_{BC}$). The individual loss terms can be formulated as:
\begin{align}
\mathcal{L}_{PDE} = \frac{1}{N_p} \sum_{k=1}^{N_p} \left \| \nabla \cdot \hat{\mathbf{x}}(\mathbf{i}_k) \right\|^2, \quad
\mathcal{L}_{BC} = \frac{1}{N_b} \sum_{k=1}^{N_b} \left\| \hat{\mathbf{x}}(\mathbf{i}_k) - \mathbf{0} \right\|^2,
\end{align}
where, \(\hat{\mathbf{x}}\) represents the \ac{PINN}-predicted solution. The vector \(\mathbf{i}_k\) denotes the spatial coordinates of the \(k\)-th point within the domain or on the boundary, respectively for \ac{PDE} and boundary condition losses. \(N_p\) is the total number of collocation points used to evaluate the \ac{PDE} loss, distributed throughout the domain \(\Omega\). \(N_b\) is the total number of collocation points on the boundary \(\partial\Omega\), used to evaluate the boundary condition loss. Additionally, an optional data loss term ($\mathcal{L}_{Data}$) can be included if the solution to the \ac{PDE} is known at sparse locations:
\begin{equation}
    \mathcal{L}_{Data} = \frac{1}{N_d} \sum_{k=1}^{N_d} \left\| \hat{\mathbf{x}}(\mathbf{i}_k) - \mathbf{x}(\mathbf{i}_k) \right\|^2,
\end{equation}
where ${N_d}$ represents the number of data points where direct measurements of the velocity field $\mathbf{x}(\mathbf{i}_k)$ are available. For benchmarking the efficacy of \ac{PINN}-based frameworks, \ac{MAE} is calculated between the \ac{PINN} predicted solution and a known solution (analytical or numerical), though \ac{PINN}s themselves are employed as solvers for \ac{PDE}s:
\begin{equation}
\mathcal{L}_{\text{MAE}}(\hat{\mathbf{x}}) = \frac{1}{N_{v}} \sum_{k=1}^{N_{v}} \left\| \hat{\mathbf{x}}(\mathbf{i}_k) - \mathbf{x}(\mathbf{i}_k) \right\|,
\end{equation}
where \(\mathbf{x}(\mathbf{i}_k)\) represents the true velocity field at the point \(\mathbf{i}_k\), and \(N_v\) is the number of data points used for validation. This metric $\mathcal{L}_{\text{MAE}}$ which is not included in the training loss function, provides a measure of the average magnitude of error across all sampled points.

Importance sampling is a technique used to enhance the efficiency of the training process in machine learning models, including \ac{PINN}s. A probability distribution is first constructed that is proportional to the loss distribution observed across the computational domain. This distribution then guides the sampling process, whereby training points are more frequently selected from regions where the current model exhibits higher training loss. This method ensures that the neural network training concentrates on areas of the domain where the model is under-performing, thereby enhancing overall model accuracy. Essentially, this strategy allows for an adaptive allocation of computational resources, focusing efforts on 'problematic' areas much like adaptive mesh refinement in traditional computational methods~\cite{robert_monte_1999, martino_effective_2017, nabian_efficient_2021}.

Addressing the disparity in loss term magnitudes, which could bias learning towards boundary condition at the expense of accurately solving differential equation, balancing coefficients have been introduced. This, along with advancements in adaptive coefficient adjustment, underscores efforts to equalise the contribution of each loss component to the learning process~\cite{zhao_solving_2020, mcclenny_self-adaptive_2019, shi_non-fourier_2021}.

The development of adaptive activation functions represents another stride, introducing trainable parameters that fine-tune the activation function's slope, thus enhancing the network's ability to model complex patterns and behaviours~\cite{jagtap_adaptive_2020, jagtap2020locally}. Several neural network architectures have been developed to project the low-dimensional input of the \ac{PINN} to a higher dimension. This strategy is useful in addressing pathologies such as the spectral bias in neural networks, particularly beneficial for learning high-frequency functions indicative of abrupt changes such as discontinuity in the solution~\cite{tancik_fourier_2020, sirignano_dgm_2018, sharma_stiff-pdes_2023, sharma2023hyperparameter}. 

Domain decomposition within \ac{PINN}s, analogous to finite element methods but with enhanced flexibility, enables tackling complex geometries and discontinuities. Techniques like Conservative Physics-Informed Neural Network, Extended Physics-Informed Neural Network and parallel \ac{PINN}s have emerged, focusing on computational efficiency and modelling for discontinuities~\cite{Jagtap_XPINN, shukla2021parallel, hu_when_2021}. 
The ability to decompose the domain has been further enhanced in Finite Basis Physics-Informed Neural Network~\cite{moseley2023finite} by employing overlapping subdomains. Additionally, the Geometry Aware Physics-Informed Neural Network~\cite{oldenburg2022geometry} effectively compresses the complexity of irregular geometries, typically represented through unstructured meshes, into a latent space. This latent representation, alongside spatial coordinates $x,y,z$, serves as input for a subsequent network within the \ac{PINN} framework. A separate neural network is employed specifically to enforce boundary conditions with hard constraints during the training phase of the \ac{PINN}.

In recent developments, input encoding strategies for \ac{PINN}s have been crucial in handling unstructured data more effectively. These encodings can generally be categorised as Fourier type, involving combinations of sine and cosine functions, or non-Fourier types such as radial basis function and hash encoding. In~\cite{kast2024positional}, Fourier features were utilised to encode spatial coordinates of unstructured data points, alongside implementing a hard-constrained output in the neural network, which effectively eliminates the boundary condition loss ($\mathcal{L}_{BC}$). This study also introduced an active sampling technique, an advancement over traditional importance sampling, to efficiently resample unstructured points. Conversely, Zeng et al~\cite{zeng2024rbf} employed non-Fourier radial basis function encoding to manage inputs from unstructured datasets, complemented by a custom finite difference scheme designed to compute derivatives in scenarios involving discontinuous solutions. Meanwhile, Huang et al~\cite{huang2024efficient} demonstrated that hash encoding could significantly accelerate \ac{PINN} training, achieving up to a tenfold decrease in the training time. These encoding strategies not only enhance the computational performance of \ac{PINN}s but also significantly improve the management of unstructured data, influencing the convergence behaviours and learning dynamics. This necessitates careful consideration of their integration into existing architectures, particularly as the discontinuous nature of hash encoding poses unique challenges in derivative handling, impacting the stability and robustness of the model.

Additionally, a significant advancement in addressing the limitations of conventional \ac{PINN}s, particularly their struggle with temporal dependencies in dynamic physical systems, has been the development of a transformer-based framework known as PINNsFormer~\cite{zhao2023pinnsformer}. This framework utilises multi-head attention mechanisms not only to more accurately capture temporal dependencies but also to efficiently process unstructured data by transforming point-wise inputs into pseudo sequences. This adaptation enhances the model’s ability to deal with non-uniform and scattered data typically encountered in complex physical scenarios. 

Recent advancements in operator learning frameworks like DeepONet \cite{wang_learning_2021} and \ac{FNO} \cite{li2020fourier} have extended their application to unstructured grid data, addressing a long-standing limitation in solving \ac{PDE} across irregular geometries. For instance, the Mesh-Independent Neural Operator \cite{lee2022mesh,franco2023mesh} and Geo-FNO \cite{li2023fourier} have enabled efficient transformations of non-uniform domains into latent spaces, allowing for operations like fast Fourier transform to function seamlessly on arbitrary geometries. Similarly, frameworks such as Non-Uniform Neural Operator \cite{liu2023nuno} leverage domain decomposition techniques like K-D trees to handle non-uniform data while maintaining computational efficiency. \ac{FNO} with localised integral and differential kernels further address challenges of over-smoothing by introducing locally supported kernels, effectively bridging the gap between global and local feature representations \cite{liu2024neural}.

DeepONet variants have also seen remarkable progress in addressing problems related to unstructured grids. Geom-DeepONet \cite{he2024geom} augments inputs with signed distance functions and employs sinusoidal representation networks to predict field solutions across parameterised 3D geometries, demonstrating robust generalisation to unseen configurations. Physics-Constrained DeepONet \cite{jnini2024physics} integrates physical laws like divergence-free constraints to enhance learning efficiency, especially with sparse data. Furthermore, Enriched-DeepONet \cite{haghighat2024deeponet} and Decoder-DeepONet \cite{chen2024hybrid} tackle challenges such as moving-solution operators and unaligned observation data, improving accuracy by orders of magnitude compared to conventional approaches. These advancements underscore the growing capability of neural operators to efficiently solve \ac{PDE}s on unstructured grids.

In summary, \ac{PINN}s and neural operators represent a significant advancement in computational science, combining the strengths of neural networks with accurate physical modelling to address many limitations of traditional numerical methods for solving differential equations. By integrating input encoding techniques such as Fourier and non-Fourier methods, and employing novel strategies like multi-head attention mechanisms from transformer-based models, \ac{PINN}s are uniquely equipped to handle unstructured data. Additionally, new techniques in sampling, balancing loss terms, and adapting activation functions, along with domain-decomposition, enable \ac{PINN}s to work effectively with experimental data and solve complex physical equations.

\subsection{Reinforcement learning for unstructured mesh generation}
\label{sec:reinforce}
\subsubsection{Introduction to reinforcement learning}
Reinforcement learning is a machine-learning technique where an agent learns to make decisions by interacting with its environment~\cite{sutton2018reinforcement}. The learning process is guided by the rewards or penalties received from trial-and-error interactions, resulting in an optimal decision-making policy that maximises cumulative rewards over time. Since a complex problem could often be divided into a sequence of manageable sub-tasks, the overall performance hinges on the ability to effectively solve these sub-problems. \ac{RL} starts to be applied to solve complex geometrical problems by providing various finite element analysis methods.

\begin{figure*}[!ht]
\centering
\includegraphics[width=1.0\textwidth]{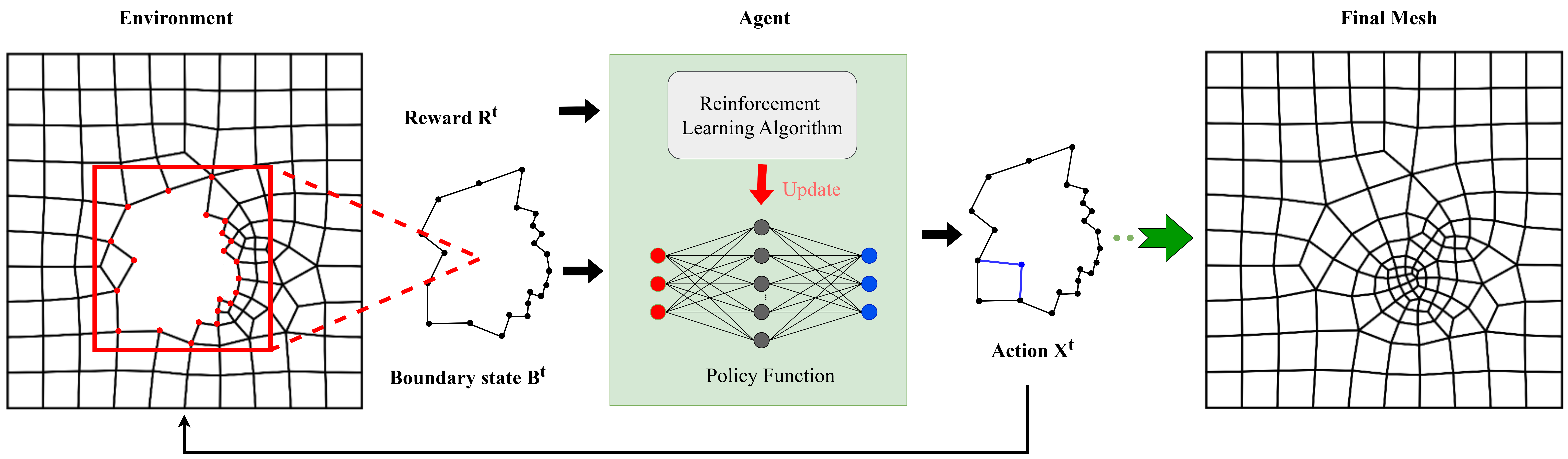}
\caption{Reinforcement learning for unstructured mesh generation.}
\label{fig:rl_mesh}
\end{figure*}

\ac{RL} has recently been applied to solve unstructured mesh generation problems, due to its underlying mechanism of recursive interactions between element extractions and its domain boundary environment~\cite{zeng1993knowledge, yao2005ann}. The work of~\cite{pan2021self, pan2023reinforcement} formulated the mesh generation problem as sequential decision making problems and developed RL-based methods to generate quadrilateral meshes for arbitrary \ac{2D} geometries. The overall mesh generation procedure is shown in Figure~\ref{fig:rl_mesh}. The meshing process is formulated as a Markov decision process, consisting of a set of boundary environment states $\mathcal{B}$, a set of possible actions $\mathcal{X}(b)$, a set of rewards $\mathcal{R}$, and a state transition probability 
\begin{align}
    P(B^{t+1} = b', R^{t+1} = r|B^t = b, X^t = x),
\end{align}
 where $b'$, $b \in \mathcal{B}$, $x \in \mathcal{X}(b)$, $r \in \mathcal{R}$, $b'$ is the new state at $t+1$ and $r$ is the reward after action $x$ on the boundary state $b$.  
It is important to note that the state (solution) variable $x$ in this section represents the action of mesh generation, unlike in other sections where it denotes simulation or prediction results.

The mesh generator at each time step $t$, acting as the \ac{RL} agent, observes a state $B^t$ from the environment, which is the contour of the geometry consisting of a set of vertices and segments. It then performs an action $X^t$ to generate a new element from the given state based on an implicit or explicit policy. The environment responds to the action by excluding the newly generated element and transits into a new state $B^{t+1}$, which is the updated geometry contour. It provides a reward or penalty $R^t$ simultaneously that measures the quality of the generated element and the updated contour, such as a value of 1 for a well-balanced quality and -1 for an imbalance. The latter could potentially lead to sharp angles and narrow regions. This element generation process continues until the updated contour includes only four vertices, forming the last element automatically. Eventually, it produces a sequence of $[B^0, X^0, B^1, R^1, X^1, ...]$. The agent's goal is to learn such a policy $\pi(x|b)$ that maximises the cumulative rewards $G^{r}_t$, 
\begin{equation} 
G^{r}_t= \sum_{k=t+1}^{T} \gamma^{(k-t-1)} R_k.
\end{equation}
where $0 \leq \gamma \leq 1$ is a discount rate that determines the weight between short-term and long-term rewards, and $T$ is the total time step.

In deep \ac{RL}, the policy $\pi_\Phi$ is represented by a nonlinear neural network with parameters $\theta$, which were updated during the training process. Different types of deep \ac{RL} algorithms could be applied to find an optimal policy $\pi_\theta^*$ that maximises an objective function $J$ defined as the expectation of the return. $\pi_\theta$ is updated by $\nabla_\theta{J}$ which is calculated using the 
deterministic policy gradient algorithm by applying the chain rule to $J$ as follows:
\begin{equation}
    \nabla_\theta(J)= \mathbb{E}_{(b^t,x^t)}\left[\nabla_\theta{Q^v(b^t, x^t)}\right],
\end{equation}
where $Q^v(b^t, x^t)$ is an action value function defined by
\begin{equation}
    Q^v(b^t, x^t)= \mathbb{E}_{\pi}\left[R^t|b^t, x^t\right].
\end{equation}
Here the notation $Q^v$ for the value function is chosen to avoid conflict with the query vector notation of the transformers.
The action value function estimates the long-term benefit of taking an action for a state, which is how well the newly generated element will contribute to the optimal mesh quality. The mesh quality is measured by metrics related to the mesh’s geometrical and topological properties~\cite{sorgente2023survey}. Common geometrical metrics for a mesh include minimum and maximum angles, edge ratio, aspect ratio, stretch, taper, skewness, and scaled Jacobian, whereas topological metric includes singularity. Generally, a mesh cannot achieve high scores on every metric because of the complexity of geometries and specific computational requirements. The choice of metric is often related to downstream applications. The stepwise reward in the Q-function needs to balance the element quality and mesh quality because the optimal individual element does not always lead to optimal meshes. Pan et al.~\cite{pan2021self} demonstrated that a reward function considering the trade-off between element quality and the remaining contour could achieve overall good performance. 

Once the reward function is defined, different types of \ac{RL} algorithms could be applied to estimate the Q-function and the optimal policy. Pan et al.~\cite{pan2021self} applied Advantage Actor-Critic to learn an initial meshing policy and then trained a feedforward neural network as the final policy with high-quality samples from the RL agent. With the similar formulation, Tong et al.~\cite{tong2023srl} further expanded the action space of extracting a new element. To achieve a stable learning efficiency, Pan et al.~\cite{pan2023reinforcement} solely implemented Soft Actor-Critic to learn the meshing policy, which has enabled it for more complex geometries. 

\subsubsection{Reinforcement learning for mesh optimisation}
Mesh optimisation for complex systems plays an important role in dynamically refining mesh regions with low or high solution variability, facilitating a favorable trade-off between computational speed and simulation accuracy. 
The application of \ac{RL} to this research area has recently started.
Wang et al.~\cite{wang2024unstructured} proposed a smoothing method to improve the quality of triangular meshes by combining a heuristic Laplacian method with the Deep Deterministic Policy Gradient algorithm. The learned policy maximises the overall mesh skewness quality by adjusting the positions of free nodes. 

Some research directly optimise meshes towards an accurate and efficient numerical approximation to solve \ac{PDE}. Reinforcement learning is applied to automate this process, sidestepping a large amount of heuristic rules. The recent work of~\cite{yang2023reinforcement} formulated adaptive mesh refinement into a Markov decision process with variable sizes of state and action spaces. It makes elementwise decisions with policy trained by REINFORCE and Proximal Policy Optimisation (PPO), to minimise the final \ac{PDE} solution error. Gillette et al.~\cite{gillette2024learning} applied Proximal Policy Optimisation to train a policy that marks a set of elements to be refined based on existing error estimates. The specific refinement strategy was based on prior knowledge. Instead of only focusing on refining meshes, Foucart et al.~\cite{foucart2023deep} formulated the Adaptive Mesh Refinement as a partially observable Markov decision process and learned a policy that could de-refine mesh elements with local surrounding information. Lorsung et al.~\cite{lorsung2023mesh} applied a deep Q network to iteratively coarsen meshes, thereby reducing simulation complexity while preserving the accuracy of the target properties.

To avoid iteratively refine the elements, some studies applied multi-agent \ac{RL}. Yang et al.~\cite{yang2022multi} further investigates multi-agent \ac{GNN} with a team reward to boost the refinement speed for arbitrary meshes. Freymuth et al.~\cite{freymuth2024swarm} formulated the adaptive mesh refinement as a Swarm Markov Decision Process, treating each mesh element as an agent and training all agents under a shared policy using \acp{GNN}. Dzantic et al.~\cite{dzanic2024dynamo} proposed a multi-agent PPO to learn anticipatory refinement strategies, specifically for discontinuous Galerkin finite difference method.

A few studies were using RL to optimise the shape of geometries with existing mesh generators, such as fin-shaped geometries for optimal heat exchange~\cite{keramati2022deep} and blade passages with optimal meshing parameters~\cite{kim4852465automatic}. Some complex geometries often require a well decomposition before meshing. Diprete et al.~\cite{diprete2024reinforcement} used PPO to train an agent to perform optimal cuts on Computer Aided Design models, decomposing them into well-shaped rectangular blocks suitable for generating high-quality meshes. 

In summary, \ac{RL} is making significant strides in generating and optimising unstructured meshes for various engineering applications, enhancing both computational efficiency and simulation accuracy. \ac{RL} generally reduces reliance on heuristic rules and enhances the accuracy of numerical approximations for \ac{PDE}s. \ac{RL}-based unstructured meshes could be applied to fields, including fluid dynamics by capturing complex flow patterns, refining meshes in structural engineering by predicting stress distributions and material behaviour, and modelling biological tissues and organs, which facilitates better surgical planning and medical device design. Future research in \ac{RL} for unstructured meshes, such as:

\begin{enumerate}
    \item Leveraging multi-agent \ac{RL} to coordinate meshing and refinement strategies across mesh elements, as seen in the use of \acp{GNN} and swarm intelligence.
    \item Extending \ac{RL} applications to optimise geometry shapes in conjunction with mesh generation, improving design efficiency in fields like heat exchange and fluid dynamics.
    \item Developing \ac{RL} agents capable of performing optimal preprocessing on Computer Aided Design models, enabling better decomposition for high-quality mesh generation.
    \item Evaluating the \ac{RL}-based methods in diverse applications, including fluid dynamics, structural engineering, and biomedical engineering.
\end{enumerate}

\subsection{Generative AI models with unstructured grid data }

\subsubsection{Introduction to generative models}

Similar to physics-informed machine learning (Section \ref{sec:PINNS}) and reinforcement learning (Section \ref{sec:reinforce}), generative modelling characterises a set of techniques towards a specific inference goal, independent from specific neural network architectures.
As such, all previously discussed architectures for handling unstructured data can also be applied to generative modelling.
Generative modelling seeks to learn an efficient sampler from a state distribution $\bx \sim p_{\theta}(\bx)$ as parametrised by the weights and biases $\theta$ of a neural network.
By modelling a conditional distribution ${\bx \sim p_{\theta}(\bx\mid\mathbf{c})}$, where $\mathbf{c} \in \mathcal{C}$ is the conditioning information, e.g., initial conditions, a coarse-grained field, or observations, drawn from the conditioning space $\mathcal{C}$, such a sampler can solve many tasks like forecasting, downscaling, or Bayesian inference.
To avoid specifying an explicit distribution, implicit generative modelling represents the distribution through samples generated with a learnable generator applied to samples from a known and possibly simpler distribution \citep{mohamed_learning_2016}.
For a more detailed overview of generative modelling, we refer to the books of \citet{tomczak_deep_2022, murphy_probabilistic_2023}.

A straightforward way to learn such a generative model is to maximise the data likelihood given the assumed model, known as the maximum likelihood approach.
Most models for generative modelling can be trained this way: energy-based models \citep{jaynes_information_1957, lecun_tutorial_2006} and normalizing flows \citep{laparra_iterative_2011,papamakarios_normalizing_2021} are examples, as well as \acp{ARM} \citep{frey_graphical_1998, larochelle_neural_2011}, which often rely on maximum likelihood estimation.
Aiming to maximise a lower bound on the data likelihood, \acp{VAE} \citep{kingma_auto-encoding_2013, rezende_stochastic_2014} and \acp{DDM} \citep{sohl-dickstein_deep_2015, ho_denoising_2020, song_generative_2020} optimise the so-called \ac{ELBO}.
Differing from that maximum likelihood approach, \acp{GAN} \citep{goodfellow_generative_2014, arjovsky_wasserstein_2017} are trained with an adversarial loss: concurrently to the generator, a deep learning-based discriminator, or critic, is trained to discriminate between true data samples and samples from the generator. 

\begin{figure}[ht]
\centering
\includegraphics[width=0.4\textwidth]{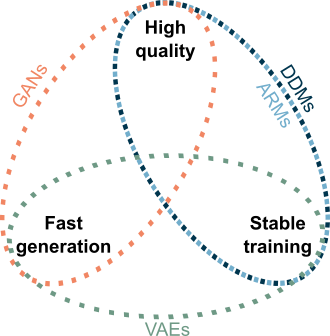}
\caption{
    The trilemma of the four most popular generative model types: all models can be framed into this triangle, and there is yet to find a method that could concurrently exhibit the three assets.
    Inspired by Figure 1 from \citet{xiao_tackling_2021}.
    \label{fig:genai_trilemma}
}
\end{figure}

\acp{GAN} can generate high-quality samples with high throughput but are notoriously unstable to train \citep{arjovsky_towards_2022} and result into mode collapse, where only partial modes of the data are generated.
Contrastingly, maximum likelihood approaches are more stable and easier to train, and they can exhibit much higher sample diversity than \acp{GAN}.
\acp{VAE} can generate samples in one step like \acp{GAN}, but they often produce lower quality samples due to data compression.
Conversely, \acp{DDM} and \acp{ARM} can generate high-quality samples but rely on slower multi-step sampling.
Thus, there is a trilemma between training stability, sample quality, and generation speed \citep{xiao_tackling_2021}.
Figure \ref{fig:genai_trilemma} categorises the main generative modelling approaches within this trilemma, while a general overview over the most popular approaches and their original citations is shown in Table \ref{tab:genai_overview}.
Since \acp{DDM} are currently the most popular generative modelling approach, we concentrate on its application to unstructured data; however, most of the reviewed applications could also work with other types of generative modelling.

\begin{table}[ht]
    \centering
    \caption{
        The main generative deep learning methods, the criterion on which their loss function is based, and the original or overview papers where they are further described.
        The abbreviations for the loss functions are: \mbox{MaxLL = maximum likelihood; ELBO = evidence lower bound.}
        Table inspired by Table 20.1 from \citet{murphy_probabilistic_2023}.\label{tab:genai_overview}
    }
    \begin{tabular}{lcr}
        \hline
        \hline
        Method (abbreviation) & Loss & Citation\\
        \hline
        \hline
        Energy-based models & MaxLL & \cite{jaynes_information_1957, lecun_tutorial_2006} \\
        Normalizing flows & MaxLL  & \cite{laparra_iterative_2011,papamakarios_normalizing_2021}\\
        Autoregressive models (ARMs) & MaxLL & \cite{frey_graphical_1998, larochelle_neural_2011}\\
        Variational autoencoders (VAEs) & ELBO  & \cite{kingma_auto-encoding_2013, rezende_stochastic_2014}  \\
        Denoising diffusion models (DDMs) & ELBO  & \cite{sohl-dickstein_deep_2015, ho_denoising_2020, song_generative_2020} \\
        Generative adversarial networks (GANs) & Adversarial  & \cite{goodfellow_generative_2014, arjovsky_wasserstein_2017} \\
        \hline
        \hline
    \end{tabular}
\end{table}

\acp{DDM} establish bidirectional pathways between the distribution of training data and a stationary noise distribution, typically a Gaussian distribution \citep{sohl-dickstein_deep_2015, ho_denoising_2020}.
In the forward path, noise gradually replaces information in the data through a diffusion process.
Conversely, the backward path requires training neural networks to denoise the data given a noised field and conditional information.
Trained across all noise magnitudes, the neural network can then iteratively map initial fields from the stationary noise distribution back to data space.
Being inherently probabilistic, the diffusion process is governed by a \ac{SDE} \citep{song_score-based_2021}, and the backward path corresponds to a reverse \ac{SDE} \citep{anderson_reverse-time_1982}.
This connects \acp{DDM} to score-based generative modelling \citep{vincent_connection_2011, song_generative_2020, song_improved_2020}, where the neural network approximates the data distribution's score.
Additionally, this connection allows to draw similarities to more traditional physical modelling approaches \citep{finn_representation_2023, ruhling_cachay_dyffusion_2023, Holzschuh_Vegetti_Thuerey_2023}.
There exists also a deterministic \ac{ODE} for the backward path \citep{song_denoising_2020, song_score-based_2021}, yielding the same marginal distribution as the reverse \ac{SDE}.
\acp{DDM} enable advanced generative tasks such as text-to-image or text-to-video generation, enhancing generative training stability and quality \citep{dhariwal_diffusion_2021}.
However, their iterative nature incurs high computational costs and ongoing research aims to mitigate these costs \citep{rombach_high-resolution_2022}.

\subsubsection{Generative models applied on unstructured data}

\begin{figure}[ht]
\centering
\includegraphics[width=0.68\textwidth]{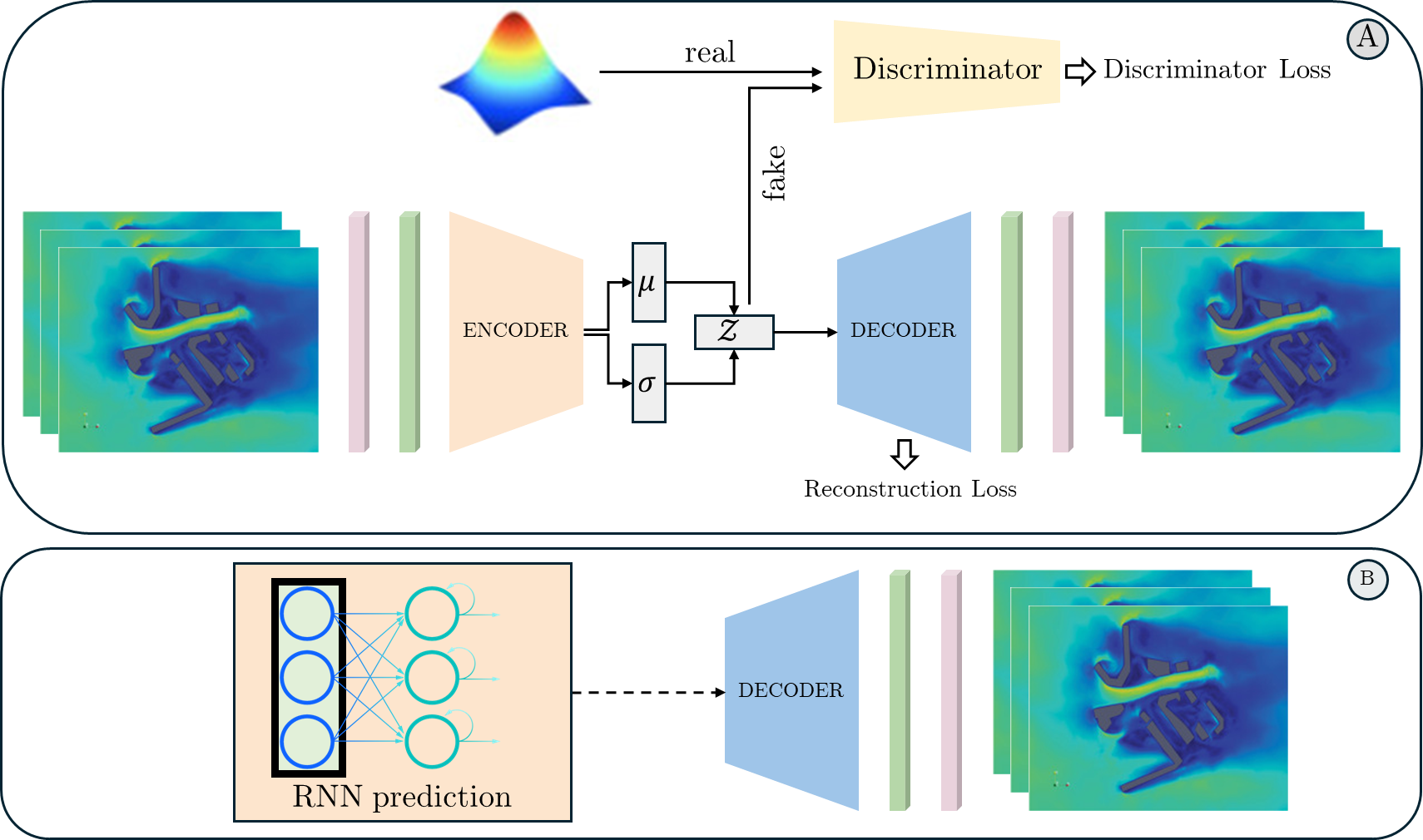}
\caption{Variational autoencoder and generative adversarial networks can be combined for forecasting on unstructured grids. Figure from \citet{quilodran2023data}.\label{fig:combine_vae_gan}}
\end{figure}

Early application of generative models to unstructured data are especially based on \acp{GAN} or \acp{VAE}.
Specific examples are their application to material modelling \citep{Kadeethum_material_2022} and flood forecasting \citep{Cheng_flood_2021}.
Combining these methods, \citet{quilodran2023data} uses together a \ac{VAE}, a \ac{GAN}, and a \ac{RNN} in latent space for forecasting on unstructured grids, as exemplarily shown in Figure
\ref{fig:combine_vae_gan}.
Such applications of GANs on spatio-temporal problems are further surveyed in \citet{Gao_survey_2022}.

However, the first applications of \acp{DDM} highlighted their potential for complex fluid prediction \citep{yang_denoising_2023, finn_generative_24}, including for unstructured grids.
In Google DeepMind's GenCast \citep{price_gencast_2024}, a DDM designed for global weather prediction outperforms all other ensemble-based prediction methods, including ensemble forecasts from the European Centre for Medium-range Weather Forecasting (ECMWF).
GenCast uses a graph neural network inspired by the GraphCast model \citep{lam_learning_2023}, incorporating graph transformer blocks to represent data on a spherical grid, an approach also applicable to unstructured grids.

\acp{DDM} based purely on transformer blocks and attention mechanisms \citep{peebles_scalable_2023} are well-suited for unstructured data.
Additionally, the latent mapping in latent diffusion models \citep{vahdat_score-based_2021, rombach_high-resolution_2022} can be readily adapted using the methods from Section \ref{sec:reordering} to transition from an unstructured to a latent structured grid, facilitating the application of diffusion models.
However, as discussed in \cite{price_gencast_2024}, the diffusion process must isotropically replace information with noise, potentially requiring alternative noise sampling strategies beyond purely random sampling on unstructured grids.

\acp{DDM} can also interact with \acp{NERF}, which are analogous to \acp{PINN} without the physics-informed loss function.
In this context, \acp{DDM} can regularise the sampled fields \citep{wynn_diffusionerf_2023} to more closely follow the DDM-generated data distribution. 
Furthermore, \acp{NERF} enable to train \acp{DDM} for 3D synthesis based on pre-trained 2D text-to-image models \citep{lin_magic3d_2023, poole_dreamfusion_2022, wang_prolificdreamer_2023, wang_score_2022}.
As \acp{NERF} represent functions of coordinates, they are inherently grid-independent and suitable for unstructured grids. 
However, the combined application of \acp{NERF} and \acp{DDM} can be computationally demanding, as both methods typically require substantial computational resources.

Neural operators aim to map between infinite-dimensional function spaces, allowing evaluation at arbitrary spatial positions \citep{kovachki_neural_2023, li2020fourier}.
To learn \acp{DDM} on such spaces, data at arbitrary positions can be interpolated to a structured grid, enabling the application of CNNs \citep{bond-taylor_infty-diff_2024}.
Additionally, convolutional layers can be replaced with implicit NERF-like representations \citep{gao_implicit_2023}, facilitating a grid-independent super-resolution operator.
By integrating \acp{DDM}, neural operators can leverage the robust generative capabilities to efficiently handle complex and stochastic function mappings, potentially enhancing the accuracy and generalisation of mappings between function spaces.

Generative models have further been applied to three-dimensional meshes and point clouds, which are crucial for modelling entities such as the human body.
\acp{VAE} and \acp{GAN} are frequently utilised in this context \citep{Liu_interactive_3dgan_2017, Tan_vae3d_2018, Foti_3dlatentvae_2022, Tata_3dgan_survey_2023, Molnar_vae_3d_2024}.
Nonetheless, with their recent emergence in image generation, \acp{ARM} \citep{Chen_meshxl_2024} and \acp{DDM} \citep{Alliegro_polydiff_2023,Liu_meshdiffusion_2023} are increasingly employed for the generation of such data.

In all these applications, generative models have emerged as powerful tools for handling unstructured data, offering a flexible approach to model complex and irregular datasets.
These models excel in capturing the underlying distributions of unstructured data, enabling sophisticated applications such as advanced weather prediction, as demonstrated by models like GenCast from Google DeepMind.
Furthermore, their adaptability to unstructured grids through techniques like graph neural networks and transformers enhances their utility in geospatial and scientific computing domains.
However, the computational demands of these models, particularly when integrating with Neural Radiance Fields (NERFs) and neural operators, present significant challenges that require high computational resources and efficient algorithms.
Despite these challenges, the ongoing research and development in optimising these models and expanding their applications suggest a bright future, promising enhanced accuracy, efficiency, and broader adoption across various scientific disciplines.

\section{Public study cases and benchmarks }


This section introduces the currently available open-access dataset for computational problems involving unstructured grid data, which can serve as test cases for evaluating the efficacy of machine learning models in simulating dynamical systems on unstructured meshes.

For computational solid dynamics, it faces challenges in defining universal standard test cases due to the diverse nature of research areas. For example, in material mechanics, common tests for unstructured meshes include material tension/compression~\cite{pagan2022graph}, bending~\cite{maurizi2022predicting}, and fatigue tests~\cite{bomidi2013experimental, hanlon2019artificial}, each tailored to specific material properties and behaviors.
Shifting to \ac{CFD}, we classify test cases with unstructured meshes into three distinct categories based on the temporal characteristics of the flow and the adaptability of the mesh. Each category is designed to evaluate specific aspects of numerical methods and algorithms, reflecting the broad range of flow phenomena and challenges encountered in fluid dynamics simulations.

\begin{figure*}[!ht]
\centering
\includegraphics[width=0.8\textwidth]{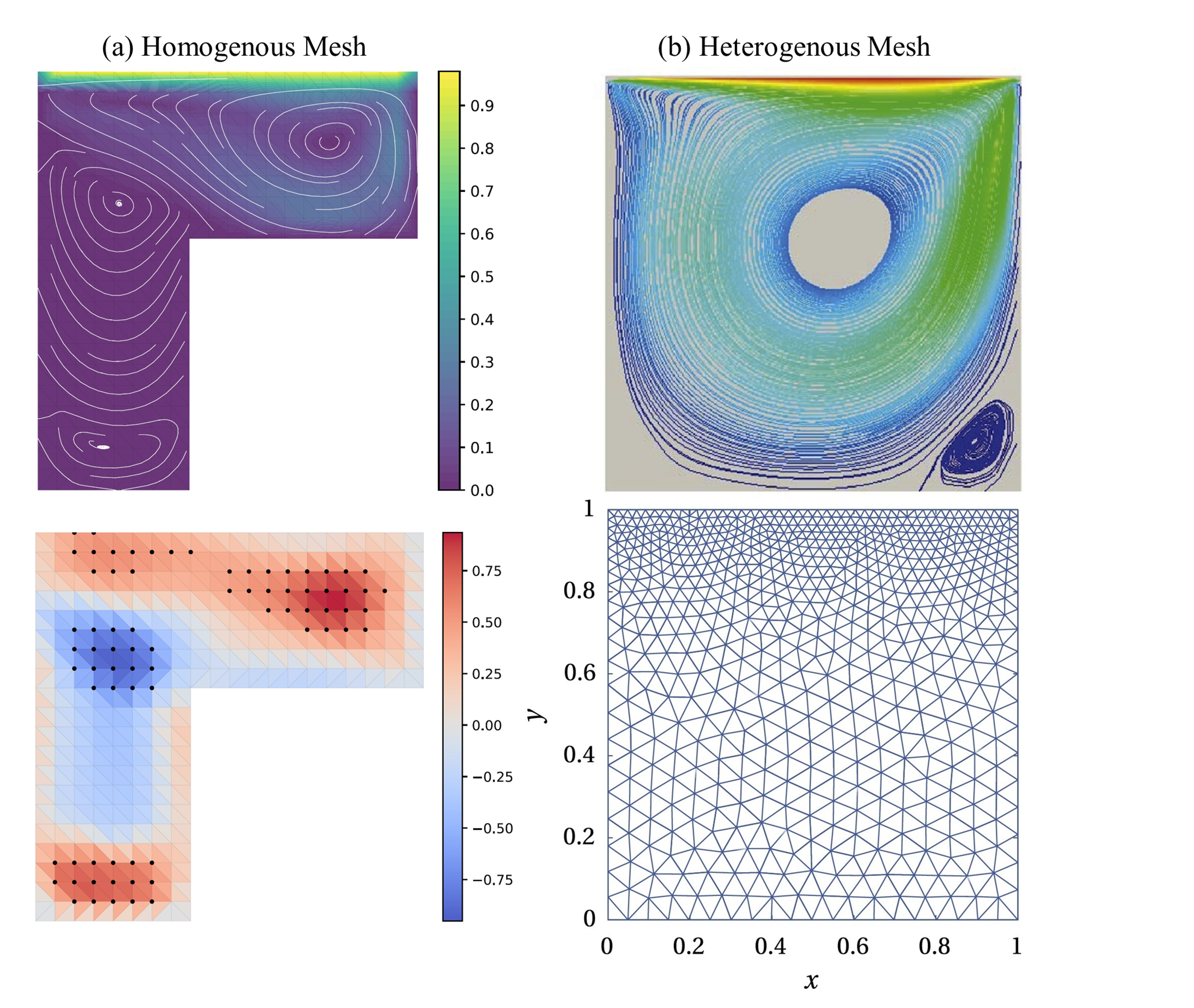}
\caption{Steady flow cases: Computational domain and computational mesh for lid-driven cavity flow of (a) homogenous mesh~\cite{hu2023graph} and (b) heterogenous mesh~\cite{hesthaven2018non}}
\label{fig:Steady Flow Cases}
\end{figure*}

\paragraph{Steady flow with unstructured meshes} This category includes open-access datasets such as lid-driven cavity flow ~\cite{hesthaven2018non, hu2023graph}, isentropic vortex~\cite{zhou2024machine} and fluid dynamics around diverse shapes at low \ac{Re}~\cite{tlales2024machine, chen2021graph}. These datasets primarily utilise triangular and quadrilateral mesh, offering a range of mesh densities and configurations. For example, in the case of flow around a cylinder when \ac{Re} ranging from 10 to 40, the mesh resolution is varied to accurately capture the boundary layer and wake regions. Finer meshes are employed near the cylinder surface and in areas with steep gradients to resolve complex flow features at different \ac{Re}s~\cite{ranade2021discretizationnet}. Additionally, these datasets provide comprehensive details on flow parameters, boundary conditions, and mesh specifications. Specifically, the lid-driven cavity flow datasets feature variations in \ac{Re}, ranging from hundreds to thousands, enabling the study of flow behaviours from laminar to transitional regimes. Furthermore, the shape of the cavity can be modified to introduce additional complexity, as demonstrated in Figure~\ref{fig:Steady Flow Cases}(a). These variations provide a versatile framework for comprehensively testing numerical accuracy, convergence, and the adaptability of computational methods to different geometrical configurations. The comparison between homogeneous and heterogeneous meshes is illustrated in Figure~\ref{fig:Steady Flow Cases}, highlighting how uniform triangular meshes offer consistent resolution throughout the flow domain, while heterogeneous meshes adaptively refine regions of interest, such as areas with sharp gradients. 

\begin{figure*}[!ht]
\centering
\includegraphics[width=1\textwidth]{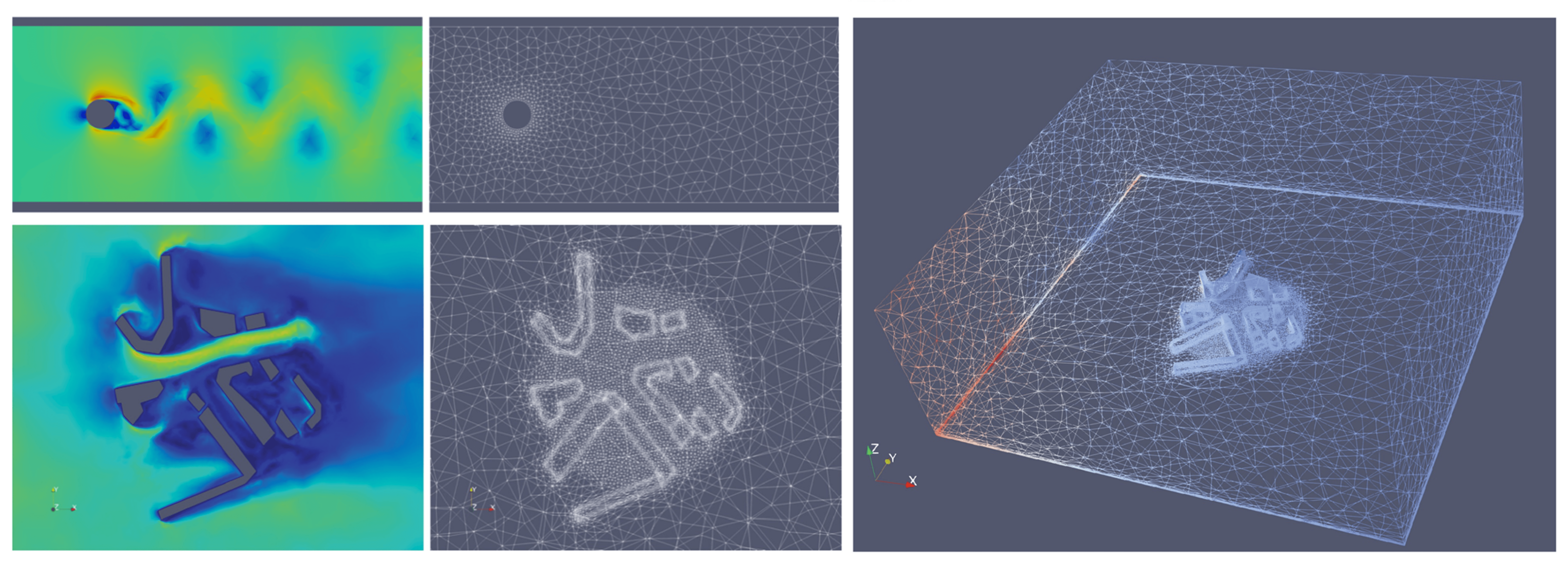}
\caption{Dynamic flow with fixed unstructured meshes cases: Computational domain and computational mesh for flow of water around a cylinder, and azimuthal and isometric view of urban air pollution simulation~\cite{quilodran2023data}}
\label{fig:Dynamic with Fixed Unstructured Meshes}
\end{figure*}

\paragraph{Time-dependent flow with fixed unstructured meshes} This category is pivotal for simulating dynamic flows using a static mesh, essential for assessing algorithms' effectiveness in depicting time-varying fluid behaviours. Unlike the previous category, which focuses on steady-state simulations, this category deals with unsteady flows and incorporates time series data to capture transient phenomena. It is the most broadly utilised technique in \ac{CFD}, dedicated to test cases that include the dynamics of flow around various shaped objects in high \ac{Re}~\cite{lav2018improvement, reddy2019reduced, he2022flow} such as cylinders, triangles~\cite{lee2024grid} and airfoils~\cite{xu2021convolutional, nemati2023data}, alongside complex configurations~\cite{wang2023graph, chang2021development} such as channel and backward-facing step flows, and extend to large-scale environmental phenomena~\cite{quilodran2023data, shao2023pignn}. The general mesh settings are similar to those in the 'steady flow with unstructured meshes' category, employing a mix of triangular and quadrilateral meshes~\cite{he2022flow}, often in irregular formations. However, the main differences lie in the physical parameters, data structure and typically involve larger meshes with higher resolutions. Datasets in this category include more mesh elements to accurately capture complex, unsteady flow phenomena over time. This increased mesh density is essential for resolving finer details in high-fidelity simulations of dynamical flows. An illustrative example is the flow around a cylinder~\cite{tlales2024machine}, as \ac{Re} increases from $40$ to $3900$, the flow regime transitions from steady laminar to unsteady. This transition necessitates increasing the mesh size from $648$ to $20,736$ elements to accurately capture the complex flow dynamics. Figure~\ref{fig:Dynamic with Fixed Unstructured Meshes} exhibits typical examples of time-dependent flow with fixed meshes. The two-dimensional visualisation captures flow around a cylinder at a \ac{Re} of $2300$, using a mesh with $5166$ nodes across $1000$ time steps. In a three-dimensional context, the figure portrays an urban air pollution simulation that incorporates a more extensive mesh with $100,040$ nodes per dimension over $1000$ time steps.

\begin{figure*}[!ht]
\centering
\includegraphics[width=0.85\textwidth]{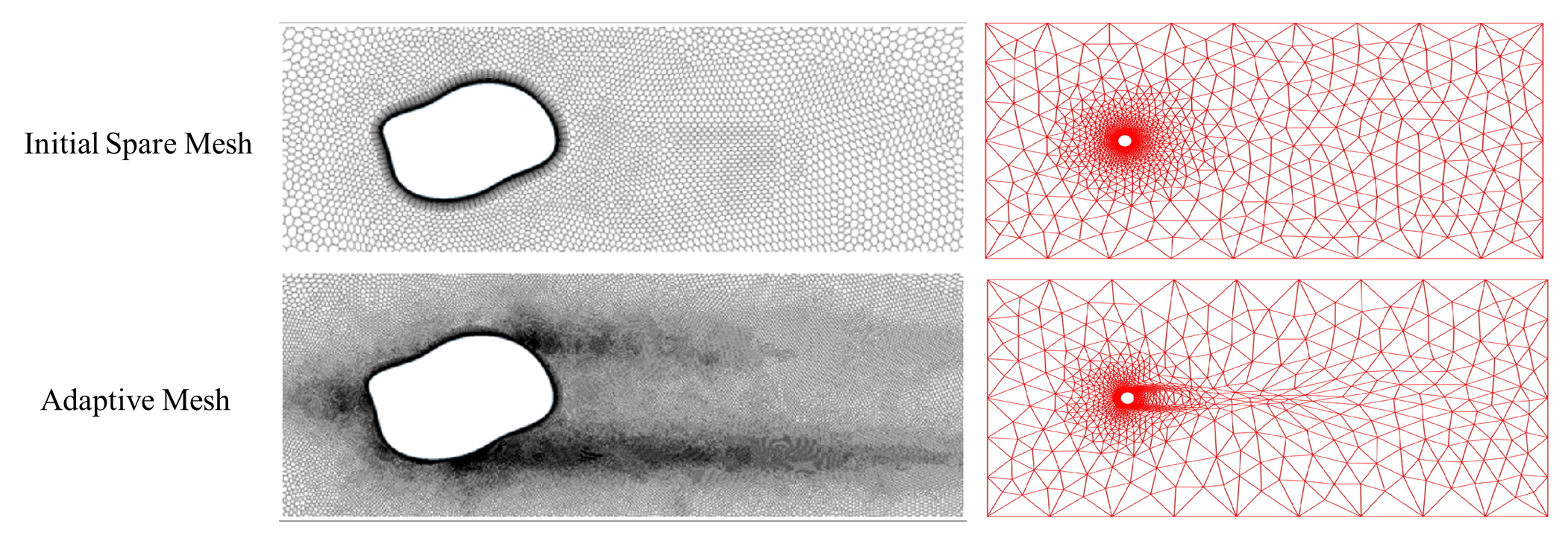}
\caption{Dynamic flow with adaptive unstructured meshes cases: Initial spare mesh and adaptive mesh for flow of water around a irregular object (left)~\cite{huang2021machine} and cylinder (right)~\cite{tingfan2022mesh}}
\label{fig:Dynamic with Adaptive Meshes}
\end{figure*}

\paragraph{Time-dependent flow with adaptive meshes} This category features datasets where the mesh resolution is dynamically adjusted during simulations. This approach enhances simulation efficiency and accuracy by refining the mesh in regions with complex flow dynamics and coarsening it where less detail is needed, significantly reducing computational costs while capturing essential flow features. It is particularly effective for physical disturbances such as aerodynamic flows around objects~\cite{pfaff2020learning, ojha2022initial}, turbulence~\cite{huang2021machine, hasegawa2020machine}, and fluid-structure interactions~\cite{peng2022grid, whisenant2020galerkin}, benefiting from the adaptive refinement's precision in complex flow areas. Specifically, adaptive meshing enables more intricate adjustments, which is particularly beneficial when dealing with scenarios that are similar but not identical. For instance, in aerodynamic studies, datasets often include simulations of flow around airfoils with slight variations in angle of attack or different \ac{Re}s~\cite{ojha2022initial}. These variations introduce changes in flow features such as boundary layer separation, vortex shedding, and pressure distribution, requiring the mesh to adapt dynamically to accurately capture these phenomena. The integration with multi-scale analysis is less common in this category, as it can be inherently related to multi-scale analysis by its nature. Figure~\ref{fig:Dynamic with Adaptive Meshes} shows the evolution from initial to adaptively refined meshes in two cases. Initially, heterogeneous meshes concentrate on refinement in regions anticipated to require higher resolution. Over time, through adaptive mesh refinement processes, these meshes undergo successive refinements, optimizing the mesh density according to the evolving flow dynamics, such as areas of high gradient or vorticity.

By summarising the characteristics of the test cases in these three categories, we provide a reference for researchers applying \ac{ML} to \ac{CFD}. This overview assists in selecting appropriate benchmarks to validate model performance. Table~\ref{tab:test-cases} complements this discussion by presenting a comprehensive array of CFD test scenarios, detailing the specific characteristics of each test case along with available codes and datasets.

\begin{sidewaystable}[!ht]
\centering 
\caption{Summary of test cases in fluid dynamics with unstructured meshes} 
\label{tab:test-cases} 

\scalebox{0.8}{ 
    \begin{tabular}{@{}ccccccccc@{}}
    \toprule
    Category & Test case & Driven equation & Method & Solver & Adaptive mesh & Multi-scale & Available code/Dataset \\ \midrule
    \multirow{4}{*}{\parbox{2cm}{\centering Basic fluid flow}} 
    & Advection & Advection equation & FDM & - & No & Yes & \href{https://dynabench.github.io/}{Data}
\cite{dulny2023dynabench} \\
    & Burgers' equation & Burgers' equation & FDM & - & No & Yes & \cite{kumar2023deep} \\
    & \multirow{2}{*}{Wave problem} & \multirow{2}{*}{\shortstack{Wave equation/\\Shallow water equation}} &  \multirow{2}{*}{FDM/FVM} & \multirow{2}{*}{OpenFOAM/$\cdots$} & \multirow{2}{*}{No} & \multirow{2}{*}{Yes} & \multirow{2}{*}{\href{https://dynabench.github.io/}{Data}\cite{dulny2023dynabench}} \\
    &  &  &  &  &  &  \\
    \midrule
    \multirow{5}{*}{\parbox{3cm}{\centering Common physical phenomena}} & Lid-driven cavity & NS & FVM/FEM & FLUENT/OpenFOAM$\cdots$ & No & No & - \\
    & \multirow{2}{*}{Disturbance around objects} & \multirow{2}{*}{NS} & \multirow{2}{*}{RANS/FVM} & \multirow{2}{*}{SU2/FLUENT/$\cdots$} & \multirow{2}{*}{Yes} & \multirow{2}{*}{No} & \multirow{2}{*}{\shortstack{\href{https://github.com/cfl-minds/gnn_laminar_flow}{Data}\cite{chen2021graph}\cite{zhou2024machine} \\ \href{https://github.com/google-deepmind/deepmind-research}{Data}\cite{pfaff2020learning}\href{https://github.com/locuslab/cfd-gcn}{Data}\cite{belbute2020combining}\cite{selig1996uiuc}}} \\
    &  &  &  &  &  &  \\
    & Channel flow & NS & RANS/FVM & OpenFOAM/Code Saturne/$\cdots$ & Yes & Yes & \cite{wang2023graph,archambeau2004code} \\
    & Large-scale phenomena & - & RANS/FVM & PHIFLOW/FLUENT/$\cdots$ & Yes & Yes & \cite{quilodran2023data, hersbach2020era5}\href{https://data.marine.copernicus.eu/products}{Data}\cite{copernicus2024} \\
    \bottomrule
    \end{tabular}
}

\raggedright 
\scriptsize 
\textbf{Note:}
\begin{itemize}
    \item The methods and solvers listed in this table represent a summary of findings from select publications.
    \item Each method and solver has its advantages and limitations, and should be chosen based on specific project requirements.
    \item FDM: Finite Difference Method; FEM: Finite Element Analysis; FVM: Finite Volume Analysis.
\end{itemize}
\end{sidewaystable}
\label{sec:dataset} 

\section{Discussion and conclusion}

This review highlights significant advancements and current achievements in the application of machine learning for handling unstructured grid data in computational physics. Notable strides include the adaptation of neural networks such as \acp{CNN}, \acp{GNN} and transformers for irregular geometries in complex dynamical systems. These models have demonstrated potential in various circumstances and specific applications. One may choose different neural network structures to work with, depending on the specific requirements and the availability of data or computational resources, as summarised at the end of Section \ref{sec:ML_unstructure}.

On the other hand, different learning paradigms can be seamlessly applied to unstructured grid data, as discussed in this review. Physics-Informed Neural Networks provide a meshless solution by leveraging the power of auto-differentiation in neural networks and incorporating physics knowledge. Reinforcement learning, from another perspective, can be utilised to optimise mesh generation in computational systems. The challenge of data irregularity has also been addressed within popular generative AI paradigms, such as \acp{VAE} and diffusion models. These learning workflows are model-agnostic, meaning they can potentially be implemented using any of the neural network structures mentioned above.

However, significant challenges remain that limit the widespread adoption and scalability of these methods. A critical concern is computational efficiency when dealing with high-dimensional systems containing irregular data, as the resource demands of ML algorithms often scale exponentially with complexity, particularly for graph-based or transformer-type neural networks. Another pressing issue is the limited generalisability of developed ML models to unseen scenarios or data grids. While this challenge is common for structured data, it becomes even more pronounced for unstructured data points. Models trained on specific geometries or boundary conditions may struggle to adapt to novel configurations, topologies, or external forcing mechanisms, such as multi-physics interactions or dynamic environmental changes.

Furthermore, the lack of comprehensive and general-purpose datasets for benchmarking different approaches remains a significant challenge for the community. Existing datasets often target specific domains, making it difficult to objectively compare the performance of various ML techniques. The development of standardised, diverse datasets that encompass a wide range of conditions and topologies is essential for fostering progress. Similarly, the absence of suitable evaluation metrics for comparing systems that predict irregular data adds to the complexity. Metrics that account for irregularities in spatial resolution, topological structure, and multi-scale characteristics are needed to ensure fair and meaningful comparisons.

By tackling these challenges, interdisciplinary research combining computational physics, advanced machine learning, and applied mathematics can pave the way for more robust, scalable, and generalisable models, unlocking the full potential of ML in modelling unstructured grid data.


\label{sec:conclusion} 

\clearpage

\section*{Notations}
\begin{table*}[htp]
    \centering
\begin{tabular}{ p{3.5cm} p{15cm}}
$\bx^t$ & current state vector/field in the full space ($t$ is the time) \\
$x_{i,j}^t$ & one element (of index $\{i,j\}$) of the state field \\
$\tilde{\bx}^t$ & state vector in the reduced space\\
$\hat{\bx}$ & predicted field vector\\
$\{\bX\}$ & an ensemble of state vectors\\
$\mathbf{i}$ & vector of node coordinates\\
$\tilde{n}$ & dimension of the compressed state\\
$ \mathcal{L}$ & loss function in training neural networks \\
$\mathcal{D}$ and $\mathcal{E}$ & decoder and encoder for state variables ($\tilde{\bx} = \mathcal{E}(\bx), \bx = \mathcal{D}(\tilde{\bx})$)\\
$\bW^l$ & weight matrix of the $l^{th}$ layer in the neural network\\
$\nabla f$ & gradient of a function $f$ \\
$\frac{\partial f}{\partial x}$ & partial derivative of function $f$ with respect to variable $x$ \\
$\mathbb{E}[\mathbf{x}]$ & expected value of a random vector $\mathbf{x}$ \\
$\mathcal{N}(\mu, \sigma^2)$ & normal distribution with mean $\mu$ and variance $\sigma^2$ \\ 
$\bU, \bV$ & unitary matrices in POD\\ 
$\mathbf{\Sigma}$ & diagonal matrix in POD\\ 
$\bI$ & Identity matrix\\ 
$\phi_s$ & POD vectors\\ 
$\|\bx\|$ & norm of vector $\bx$ \\ 
$\bA^G$ & adjacency matrix of graph $\mathbf{G}$\\
$\bV^G$ & node-level features of graph $\mathbf{G}$\\
$\mathbf{E}^G$ & edge-level features of graph $\mathbf{G}$\\
$Q,K,V$ & attention head in transformer\\ 
$N_b$ & total number of collocation points on the boundary \\
$N_p$ & total number of collocation points inside the domain \\
$N_d$ & total number of data points where direct measurements are available \\
$N_v$ & total number of data points in the validation dataset \\
$\mathcal{L}_{PDE}$ & PDE loss \\
$\mathcal{L}_{BC}$ & boundary condition loss \\
$\mathcal{L}_{IC}$ & initial condition loss \\
$\mathcal{B}$ & geometry boundary \\
$\mathcal{R}$ & reward for assessing the quality of a new mesh element \\
$Q^v$ & action value in reinforcement learning\\
$G^r$ & cumulative rewards in reinforcement learning\\
\end{tabular}
\end{table*}
\clearpage

\section*{Acronyms}
\begin{acronym}[htp]
\footnotesize{
\acro{2D}{two-dimensional}
\acro{AE}{Autoencoder}
\acro{ARM}{Autoregressive model}
\acro{CFD}{Computational fluid dynamics}
\acro{CNN}{Convolutional neural network}
\acro{DDM}{Denoising diffusion model}
\acro{DMD}{Dynamic mode decomposition} 
\acro{ELBO}{Evidence lower bound}
\acro{FNO}{Fourier neural operator}
\acro{GAN}{Generative adversarial network}
\acro{GNN}{Graph neural network}
\acro{KNN}{K-Nearest neighbours}
\acro{LSTM}{Long short-term memory}
\acro{MAE}{Mean absolute error}
\acro{ML}{Machine learning}
\acro{MLP}{Multi layer perceptron}
\acro{MSE}{Mean squared error}
\acro{NERF}{Neural radiance fields}
\acro{NLP}{Natural language processing}
\acro{ODE}{Ordinary differential equation}
\acro{PCA}{Principal component analysis}
\acro{PDE}{Partial differential equation}
\acro{PINN}{Physics-informed neural network}
\acro{POD}{Proper orthogonal decomposition}
\acro{EOF}{Empirical orthogonal functions}
\acro{Re}{Reynolds number}
\acro{RANS}{Reynolds-averaged Navier-Stokes}
\acro{RF}{Random forest}
\acro{RL}{Reinforcement learning}
\acro{RNN}{Recurrent neural network}
\acro{ROM}{Reduced-order modelling}

\acro{SDE}{Stochastic differential equation}
\acro{SINDy}{Sparse identification of nonlinear dynamics}
\acro{SVD}{Singular value decomposition}
\acro{VAE}{Variational Autoencoder}
\acro{XGBoost}{eXtreme gradient boosting}
}

\end{acronym}

\section*{Acknowledgement}
Sibo Cheng acknowledges the support of the French Agence Nationale de la Recherche (ANR) under reference ANR-22-CPJ2-0143-01. Weiping Ding acknowledges the support of the National Key R\&D Plan of China under Grant 2024YFE0202700, the National Natural Science Foundation of China under Grant  U2433216, and the Natural Science Foundation of Jiangsu Province under Grant BK20231337. Jinlong Fu acknowledges the financial support from Alexander von Humboldt Foundation. Dunhui Xiao acknowledges the Top Discipline Plan of Shanghai Universities-Class I and Shanghai Municipal Science and Technology Major Project (No.2021SHZDZX0100), National Key R\&D Program of China(No.2022YFE0208000). This work is supported in part by grants from the Shanghai Engineering Research Center for Blockchain Applications And Services (No.19DZ2255100) and the Shanghai Institute of Intelligent Science and Technology, Tongji University. CEREA is a laboratory of Institut Pierre-Simon Laplace.

\footnotesize
\bibliographystyle{elsarticle-num-names}
\bibliography{main.bib}
\end{document}